\documentclass[preprint,12pt]{elsarticle}

\usepackage{graphics}
\usepackage{epsfig}
\usepackage{soul}
\setstcolor{red}

\usepackage{subfigure}

\usepackage{amssymb}
\usepackage{textcomp}
\usepackage{gensymb}
\usepackage{xcolor}
\usepackage {appendix}
 \usepackage{amsthm}

\usepackage{lineno}

\usepackage{tabularx}
\usepackage{caption}
\usepackage{amssymb}
\usepackage{latexsym}
\usepackage{amsmath}
\usepackage{placeins}

\usepackage{url}
\usepackage{soul}
\usepackage{xcolor}
\usepackage{tabularx}
\usepackage{graphicx}
\usepackage{xcolor}
\usepackage{algorithmic}
\usepackage{algorithm}
\usepackage[colorlinks = true,
            linkcolor = blue,
            urlcolor  = blue,
            citecolor = blue,
            anchorcolor = blue]{hyperref}
\usepackage{amsmath,amssymb,amsfonts}

\usepackage{array}

\renewcommand{\appendixname}{Appendix}

\usepackage{adjustbox}

\usepackage{cancel}

\usepackage{algorithm}

\usepackage{float}

\makeatletter
\def\algbackskip{\hskip-\ALG@thistlm}
\makeatother

\usepackage{url}
\usepackage{amsmath}
\usepackage{hyperref}

\usepackage{fullpage}
\usepackage{tikz}
\usetikzlibrary{arrows,%
                shapes,positioning}

\usepackage{hyperref}
    \hypersetup{
        colorlinks   = true,
        citecolor    = blue
    }
\journal{}

\makeatletter
\def\ps@pprintTitle{%
 \let\@oddhead\@empty
 \let\@evenhead\@empty
 \def\@oddfoot{}%
 \let\@evenfoot\@oddfoot}
\makeatother

\renewcommand\refname{References}

\usepackage{lineno}
\renewcommand{\appendixname}{Appendix}

\begin{document}

\begin{frontmatter}



%\title{High-resolution dynamic MRI reconstruction to investigate Biomechanics of the Ankle Joint in children with cerebral palsy}
%\title{Non-invasive characterization of surface motion patterns in highly deformable soft tissue organs from dynamic Magnetic Resonance Imaging}
\title{Characterization of surface motion patterns in highly deformable soft tissue organs from dynamic MRI: An application to assess 4D bladder motion}

\author[]{Karim Makki $^{a,*}$, Amine Bohi$^{a}$, Augustin C. Ogier$^a$, and Marc Emmanuel Bellemare$^a$}
\address[$^{*,1}$]{Aix Marseille Univ, Université de Toulon, CNRS, LIS, Marseille, France\\}
% $^*$ These authors contributed equally to this work}

\begin{abstract}
\textit{Background and objectives:} Dynamic Magnetic Resonance Imaging (MRI) may capture temporal anatomical changes in soft tissue organs with high-contrast but the obtained sequences usually suffer from limited volume coverage which makes the high-resolution reconstruction of organ shape trajectories a major challenge in temporal studies. Because of the variability of abdominal organ shapes across time and subjects, the objective of the present study is to go towards 3D dense velocity measurements to fully cover the entire surface and to extract meaningful features characterizing the observed organ deformations and enabling clinical action or decision. \\

\textit{Methods:} We present a pipeline for characterization of bladder surface dynamics during deep respiratory movements. For a compact shape representation, the reconstructed temporal volumes were first used to establish subject-specific dynamical 4D mesh sequences using the large deformation diffeomorphic metric mapping (LDDMM) framework. Then, we performed a statistical characterization of organ dynamics from mechanical parameters such as mesh elongations and distortions. Since we refer to organs as non-flat surfaces, we have also used the mean curvature changes as metric to quantify surface evolution. However, the numerical computation of curvature is strongly dependant on the surface parameterization (\textit{i.e}. the mesh resolution). To cope with this dependency, we employed a non-parametric method for surface deformation analysis. Independent of parameterization and minimizing the length of the geodesic curves, it stretches smoothly the surface curves towards a sphere by minimizing a Dirichlet energy. An Eulerian PDE approach is used to derive a shape descriptor from the curve-shortening flow. Intercorrelations between individuals' motion patterns are computed using the Laplace–Beltrami Operator (LBO) eigenfunctions for spherical mapping.\\

\textit{Results:} Application to extracting characterization correlation curves for locally-controlled simulated shape trajectories demonstrates the stability of the proposed shape descriptor. Its usability was shown on MRI acquired for seven healthy participants for which the bladder was highly deformed by maximum of inspiration. As expected, the study showed that deformations occured essentially on the top lateral regions.\\

\textit{Conclusion:}
Promising results were obtained, showing the organ in its 3D complexity during deformation due to strain conditions. Smooth genus-0 manifold reconstruction from sparse dynamic MRI data is employed to perform a statistical shape analysis for the determination of bladder deformation. 

\end{abstract}


\end{frontmatter}

\section{Introduction}

Pelvic floor disorders affect approximately 50\% of women older than 50 years~\cite{law2008mri}. Related health problems such as urinary and fecal incontinences get worse with age which affects activities of daily living. 
Dynamic MRI examinations are now essential for the investigation of the pelvic area as they allow a non-invasive observation of the main organ deformations~\cite{el2017magnetic}. Current clinical practice involves 2D dynamic MRI acquiring a single sagittal plane per time frame~\cite{jourdan2021semiautomatic}. However, in-plane and out-of-plane deformations can simultaneously occur, thus limiting the ability to assess the effective 3D deformations from the 2D images. In the context of pelvic floor, 3D imaging has remained exclusive to static MRI so far, in order to observe pelvic floor anatomy at rest during instructed \emph{apnea}.
In fact, the MRI-modalities are still too slow to follow the movements within a reasonably short duration. Moreover, most of non‐invasive and fast 3D acquisitions suffer from intrinsically low volume rates~\cite{garetier2020dynamic,chen2021blood}. 

Assuming that $58\%$ of women who undergo surgery have reported a recurrence of pelvic disorders~\cite{gorji2020effect}, one of the most challenging tasks is to enhance the understanding of the organ connections and their morphological changes under strain conditions.
Beyond visual inspection by a radiologist, a more relevant quantitative information about pelvic organ deformations is required to support surgeons' needs for optimization of implant position and post-surgical follow-up of patients.
To address these needs, some recent studies~\cite{chen2015female,courtecuisse2020three} have begun to create a biomechanical model of organ interactions, while other studies focused on the role of 3D reconstruction in the assessment of organ motion~\cite{ogier20193d,makki2020new}. Both these approaches pave the way for a better understanding of organ behaviors and they would profit from an efficient descriptor of organ temporal 3D deformations.
Indeed, such a tool could be appreciable to simplify the visualization and the statistical identification of the most common deformations, and to compare organ dynamics simulations to organ dynamics observations, to name a few gains.
Furthermore, comparing organ movement between patients would allow to identify and grade abnormal motion in order to classify patients into healthy and pathological, or into subgroups sharing similar organ motion characteristics within large data sets, which will optimize surgical and non-surgical treatments by avoiding unnecessary duplication of effort across patients.

% Surgeons and clinicians can also benefit from comparing bladder movement between two or more patients 

% is to use this variance–covariance matrix for further robustifying
% the

% we perform characterization of organ surface motion patterns employ realistic boundary conditions which are typically able to provide good prior knowledge about such deformations in order to robustify and validate existing simulators of the dynamic behavior of the pelvic organs}~\cite{chen2015female}.

From a methodological point of view, \textit{in vivo} characterization of the dynamical behavior of human joints, organs, and soft tissues during daily physical activities remains challenging because of the complexity and the non-linearity of their shape dynamics~\cite{bruning2020characterization,pennec2020advances, zheng2019unsupervised, abbas2019analysis, hong20184d, makki2019vivo}. During the last decade, statistical shape analysis tools have been shown to be clinically useful, in particular, to help understand patterns in large clinical data sets and to characterize the functioning soft tissue organs or anatomical structures undergoing large deformations~\cite{heimann2009statistical,pennec2020statistical,fishbaugh2017geodesic,zhang2016statistical,fishbaugh2013geodesic,abi2020monotonic}. However, one should keep in mind that simply encoding deformations with some displacement vector fields is not sufficient to identify clinically relevant patterns of the induced organ deformations because of the complexity of their dynamics.
% (i.e. the Jacobian determinant of the deformation field is not a meaningful parameter in statistical characterization).
Therefore, space-time statistics should be based on robust and stable surface features or even better, geometric shape descriptors\footnote{The two terms are used interchangeably throughout the paper.}. In other words, the temporal feature changes should characterize the local shape deformations. Since shape space is not necessarily flat, then statistical tools derived from Euclidean geometry are not well adapted to study organ shape dynamics. Under such circumstances, it is practically meaningful to employ Riemannian geometry for generating shape trajectories which belong to non-linear
manifolds~\cite{zolfaghari2014multiscale, bone2018learning}. Relying on the notion of geodesic distances on manifolds to compute the shortest paths between two points on a curved surface~\cite{peyre2009geodesic}, these tools cover topics that start from the fundamentals of general relativity theory~\cite{malament2012remark}. 

In practice, several studies for characterizing spatio-temporal shape trajectories have used longitudinal datasets aiming at quantifying: the shape growth over a long period of time~\cite{sun2015fast} (intrasubject variations in geometry), or the shape variability in an inter-subject context (\textit{e.g.} for atlas building at the population scale~\cite{kim2020framework}). Longitudinal data consist of a collection of stationary scans (\textit{e.g.} conventional high-resolution MRI scans) portraying anatomical changes during a large period of time (over weeks, months, or even years depending on the studied pathology or phenomenon). Such studies have served to highlight the need for clinicians to understand anatomical changes occurring during development or disease progression (e.g. linear geodesic regression model for shape time-series with sparse parameters~\cite{fishbaugh2013geodesic, fishbaugh2017geodesic}). In the context of spatio-temporal bladder imaging~\cite{rios2017population}, longitudinal samples are often limited by bladder volume variation which may cause an increase in scaling effects. A solution is to perform short-time non invasive imaging in order to assess the effective deformations of constant bladder volumes. The latter condition is essential to direct the radiotherapy of pelvic tumors with precision~\cite{luo2016interfractional}.

As compared to longitudinal studies, temporal studies using dynamic MRI may non‐invasively capture anatomical changes during motion over a short period of time, but at the cost of having: (1) less spatio-temporal resolution of anatomical data sequences, (2) higher sensitivity to motion artifacts, and (3) an increased effect of image noise. 
Overcoming these difficulties that appear in sparse spatio-temporal data is a whole new challenge, especially when it becomes necessary to have a high spatio-temporal resolution of shape trajectory in order to carry out statistically significant studies. 
%\st{In the context of spatio-temporal bladder imaging}~\cite{rios2017population}, \st{longitudinal samples are often limited by bladder volume variation which may cause an increase in scaling effects. A solution is to perform short-time non invasive imaging in order to assess the effective deformations of constant bladder volumes. The latter condition is essential to direct the radiotherapy of pelvic
%tumors with precision}~\cite{luo2016interfractional}.

% Surmounting these difficulties arising in the data sequences represents a newest challenge especially when a high-resolution temporal reconstruction of shape trajectory becomes necessary for performing  statistically significant studies. 

In this paper, we have employed diffeomorphic statistical shape tools to evaluate organ surface dynamics. We show the strength of the approach by characterizing 3D MRI observations of bladder under strain conditions. Starting from sets of high resolution reconstructed temporal bladder volumes, we have first employed the large deformation diffeomorphic metric mapping (LDDMM) framework to encode the organ's large deformations with a reduced number of significant surface points while covering properties in shape geometry. A dynamical quadrilateral mesh for the organ surface is then established to address the need for preserving neighborhood structures when computing geometric descriptors, that are classically used in biomechanics, such as mesh elongations and distortions. Then, characterizations of organ shape dynamics are performed by computing mean curvature changes. Moreover, a new geometric feature is also proposed. It allows detecting salient motion patterns from  geodesic shortest length paths for mapping a shape to a sphere. It satisfies the "invariance" conditions of Kendall's shape space~\cite{kendall1984shape}, by filtering out location, size and rotation. This feature may capture local surface variations, with no need to compute Riemann's tensor. This last point is an interesting advantage as computing tensors not only incurs high computational costs but also impacts numerical stability~\cite{nava2019geodesic}. Results demonstrate that the proposed feature, possessing a high repeatability score of measures throughout cyclic shape trajectories, is much more numerically stable.

Since we are interested in comparing organ shapes in terms of their geometry rather than their size, we performed all comparisons in a common \textit{shape space}. Indeed, motion patterns derived from the different feature vectors were projected onto the unit sphere $\mathbb{S}^2$ (point-to-point anatomical correspondences between these geometric features were established across subjects).

% , in order to compute intercorrelations between individuals' motion patterns with the help of the Laplace–Beltrami Operator (LBO) eigenfunctions for spherical mapping~\cite{lefevre2015spherical}.
  
\section{Related work}
In the literature, some studies show attempts to quantify or model human soft tissue in a non-invasive manner during daily living activities~\cite{billet2008cardiac, pennec2020statistical, makki2019vivo}. In~\cite{lee2014semi}, the authors proposed a semi-automatic segmentation framework to track the motion of the tongue and to measure its internal deformation during speech and swallowing using dynamic MRI. In contrast, the temporal resolution was limited and the tongue trajectory was only represented by $26$ reconstructed volumes, each with a voxel size of $1.875 \times 1.875 \times 1.875$ mm. 
In ~\cite{makki2018high, makki2019vivo}, a first attempt to quantify the ankle joint motion patterns through a combination of static and dynamic MRI data was presented. A tracking of bones and surrounding soft tissues was performed by estimating a dense deformation field covering the entire field of view from the static scan to dynamic time frames using the Log Euclidean Polyrigid registration Framework (LEPF)~\cite{arsigny2009fast}.  
However, in the context of pelvic floor dynamics, most biomechanics simulation experiments have been hampered by the lack of relevant data due to limitations of spatio-temporal resolution of dynamic MRI~\cite{chen2015female, courtecuisse2020three}. In~\cite{courtecuisse2020three}, 2D dynamic MRI images were combined with 3D biomechanical models in order to extrapolate the complete 3D dynamic motion of abdominal organs. A validation attempt was performed by checking that the reconstructions were well conducted from the first scan towards the end of dynamic sequence. However, the validation itself was not founded on a clinically relevant ground truth (i.e. the validation was not performed in the high resolution domain). Furthermore, only two simulated sequences have been used to validate the model. 

The above issues were addressed in~\cite{ogier20193d}: a high-resolution spatio-temporal reconstruction of the non-linear dynamics of an abdominal organ motion was introduced for designing a continuous-time dynamics that allowed us to infer inter-frame deformations.
Promising results were obtained, showing the bladder in its 3D complexity during deformation due to strain conditions with an estimation of the most deformed tissue areas. However,  bladder motions were quantified based on the temporal changes in Jacobian determinant of the estimated deformation fields. Although this parameter can quantify the organ volume changes, it fails to identify its local morphological changes throughout motion. Since we are interested in comparing differences in geometry (rather than in size), these high-resolution temporal data were then employed to introduce a compact characterization of moving bladder surfaces through the use of a geodesic-based shape descriptor in~\cite{makki2020new}. In this paper, we have extended the study group to include more subjects and we have proposed a technique to simulate pathological data sequence. Moreover, more results for comparing between shape descriptors and individual motion patterns were provided to show how the use of such methods is intended to ultimately inform clinical practice and research.
% To achieve full volume coverage, these issues have already been addressed in~\cite{ogier20193d}, where the reconstruction of realistic 3D bladder trajectories was assessed by preserving the static volume undergoing large deformations to fit each of the low resolution volumes. The reconstruction pipeline was based on diffeomorphic registration (ANTs tools) to perform inter-slice registration over a restricted field of view (using bounding boxes around the organ of interest to accelerate the alignment process). A set of different configurations for dynamic data acquisitions were used and the reconstruction consisted of filling the inter-slice gaps.

In~\cite{rahim2013diffeomorphic}, a characterization of pelvic organ dynamics was proposed using diffeomorphic registration on dynamic 2D slices. The objective of the present study is to go towards 3D dense velocity measurements to fully cover the entire surface. Furthermore, most of the descriptors used in~\cite{rahim2013diffeomorphic} were based on Euclidean geometry (\textit{i.e.} the geometry of a flat space). Proposing new shape descriptors which generalize Euclidean geometry to non-flat or curved spaces eventually becomes necessary with the increase in the number and complexity of motion patterns to be recognized. 

\section{Methods}

\subsection{Dynamic quadrilateral mesh}
\label{surface_param}

\subsubsection{Surface parameterization}

% A surface is represented by a structured mesh (K,V) where: $V=\{v_1,v_2,\ldots v_n \}$, $v_i \in \mathbb{R}^3$ is the set of vertex positions; and $K$ is a simplicial complex determining the topological type of the mesh by specifying the connectivity of vertices, faces, and edges.

% specifying  \\

\begin{figure}[ht]
\centering
%\begin{minipage}{0.36\textwidth}
\subfigure{\includegraphics[scale=0.2]{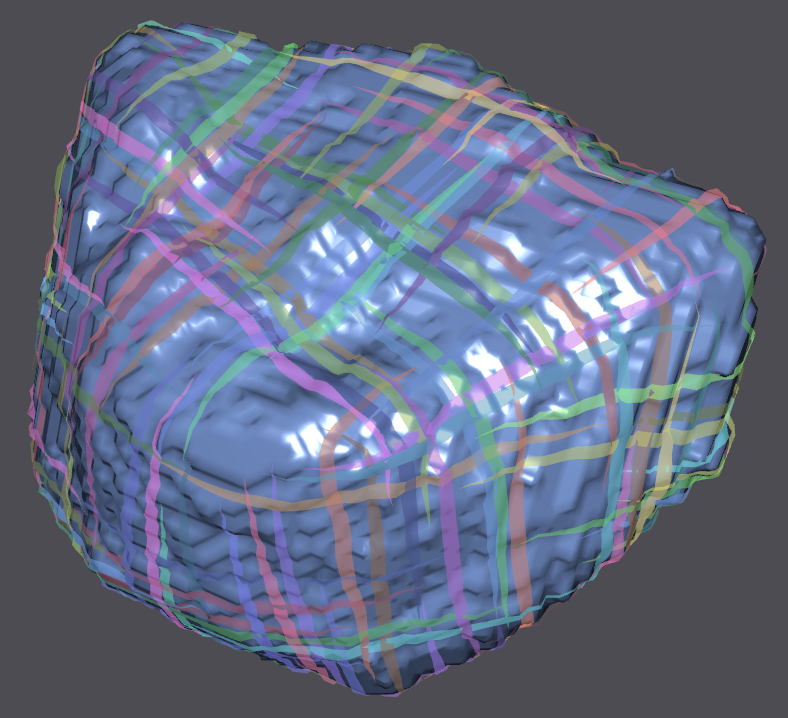}}
\hspace{0.2cm}
\subfigure{\includegraphics[scale=0.195]{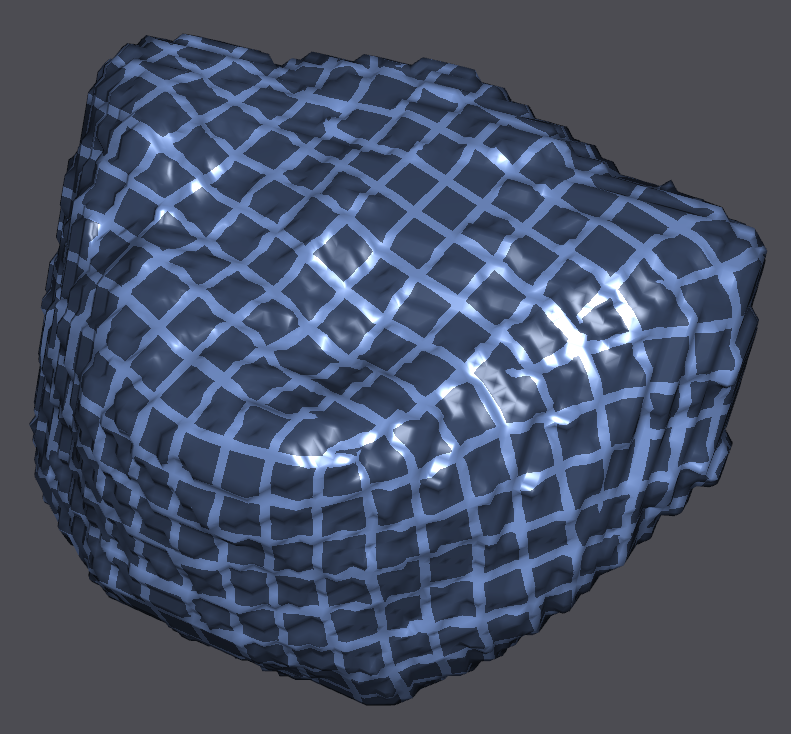}}
\hspace{0.2cm}
\subfigure{\includegraphics[scale=0.201]{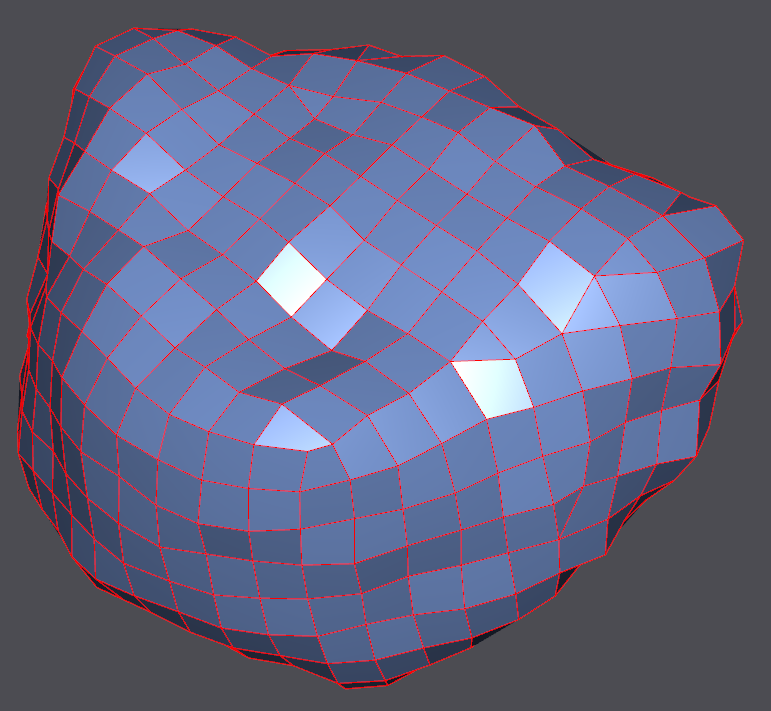}}
\hspace{0.2cm}
\subfigure{\includegraphics[scale=0.204]{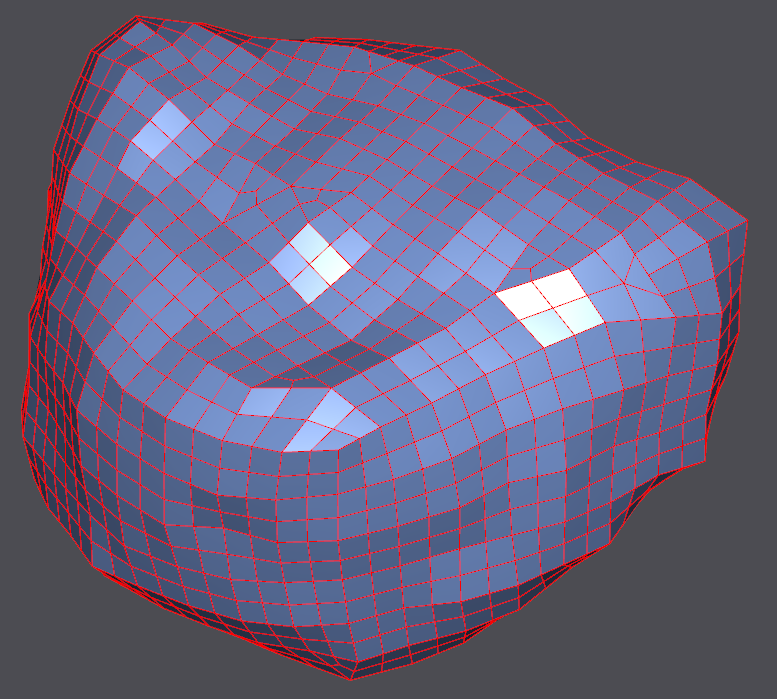}}
\caption{\label{spatial_parameterization} Parameterization of the bladder surface, from left to right: the orientation field, the position field, the quad dominant mesh, and the resulting pure quad mesh.}
\end{figure}

% To encode the shape surface with an extremely low number of meaningful variables, we have first extracted iso-surfaces from the reconstructed organ volumes (\textit{i.e.} 3D binary masks) at the first time frame using the marching cubes algorithm. The marching squares/cubes is the standard algorithm to extract
% iso-curves/surfaces from a discretized image/volume~\cite{lewiner2003efficient}. 
In a first step, we have extracted an initial control mesh from the reconstructed bladder volumes (\textit{i.e.} 3D binary masks obtained using the methods introduced in~\cite{ogier20193d}) at the first time frame using the marching cubes algorithm~\cite{lewiner2003efficient}. The simplicial surface is then converted to a quasi-regular quadrilateral mesh $(K,\mathcal{M})$ such that $\mathcal{M}=\{x_1,x_2,\ldots x_n \}$, $x_i \in \mathbb{R}^3$ is the set of vertex positions, and $K$ is a cubical complex determining the topological type of the mesh by specifying the connectivity of vertices, faces, and edges. To fully cover the entire surface with an extremely low number of meaningful variables, we have used a robust algorithm presented in~\cite{jakob2015instant}, providing a uniform distribution of vertices over the manifold by iteratively refining the initial control mesh until a pure quad mesh is obtained. This algorithm avoids irregularity problems at the shape poles, such as vertex singularities encountered in~\cite{chen2015female}. Fig.~\ref{spatial_parameterization} illustrates the quality of an obtained quadrilateral mesh.

\subsubsection{Estimation of smooth vertex trajectories}
In a second step, we propose to track the mesh vertices during respiratory motion while preserving their connectivities. This allows for constructing a spatio-temporal structured meshes which might also be used for deriving some biomechanical properties of the organ dynamics such as strains and stresses using finite element methods which are required for establishing  biomechanical models (\textit{i.e.} pure quad meshes are often desired in CAD applications).

Our model for shape analysis focus on deformations represented by diffeomorphisms acting on landmarks.
To determine point correspondences, a mesh-to-volume registration is performed. In other words, we propose to track the set of mesh vertices using the LDDMM framework that has been heuristically shown to produce natural deformation paths in the space of diffeomorphisms without having to use point-to-point correspondences between source and target point sets~\cite{beg2005computing}. 
% This framework which is based on Hamiltonian dynamics and statistical mechanics allows one to provide an hypothesis compatible with the physics of deformations by estimating geodesic shape curves without having to use point-to-point correspondences between source and target point sets. 
A smooth and continuous-time trajectory of each vertex is estimated throughout the organ range of motion (vertices travel along geodesic curves).  

The principle of control-points-based LDDMM for estimating a diffeomorphic mapping is as follows: \\
Given a set of $N$ control points $\{q_i\}_{i \in 1,\ldots,N}$, and a set of $N$ corresponding momentum vectors of $\mathbb{R}^3$  $\{\mu_i\}_{i \in 1,\ldots,N}$, the velocity vector in the tangent space $\mathcal{T}_x \mathcal{M}$ of the parametric surface $\mathcal{M}$ at a point $x \in \mathcal{M}$, is obtained through the use of a Gaussian convolution filter:
\begin{equation}
    v : x \in  \mathcal{M} \mapsto v(x)=\sum_{i=1}^{N}K(x,q_i).\mu_i
    \label{gauss_kernel}
\end{equation}
where $K(x_i,x_j) = exp(-||x_i-x_j||^2/ \sigma ^2)$ is a Gaussian kernel to ensure smooth geodesic shooting.

The temporal evolution of the organ velocity vector field can be modeled by the following Hamilton's equations of motion:

\begin{equation}
\begin{cases} \dot{q}(s) = K(q(s),q(s)).\mu(s) \\

\dot{\mu}(t) = -\frac{1}{2} \nabla_q \{K(q(s),q(s)). \mu(s)^{\top}\mu(s)\} \end{cases}
\end{equation}

The solutions to this Hamiltonian are then the same as the geodesics on a Riemannian manifold. The numerical integration of these PDEs, performed using a second-order Runge-Kutta scheme, gives a flow of a time-dependent velocity vector field parameterized with $q(s)$ and $\mu(s)$:
\begin{equation}
    v : x \in  \mathbb{R}^3 \times s \in [0,1] \mapsto  v(x,s)=\sum_{i=1}^{N}K(x,q_i(s)).\mu_i(s)
    \label{eq:temporal_vvf}
\end{equation}

The temporal displacement of each tracked point $x \in \mathcal{M}$  is governed by the following autonomous first order ODE:

\begin{equation}
   \dot{x}(s) = v(x(s),s) \quad subject \quad to \quad x(0)=x
\end{equation}

Finally, the solution of this ODE yields a flow of diffeomorphisms starting from the source points (i.e. starting from the identity in the space of transformations), $\Phi_{q,\mu}(.,s) :  \mathbb{R}^3 \times [0,1]  \mapsto \mathbb{R}^3$, such that $\Phi_{q,\mu}(.,1) = Id + \int_{0}^{1} v(\Phi_{q,\mu}(.,s)) ds $ is the
end-point of the geodesic flow matching the given point sets.

The overall algorithm for vertex tracking is described in Algorithm~\ref{algomotion}, with the following notations:  
$L$ is the length of the dynamic sequence, $\mathcal{M}_t$ gives the locations of mesh vertices at time $t$, $\mathcal{C}_t$ is the entire 3D surface point cloud at time $t$ ($\mathcal{M}_t$ is a proper subset of $\mathcal{C}_t$). Note that the registration problem is solved by iteratively minimizing the following loss function:
\begin{equation}
  f(q,\mu)= d(\mathcal{C}_{t+1} , \Phi_{q,\mu}(\mathcal{M}_t)) + R(q,\mu)
\label{loss_function}
\end{equation}
where the first term measures data-attachment while the second regularization term represents the norm of the deformation.
% Some results of our tracking algorithm are illustrated in Fig.~\ref{fig:tracking_algorithm}. 

\begin{algorithm}
\caption{LDDMM-based 4D quad mesh generation}
\begin{itemize}
\item \textbf{Inputs:} - Initial mesh structure $(K, \mathcal{M}_0)$.\\
- Reconstructed segmentation contours $\mathcal{C}_{t,\quad t \in\{1 \ldots L-1\}}$   
\item \textbf{Motion estimation:} Estimate forward successive vertex trajectories using the LDDMM $\{ \mathcal{M}_{t+1}\}_{t~=~0,\ldots,L-1}$ such that $\mathcal{M}_t \subset \mathcal{C}_t $, by aligning $\mathcal{M}_{t}$ and $\mathcal{C}_{t+1}$:

\begin{itemize}
\item \textbf{for} $t$ \textbf{in} $0\ldots L-1$:
\end{itemize}
\hspace{1.2cm} 1. Initialize $\{q_i^t(s)\}_{i \in 1,\ldots,N}$, and $\{\mu_i^t(s)\}_{i \in 1,\ldots,N}$.

\hspace{1.2cm} 2. Compute velocities $v^t$ according to Eq~\ref{eq:temporal_vvf}.

\hspace{1.2cm} 3. Integrate the flow $\Phi_{q,\mu}^t(x,s)=  Id_{\mathbb{R}^3} + \int_{0}^{1} v^t(\Phi_{q,\mu}(x,s)) ds, \quad  \forall x \in \mathcal{M}_{t}$, by 

\hspace{1.7cm} minimizing the cost function defined in Eq~\ref{loss_function}.

% \begin{enumerate}
% \item   
% \end{enumerate}

\item \textbf{Output:} 4D quad mesh sequence $(K, \mathcal{M}_t)_{t~=~0,1,\ldots,L-1}$.
\end{itemize}
\label{algomotion}
\end{algorithm}

\begin{figure}[ht]
    \centering
    \includegraphics[width=1.0\linewidth]{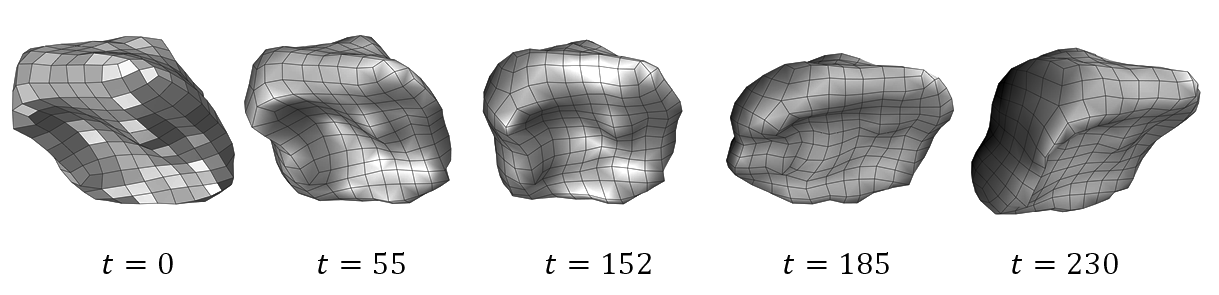}
    \vspace{-0.7cm}
    \caption{Example of 4D reconstructed quad mesh for the bladder during forced respiratory motion.}
    \label{fig:4Dmesh}
 \end{figure}

To validate the tracking process, we propose to compute the following error:

\begin{equation}
  E = \frac{1}{n} \sum_{p=1}^{n} dist(x_p,\mathcal{C}_{L-1})
\end{equation}

where: $n$ is the total number of tracked vertices, $x_p \in \mathcal{M}_{L-1}$, and $dist(x_p,\mathcal{C}_{L-1})$ is the Euclidian $\ell^2$ distance between $x_p$ and the closest point $x_l$ in the last reconstructed surface $\mathcal{C}_{L-1}$. A propagated mean error of $0.63 \pm 0.06$ mm was obtained across all subjects. The resulting error was always inferior to $1$  mm which reflects the tracking accuracy level for a given isotropic voxel size of $1\times1\times1$ mm. Fig.~\ref{fig:4Dmesh} shows the quality of our 3D+$t$ quadrilateral mesh reconstruction based on smoothly tracking vertices using the LDDMM while keeping connectivity unchanged.

% \begin{figure}[ht]
% \centering
% \subfigure{\includegraphics[scale=0.2]{point_tracking_view1.png}}
% \hspace{0.008cm}
% \subfigure{\includegraphics[scale=0.201]{point_tracking_view2.png}}
% \hspace{0.008cm}
% \subfigure{\includegraphics[scale=0.2]{point_tracking_view3.png}}
%     \caption{Vertex Set tracking \hl{for 3 separate deformation examples}: $\mathcal{M}_0$ in red, $\mathcal{M}_t$ in green, and $\mathcal{C}_t$ in white.}
%     \label{fig:tracking_algorithm}
% \end{figure}
\subsection{Shape descriptors}
% In this section, we present the geometric descriptors that we have used to characterize the organ dynamics from the reconstructed quad meshes.
\subsubsection{Mesh elongation and distortion}

Shape elongations and distortions have been used in medical imaging, in particular to characterize the deformations of pelvic organs in the plane from dynamical shape
contour points~\cite{rahim2013diffeomorphic}. However, such characterizations
are prone to biased interpretation of organ motion patterns because of the out-of-plane problem. 
To overcome these limitations, we propose
a 3D extension method of these 2D feature maps, compatible with the reconstructed quad meshes.

\subsection*{Mesh elongation:}
The mesh elongation, equivalent to the Green-Lagrange deformation descriptor, is a commonly used feature in biomechanics for characterizing local mesh deformations based on the spacing changes in the neighborhood of each vertex. Consider a vertex $x_{i}^{j}(t)$ and its four nearest neighbors $\{x_{i+1}^{j}(t), x_{i-1}^{j}(t), x_{i}^{j+1}(t), x_{i}^{j-1}(t)\}$ belonging to the mesh at the $t^{th}$ time frame, and a vertex $x_{i}^{j}(t+1)$ and its neighbors $\{x_{i+1}^{j}(t+1), x_{i-1}^{j}(t+1), x_{i}^{j+1}(t+1), x_{i}^{j-1}(t+1)\}$, their homologous vertices at the $(t+1)^{th}$ time frame (where $i$ and $j$ are used for indexing mesh local connectivities). We compute $\hat{d}(t)$ and $\hat{d}(t+1)$, the average Euclidean distance between $x_{i}^{j}(t)$ and its neighbors (resp. between $x_{i}^{j}(t+1)$ and its neighbors). Then, the elongation measure is given by: 
\begin{equation}
E(k)= \frac{\hat{d}(t+1) - \hat{d}(t)}{2 \hat{d}(t)}    
\end{equation}
If $E(t)=0$, then no deformation has occurred, else if $E(t)>0$, the neighborhood of the vertex has expanded. Otherwise, the neighborhood has shrunk.

\subsection*{Mesh distortion:} 

The dihedral angles are defined as the angles between the two normal vectors to each two adjacent faces~\cite{vavsa2012dihedral}. And the temporal change in these angles during motion is considered as the temporal distortion of the quad mesh.

For two adjacent quad faces $F_1=\{x_1,x_2,x_3,x_4\}$ and $F_2=\{x_1,x_2,x_{3}^{'},x_{4}^{'}\}$, with normals $N_1$ and $N_2$, respectively, the local dihedral angle is computed as follows:

\begin{equation}
\theta = \arccos\left({\frac{N_1 \cdot N_2}{||N_1||\times ||N_2||}}\right)
\end{equation}
    
% Let $V_{i}^{j}(t)$ and $V_{i}^{j}(t+1)$ be the mesh vertices corresponding to the same tracked surface point at times $t$ and $t+1$, respectively.

Assuming that the adjacent faces to each vertex of the initial quad mesh (at $t=0$) were maintained the same throughout the sequence, then both vertices $x_{i}^{j}(t)$ and $x_{i}^{j}(t+1)$ have always the same adjacent faces that we note  
$\{F_{1}, F_{2}, F_{3}, F_{4}\}$. If we consider $\theta(t)$ and $\theta(t+1)$ as the maximal dihedral angles among the six possible couple combinations of adjacent faces at times $t$ and $t+1$, respectively, then, the temporal mesh distortion around $x_{i}^{j}$ is determined by:
    
\begin{equation}
        D(t) = \mid{\theta(t+1) - \theta(t)}\mid
\end{equation}
    
If $D(t)=0$, then no local deformation has occurred. Otherwise, the neighborhood has distorted.

\subsubsection{A novel geodesic-based feature for characterization of organ surfaces:}
\label{sec: Dirichlet_problem}

In the field of computer graphics and computational geometry, shortest path problem on both flat and curved domains is a well-known problem. Several techniques have been proposed for estimating shortest path lengths, also known as geodesic distances, on curved domains. One of the most popular computational methods is the \textit{Dijkstra} algorithm for finding the shortest paths between nodes in a weighted graph~\cite{dijkstra1959note}. However, this algorithm computes just a rough approximation of the true distances. Its major drawback is that the direction along which distance increases is partially ignored since only horizontal and vertical displacements are allowed on the grid. Then, it becomes clear, for instance, that this method will overestimate the straight-line Euclidean distance of any diagonal path crossing a regular grid. To circumvent this problem,  \textit{Sethian} introduced the fast marching algorithm which is a far reaching generalization of the Dijkstra algorithm~\cite{sethian1996fast}. The fast marching algorithm computes the geodesic distance (in the viscosity sense) in $O(n . log(n))$ operation. Equivalently it solves the following boundary value problem (BVP) of the Eikonal PDE by front propagation:

\begin{equation}
    \left\{
\begin{array}{l}
 |\nabla L(x)|=1, \quad x \in \Omega,  \\
  L(\partial \Omega) = 0,  \\
\end{array} \right.
\label{eikonal}
\end{equation}

where $\Omega$ is an open set of $\mathbb{R}^n$ with well-behaved boundary, $n = card(\Omega)$, and $L$ stands for the distance or length. However, being nonlinear (hyperbolic), the Eikonal PDE is just simple to state but not at all easy to solve, in particular for two-point BVPs which will be the case in the present work. Subsequently, \textit{Crane et al.} proposed the heat method for geodesic distance computation~\cite{crane2017heat}. In the heat method, the shortest path problem is reformulated in a more elegant way by transforming it into two simpler linear elliptic problems, thus allowing the main problem to be solved in three steps:
\begin{itemize}
    \item Integrate the heat flow for a source point across the grid/mesh (problem 1).
    \item Compute the normalized gradient for the heat (a simple change of variable).
    \item Recover the true distance from the normalized gradient by solving a Poisson equation (problem 2).
\end{itemize}

%the first problem consists of finding the direction along which distance is increasing, and the second one consists of recovering the distance itself in just such a way that the Eikonal equation holds. \\

In this paper, we first propose to point out the relation between the heat method and an Eulerian PDE approach, that has been proposed in~\cite{yezzi2003eulerian} for estimating tissue thickness between two non-intersecting boundaries, which in its turn, decomposes the shortest path problem into two elliptic problems using the same change of variable in-betweens. Then, we combine them to derive a robust shape descriptor which is purely based on geodesic distances and which will enable us to quantify the non-linear deformations of organ surface with respect to the sphere.
The latter technique has been tremendously used for tissue thickness estimation~\cite{yezzi2003eulerian,acosta2009automated,cedilnik2019fully}, while, surprisingly enough, it has been used only for this particular problem, regardless of its potential to provide accurate geodesic distance maps in a more general setting since it can also handle multiple segment domains with appropriate boundary conditions. In short, the present work describes its potential to estimate the geodesic distance maps related to the mapping of a smooth surface into a sphere. In the following, we will explain in more detail how geodesic distance maps are estimated and then employed to derive our shape descriptor.

\subsection*{\textbf{Heat flow integration:}} 
First of all, let us introduce the heat transfer problem and emphasize the effect of the imposed boundary conditions on the solution. Consider the basic form of the heat equation: 
\begin{equation}
\frac{\partial h(x,t)}{\partial t}= \Delta h(x,t) 
\label{heat_equation}
\end{equation}

Consider the 1-point BVP:
\begin{equation}
\frac{\partial h(x,t)}{\partial t}= \Delta h(x,t) \quad s.t \quad h(\partial \Omega)=1
\label{one_BVP}
\end{equation}
The time-dependant solution of the above stated BVP is used in ~\cite{crane2013geodesics} to approximate the heat diffusion from a single source $\partial \Omega$ on a smooth manifold to all other points of the manifold. In practice, the heat method approximates solution to this particular problem from the heat kernel, namely the Green's function for the heat equation, in the limit, where dissipation time approaches zero (a relation between heat and distance known as Varadhan's formula~\cite{varadhan1967behavior}). For more details we refer to~\cite{wang2015novel} and~\cite{grigoryan2009heat}.

Consider now the 2-point BVP such that the solution we are looking for, satisfies the heat equation inside a region $\Omega = ]\partial_0 \Omega,\partial_1 \Omega[ \subset \mathbb{R}^n$ and takes prescribed values at the boundaries:

\begin{equation}
h(\partial_0 \Omega,t)= a(t), \quad h(\partial_1 \Omega,t)= b(t).
\label{two_BVP}
\end{equation}

Then, according to the Eq.~\eqref{heat_equation}, we expect the temperature distribution $h$ to change with time. However, if $a(t)$ and $b(t)$ are both time-independent (\textit{i.e.} if they are constant over time), then one might expect the solution to eventually reach a steady-state solution after a certain amount of time (\textit{i.e.} limits for t approaching infinity)~\cite{leveque2007finite}. Therefore, $\frac{\partial h} {\partial t} \rightarrow 0$ and the heat equation can be safely reduced to the Laplace equation, $\Delta h = 0$ inside $\Omega$. 

In classical physics, Laplace's equation arises in the description of all kinds of conservative physical systems in equilibrium. In the field of study of Laplace's equation, namely potential theory, a potential is a scalar function whose gradient vector field is divergence- and curl-free. The gradient is then said to be a conservative vector field. Since the principle remains the same if we replace the electric potential with temperature, we propose to determine the temperature distribution under the condition of the thermal equilibrium. 
The solution $h$ is a smooth scalar function whose gradient describes in which direction and at what rate the temperature, or equivalently, the distance increases the most rapidly around a particular location. 
For instance, this solution has been directly employed as an approximation for the scalar distance function on a surface in~\cite{wessner2006anisotropic}. Such an assumption can be well argued for approximating distance maps in-between two parallel plates. However, as soon as one of the two boundaries is slightly deformed, the restrictive unit length condition will be removed from the corresponding gradient vector field. To surmount this issue, and following the principle of the heat method, we will first normalize the gradient vector field of $h$ and then recover the true distance function from the resulting unit-length gradient vector field.

From another point of view, solving Laplace's equation subject to appropriate Dirichlet boundary conditions, is equivalent to solving the variational problem of finding a function $h$ that satisfies the boundary conditions and has minimal Dirichlet energy. \\

\textbf{Definition}
Let $\Omega$ be an open subset of $\mathbb{R}^n$. And let $h: \Omega \rightarrow \mathbb{R}$ be a $C^2$ function over $\Omega$, the gradient’s scalar Dirichlet energy of $h$ is defined by the following positive real number:
\begin{equation}
\label{Dirichlet_energy}
 E[h] = \frac{1}{2} \int_{x\in \Omega} ||\nabla h(x))||^2dx
\end{equation}

where $\nabla h: \Omega \rightarrow \mathbb{R}^n$ denotes the gradient vector field of $h$. Further details, from a Riemannian point of view, are provided in the Appendix~\ref{Appendix A}.

\subsection*{\textbf{Well-posedness of the 2-point boundary value problem:}} 

Before defining the BVP to be solved in this work, we shall give a remark that a solution to Laplace’s equation is uniquely determined if appropriate boundary conditions are posed. More generally, a BVP can be solvable if and only if the problem is well posed. 
In particular, the previously defined Dirichlet problem can be solvable, if and only if the boundaries are smooth curves/surfaces~\cite{krantz1999handbook,greene2006function}. 

To initialize the workflow, the binary mask of the shape is eroded with a cross-shaped structuring element which is best suited for fine structures. The choice of the structuring element is of great importance in preserving, as much as possible, the geometry and topology of any arbitrary shape. The objective here is to deal with sharp peaks particularly. In the following, we denote the eroded mask by $S_{e}$. 
Then, we perform the Principal Component Analysis (PCA) on $S$ (viewed as a point cloud where each voxel is considered as a point). The eigenvector corresponding to the largest eigenvalue gives the axis called the principal axis of Inertia. This axis intersects the shape surface at two points $p_1$ and $p_2$ (called the shape poles). To ensure that the bounding sphere will sufficiently enclose the shape and, thus, to prevent boundaries from overlapping, we use a surrounding sphere $S_s$ of radius $R= 0.8*l$ and whose center coincides with the shape centroid, where $l$ is the usual Euclidean length of the segment $[p_1,p_2]$. To determine the shape/sphere centroid, we use the median point which is a better midpoint measure for cases where a small number of outliers could drastically skew the average.
%\hl{Let us also note that we use the median of shape point coordinates and not the mean as shape centroid.}
At this level, all the shape surface points will be located between two non-intersecting boundaries: $\partial_0 \Omega=\bar{S_s}$, and $\partial_1 \Omega=S_e$, where $\bar{S_s}$ denotes the region outside the sphere.\\
% The full algorithm is described below:
In practice, we are looking for an elliptic twice-differentiable function $h :  \mathbb{R}^3  \rightarrow  \mathbb{R}$ which satisfies the Laplace's PDE $\Delta h = div(grad(h)) = 0 $ inside the region $\Omega = \overline{\partial_0 \Omega \cup \partial_1 \Omega}$, subject to the Dirichlet boundary conditions $h(\partial_1 \Omega)=1$ and $h(\partial_0 \Omega)=0$ (\textit{i.e.} $h$ is the function that has minimal Dirichlet energy for all $x \in \Omega$, also called the harmonic interpolant in potential theory). The physical intuition behind is to determine the equilibrium heat distribution in a perfectly symmetric spherical room since the divergence of the gradient vector field corresponds to some kind of fluid flow. Implicitly, the shape surface $\mathcal{S}$ can be approximated in function of the solution $h$, by the isosurface:
\begin{equation}
\mathcal{S}  =  \{x\in \Omega \quad  | \quad h(x) = c\},
\end{equation}
where $c$ is a constant in $]0,1[$. Varying $c$ continuously from 0 to 1 will give the disjoint isosurfaces of the distance function. It is also important to keep in mind that the major problem in approximating geodesic distance is that the latter fails to be smooth at points in the \textit{cut locus}, \textit{i.e.} points that are equidistant from at least two points on a boundary. Fortunately, however, since the organ surface is smooth enough, then cut locus issues are avoided and the smoothness of the solution would not affect the geometry of the isosurfaces of the distance function.

\subsection*{\textbf{Numerical integration:}}

To approximate the numerical solution of Laplace's equation, we use the Jacobi iterative relaxation method which is simple to implement and which allows fast approximation of the solution in Cartesian coordinates:
\begin{equation}
\begin{matrix} \label{jacob}
h_{t+1}(i,j,k)=\frac{1}{2(\Delta i^2\Delta j^2+\Delta i^2 \Delta k^2+ \Delta j^2 \Delta k^2)}(\Delta j^2 \Delta k^2[h_{t}(i+ \Delta i,j,k)+h_{t}(i- \Delta i,j,k)]\\
+\Delta i^2 \Delta k^2[h_{t}(i,j+\Delta j,k)+h_{t}(i,j- \Delta j,k)]+\Delta i^2 \Delta j^2[h_{t}(i,j,k+ \Delta k)+h_{t}(i,j,k- \Delta k)])
\end{matrix}
\end{equation}
where $t$ is the iteration index. In this work, we process reconstructed data with an isotropic voxel spacing of $1 \times 1\times 1$ mm (\textit{i.e.} $\Delta i = \Delta j = \Delta k = 1$).

\subsection*{\textbf{Convergence criterion:}}

The scalar function $h$ is initialized to 0 inside $\Omega$ and then iteratively relaxed by finite differences until a satisfactory heat steady state solution is reached. A parallel computational implementation is performed in such a way that all grid points inside $\Omega$ are simultaneously visited at each iteration. A convergence criterion can be defined, through a discretization of the eq~\eqref{Dirichlet_energy}, by the following expression based on the total field energy over $\Omega$: 
\begin{equation}
\epsilon_t = \sum_{ \Omega} \sqrt{ (\frac{\nabla h_t^i}{2})^2 + (\frac{\nabla h_t^j}{2})^2 + (\frac{\nabla h_t^k}{2})^2 }
\end{equation}
where $\nabla h_t^i = h_t(i+1,j,k)-h_t(i-1,j,k)$, $\nabla h_t^j = h_t(i,j+1,k)-h_t(i,j-1,k)$, \\ and $\nabla h_t^k = h_t(i,j,k+1)-h_t(i,j,k-1)$. The Jacobi iterative computational scheme of eq~\eqref{jacob} converges when the ratio $\epsilon = \frac{\epsilon_t - \epsilon_{t+1}}{\epsilon_t}$ becomes smaller than a user-defined threshold (typically about $10^{-5}$). Intuitively, this aims at encouraging smooth scalar fields by penalising large spatial gradients between neighbouring grid points. 
%This convergence criterion may provide a plausible link between the Eulerian PDE approach and the heat method for the integration of the heat flow. In fact, the latter method suggests that if the heat has taken a short time to vary/diffuse ($t \rightarrow 0$), then the local distance should also have little time to deviate from the shortest possible path.}
To speed up the algorithm, we strongly recommend to keep the number of iterations as a user defined parameter in order to avoid the repeated evaluation of $\epsilon$ at each iteration. A total number of 200 iterations is used for solving the Laplace's equation in this work. \\
%\hl{In the same context, using the same solver, a smooth solution cannot be achieved in the heat method formulation, imposing a single boundary condition (a single point/curve source). To see more clearly the problem, just imagine the case in which the heat starts spreading from the center, into the rest of the points of a regular grid. It can be immediately perceived that without specifying a smooth external boundary condition, the Jacobi iterative relaxation method, which is based on finite differences, will fail to solve the Laplace equation accurately. In other words, the optimal solution cannot be relaxed (smoothed) more and more near the grid edges (and even near the center) which will affect the whole results.}

\subsection*{\textbf{Computing the unit-length normal vectors to the tangent planes to the harmonic layers:}} As argued before, the gradient of $h$ is highly sensitive to errors in magnitude. We therefore compute the unit normal gradient field $N= - \frac{\nabla h}{\left \| \nabla h \right \|} = (N_i,N_j,N_k)^T$, so that the magnitude can safely be ignored when recovering distance maps from $N$ in the next step. The Eikonal equation will then automatically be satisfied thereafter, without being solved directly. The same operation was referred to as a change of variable in the heat method (the second step)~\cite{crane2017heat}.
Integrating this normal velocity vector field allows for a bijective mapping between pairs of points (\textit{i.e} a one-to-one mapping). This gives a set of paths of minimum Dirichlet energy (\textit{i.e.} a geodesic flow) for mapping the shape to a sphere. Let us underline that $N$ is also a conservative vector field (curl free) when evaluated at each isosurface $h = h_{iso}$. More details are provided in Appendix~\ref{Appendix C}.

\subsection*{\textbf{Recovering geodesic distance maps:}} 

This last step resembles the third and last step in the heat method which consists of computing true distance functions (potential scalar function $L_n$) whose gradient is parallel to $N$. We therefore find the closest scalar potential $G=L_0+L_1$ by minimizing $\sum_{n=0}^1 \int_{\Omega}\left | \nabla L_n - N \right |^2$, or equivalently, by solving the elliptic Poisson equation $\nabla \cdot \nabla L = \nabla \cdot N$, where $(\nabla \cdot)$ is the divergence operator, or even more simply, by aligning $\nabla L$ and $N$.

In the discrete setting, a solution to the above stated problem can be determined by solving a couple of Euler-Lagrange PDEs: $\nabla L_0 . N = -\nabla L_1 . N = 1$, subject to the boundary conditions $L_0(\partial_0 \Omega)= L_1(\partial_1 \Omega) = 0$. $L_0$ and $L_1$ are first initialized to $0.5$ and then iteratively updated inside $\Omega$ using a symmetric relaxation Gauss-Seidel method:
\begin{equation}\label{L0}
\begin{matrix}
\frac{L_0^{t+1}[i,j,k]}{\alpha} = 1+\left | N_i \right |L_0^t[i\mp1,j,k]+\left | N_j \right |L_0^t[i,j\mp1,k]+\left | N_k \right |L_0^t[i,j,k\mp1] 
\end{matrix}
\end{equation}
\begin{equation}\label{L1}
\begin{matrix}
\frac{L_1^{t+1}[i,j,k]}{\alpha} = 1+\left | N_i \right |L_1^t[i\pm1,j,k]+\left | N_j \right |L_1^t[i,j\pm1,k]+\left | N_k \right |L_1^t[i,j,k\pm1]
\end{matrix}
\end{equation}

where: $\left\{
\begin{array}{l}
  m \pm 1 = m + sgn(N_m), \quad  m \mp 1 = m - sgn(N_m)  \\
  for \quad m \in{\{i,j,k\}}, \quad and \quad  \alpha = \frac{1}{\left | N_i \right |+\left | N_j \right |+\left | N_k \right |}  \\
\end{array} \right.$,

with: $ sgn(.) $ is the sign function, $L_0(x)$ is the length of the optimal geodesic path from the point $x=(i,j,k)$ to $\partial_0 \Omega$, while $L_1(x)$ is the length of the optimal geodesic path from $x$ to $\partial_1 \Omega$. 
The sum of these two lengths $G(x) = L_0(x)+L_1(x)$, defined as \textit{thickness} in~\cite{yezzi2003eulerian}, represents in fact the length of the optimal geodesic path from $\partial_0 \Omega$ to $\partial_1 \Omega$ that passes through $x$. These geometric methods which are fully non-parametric allow one to handle time derivatives with finite differences in a restricted physical domain that exhibit large deformations.This step is also parallelized by simultaneously visiting all voxels inside $\Omega$ at each iteration.

\subsection*{\textbf{Proposed feature for surface characterization:}} Finally, we define a flexible feature by the following application: 
   % \begin{equation}
    %    \tilde{f}(x) = \frac{R}{G(x)}
    %\end{equation}
    
\begin{equation}
\begin{matrix}
\tilde{f} :  \Omega  \rightarrow  \mathbb{R}^*_+ \\
x \mapsto \frac{R}{G(x)}
\end{matrix}
\end{equation}

\begin{figure}[ht]
\centering
\includegraphics[width=0.9\linewidth]{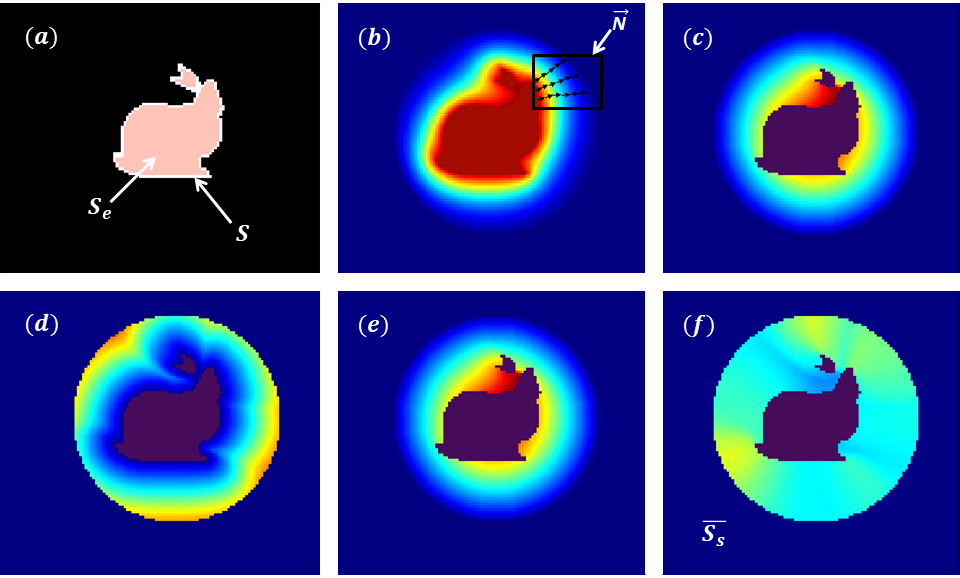}
\vspace{-0.2cm}
\caption{Proposed method (Stanford Bunny example): (a) binary mask and its eroded version, (b) harmonic map: heat values range from  blue (0) to red ($10^4$), (c) length of the geodesic path from $x$ to the surrounding sphere $L_0$, (d) length of the geodesic path from $x$ to the inner boundary $L_1$, (e) length of the resulting geodesic path $G$ (lengths are expressed in mm), and (f) feature values $\tilde{f}$ between the two boundaries (in mm$^{-1}$).}
\label{fig:proposed_method}
\end{figure}

\begin{figure}[h!]%[!h]
\centering
\includegraphics[width=0.9\linewidth]{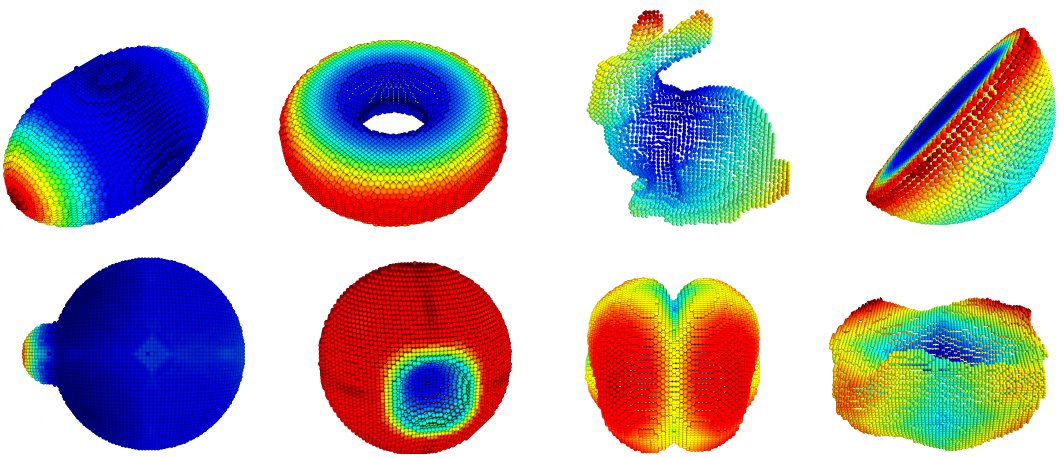}
% \vspace{-0.3cm}
\caption{\label{ellipsoid_curv} Obtained feature maps (normalized) for synthetic and realistic (\textit{e.g.} neonatal brain and bladder) surfaces. Colormap goes from blue (0) to red (1).}
\end{figure}

Note that $L_0 >> L_1$ for all the surface points so that $G \simeq L_0$ and only one PDE has to be solved to calculate the feature function (see Fig.~\ref{fig:proposed_method}\textcolor{blue}{.c},~\ref{fig:proposed_method}\textcolor{blue}{.e}). The function $\tilde{f}$ has the potential to characterize the surface variation and to delineate between concave, convex and flat regions as illustrated in Fig~\ref{ellipsoid_curv} for both torus and ellipsoid. Relatively, the largest feature values correspond to the most convex areas in the surface while the smallest values correspond to the most concave areas. The proposed feature values exhibit smooth transitions between concave and convex regions. Furthermore, this feature is scale-invariant thanks to the use of variable sphere radius, rotation-invariant thanks to spherical symmetry, and also invariant to translation since the sphere and shape are concentric. 
An illustration of all the previous steps is presented in Fig.~\ref{fig:proposed_method}. Fig~\ref{ellipsoid_curv} illustrates the obtained feature maps for a set of symmetric and non-symmetric geometries. 

% \begin{figure}[h!]
% \centering
% \subfigure{\includegraphics[scale=.03]{fm1.jpg}}
% \hspace{0.6cm}
% \subfigure{\includegraphics[scale=.027]{fm2.jpg}}
% \hspace{0.6cm}
% \subfigure{\includegraphics[scale=.13]{fm3.png}}
% \hspace{0.6cm}
% \subfigure{\includegraphics[scale=.27]{fm4.png}}

% \subfigure{\includegraphics[scale=.027]{fm5.png}}
% \hspace{0.6cm}
% \subfigure{\includegraphics[scale=.023]{fm6.jpg}}
% \hspace{0.6cm}
% \subfigure{\includegraphics[scale=.027]{fm7.png}}
% \hspace{0.6cm}
% \subfigure{\includegraphics[scale=.025]{fm8.png}}

% \caption{\label{ellipsoid_curv} Obtained feature maps \hl{(normalized)} for synthetic and realistic (\textit{e.g.} neonatal brain and bladder) surfaces. \hl{Colormap goes from blue (0) to red (1).}}
% \end{figure}

\begin{figure}[h!]
\begin{center}
\centering
\includegraphics[width=.3\linewidth]{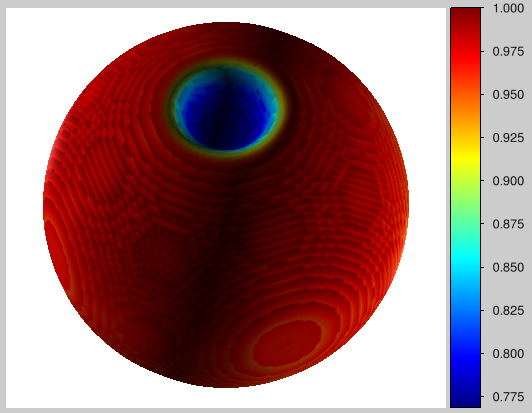}
\hspace{2cm}
\centering
\includegraphics[width=.3\linewidth]{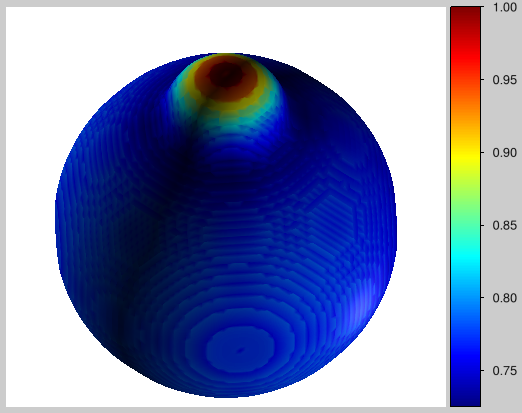}
\hspace{2cm}
\centering
\includegraphics[width=.3\linewidth]{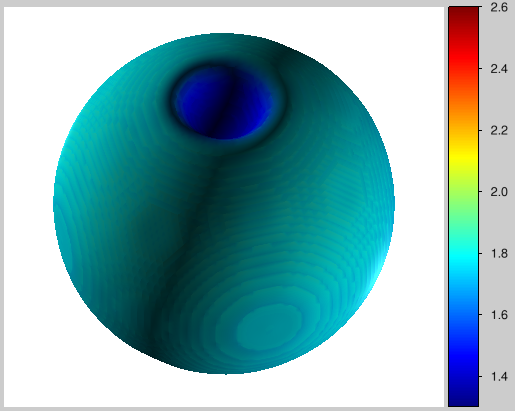}
\hspace{2cm}
\centering
\includegraphics[width=.3\linewidth]{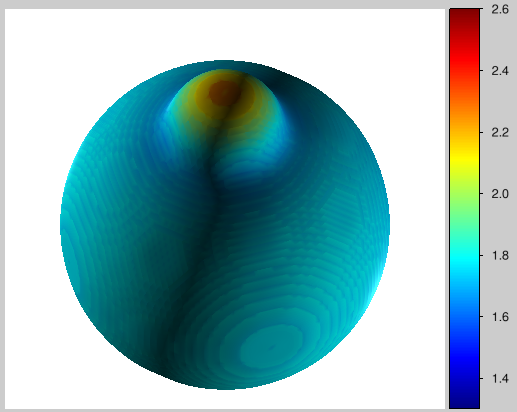}

\caption{Capacity to detect local changes in surface geometry: first row (after normalization), second row (without normalization and with a commun truncated colorbar). Feature maps are plotted on the extracted isosurface $h=0.98$.}
\label{fig:local_malformation}
\end{center}
\end{figure}

The ability to quantify organ shape changes with respect to the sphere would not only allow for deeper understanding of organ motion but also conceivably allow for improved detection of pathologies. In practice, many pathologies are associated with localized organ malformations. Fig.~\ref{fig:local_malformation} provides evidence that the proposed feature is capable of discriminating such phenomena. Moreover, assuming that the organ volume should be preserved during motion, we suppose that the sphere radius should remain constant which is a sufficient condition to detect all local changes without any small scale effect. Therefore, the feature values, being inversely proportional to a distance measure, can be expressed in mm$^{-1}$.
\section{Experiments and Results}

\subsection{Data set} \label{sec:data_set}
\subsubsection{Realistic dynamic MRI data}

Pelvis areas of seven healthy participants (five women), ranging in age from 23 to 31 years, and in weight from 58 to 81, were imaged. Subjects were imaged with a 1.5T MRI scanner (MAGNETOM Avanto, Siemens AG, Healthcare Sector, Erlangen, Germany) using a spine/phased array coil combination and $T_{1}/T_{2}$ weighted balanced steady-state free precession sequences ($T_{1}/T_{2}W$ bSSFP). 
A quasi-isotropic 3D static image was recorded during a maximum expiration apnea of 18 seconds. Multi-planar dynamic acquisition assuring full coverage of the pelvic region was recorded during a $80$ s forced breathing exercise. During this exercise, the subject alternately inspired and expired at maximum capacity. 
Subjects were also instructed to increase pelvic pressure at the maximum inspiration and then to contract abdominal muscles during the expiration. These actions increased the intra-abdominal pressure, causing deformations of the pelvic organs. 
Spatial configuration of the multi-planar acquisition is described in detail in~\cite{ogier20193d}. The study was approved by the local human research committee and was conducted in conformity with the Declaration of Helsinki. Since no extraneous liquids was injected into pelvic cavities in this study, only the segmentation of the bladder was performed and the analysis focused exclusively on this organ.
For each subject, the three-dimensional dynamic sequences acquired in multi-planar configurations allowed the reconstruction of nearly 400 bladder volumes generated at a rate of 8 volumes per second. Bladder volume of each subject was computed from the manual segmentation of the static acquisition. Bladder volume presented a large variability among subjects (values were ranged from 48 cm$^3$ to 403 cm$^3$).
Since the scanning duration is short relative to the organ motion, spatio-temporal reconstructions were performed outside the MR scanner to recover the missing data using diffeomorphic registration as detailed in~\cite{ogier20193d}.

\subsubsection{Data simulation}
\label{sec:simu_patho}

To evaluate the capability of each geometric descriptor in detecting abnormalities in bladder dynamics during breathing exercises, we have simulated a smooth continuous-time trajectory of the organ volume using a Log Euclidean Polyaffine registration framework~\cite{arsigny2009fast} to excite the organ deformation from the interior with locally controlled properties. The organ volume in the static scan is divided into four non-intersecting regions and an affine transformation is associated to each region or component (we have only used component-wise scaling transforms in such a way that the organ volume is still preserved throughout the sequence). A first motion cycle is simulated by estimating a flow of invertible diffeomorphisms for mapping the organ volume from resting state (i.e. organ volume in the stationary scan) towards the maximum of inspiration state exhibiting the largest deformation. A forward trajectory was estimated based on the integration of stationary velocity fields via the \textit{exponential map}. Then, the inverse trajectory for coming back to the resting state was obtained by smoothly interpolating the inverse of the simulated polyaffine transformations. Finally, this motion cycle is repeated 8 times for stability assessment of different measures.

\begin{figure}[h!]
 \centering
  \includegraphics[width=0.9\linewidth]{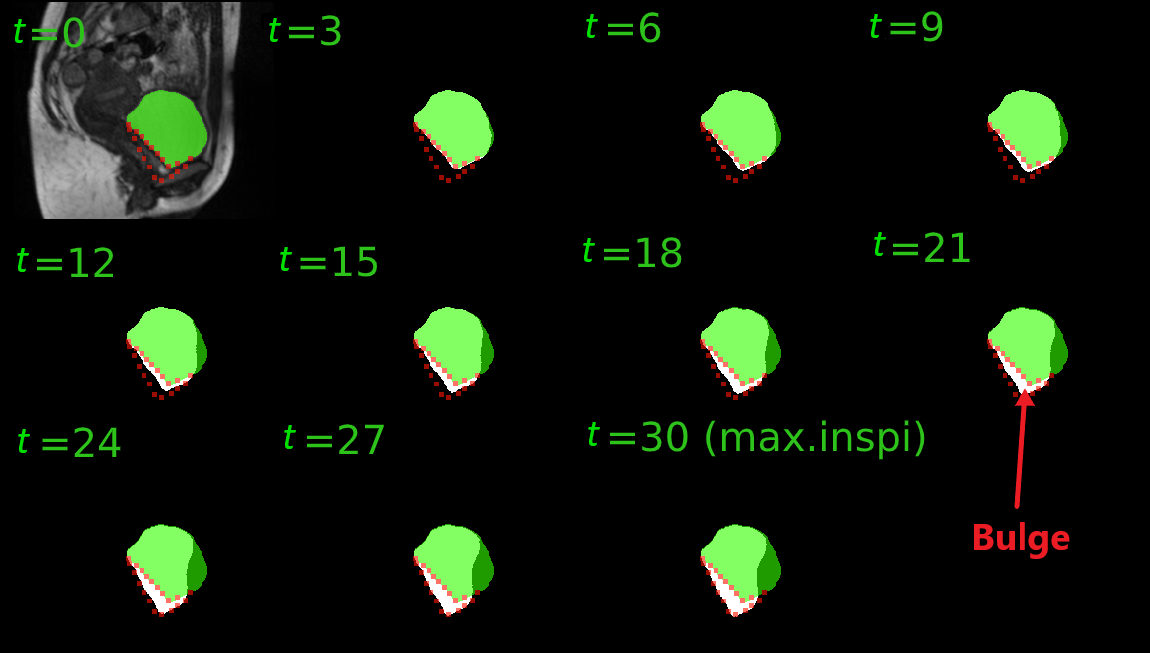}
  \caption{Simulation of large nonlinear deformations using the LEPF: the green mask corresponds to the bladder segmentation at the resting state, the white segmentations correspond to the temporal organ positions, while the red dotted area shows the regional bladder motion abnormality (the resulting bulge through the vaginal canal).}
  \label{simulated_deformations}
\end{figure}

In practice, we simulated a large deformation to make the organ fall down by maximum of inspiration which allow the floor of the bladder to sag through the muscle and ligament layers. This abnormal kind of motion occurs frequently in women with uterine and bladder prolapse for which the bladder can create a bulge into the vagina because of the weakness of their pelvic muscles and ligaments. Fig.~\ref{simulated_deformations} illustrates the simulated organ trajectory which we will call Simulated Pathology (SP) in the sequel. This sequence, for which the shape trajectory is well known, will serve to compare the different features used in this study in terms of stability and measure repeatability during forward organ movement. It will also serve to assess their capacity to detect this abnormal motion type in a mixed data set.

\subsection{Tracking of mesh vertices}
Diffeomorphic registrations were performed using the software package Deformetrica~\cite{deformetrica}. The parameters of the registration algorithm were set as follows: the standard deviation of the Gaussian kernel defined in~\eqref{gauss_kernel} was set to $\sigma = 10^{-4}$ in order to obtain a deformation field with a narrow precision. The kernel-width parameter for controlling the granularity of the deformation was set to $8$. $15$ intermediate states describing the temporal evolution of the tracked points were estimated in order to obtain a smooth continuous-time trajectory of the organ between successive time frames. The loss function defined in~\eqref{loss_function} was minimized using the gradient descent optimization method. All experiments were performed on an Intel$^{\mbox{\scriptsize{\textregistered}}}$  Xeon$^{\mbox{\scriptsize{\textregistered}}}$ Processor Silver 4214 CPU \verb+@+ 2.20GHz, with a physical memory of 96GB. For the first subject for example, Algorithm~\ref{algomotion} took $6$ sec to align a set of $342$ tracked point set $\mathcal{M}_t$, with the target set $\mathcal{C}_t$, composed of $5686$ points for which only a set of $210$ control points have been used for optimizing the shape matching.

\subsection{Geometric descriptors}
\label{sec:geom_desc}
In this study, we have used different shape descriptors that can be classified into two categories:
\begin{itemize}
\item 
Biomechanical descriptors: the deformations were quantified using mesh temporal elongations and distortions. This aimed to extend the methodologies used in ~\cite{rahim2013diffeomorphic} from 2D+$t$ to 3D+$t$.

\item 
Geometric descriptors (family of geodesic-based features): the mean curvature and the new proposed descriptor which take into account the non Euclidean geometry of three dimensional shape space.
\end{itemize}

% Figure~\ref{fig:bladder_defo} illustrates the capability of each feature to provide a specific information about organ temporal deformation. Both mean curvature and our feature provide information about surfacic tangential deformations.
% The dihedral angle based distortion may capture mesh sharp deformations, while the mesh elongation aims to capture shape dilations and contractions during motion. 
A comparison between descriptors was performed where the goal was to assess their robustness by
performing trajectory stability and sensitivity analysis.
% \hl{stability and sensitivity analysis}. 
% and to evaluate their performances in terms of:
% \begin{itemize}
% \item 
% Sensitivity to tracking error propagation, related to possible registration biases (\textit{e.g.} small errors which may be caused by the smoothing of the velocity vector field within the LDDMM framework.)

% \item 
% Numerical stability throughout a perfectly cyclic nonlinear deformations in a priori given shape trajectory (\textit{i.e.} using the simulated sequence).
% \end{itemize}

% To evaluate the numerical stability of each descriptor, we focus on the similarity between the corresponding feature map at the last revisited resting state (towards the end of sequence, $t=420$), and the reference feature map (at t=0). Figure~\ref{fig:temporal_stability2} shows that the proposed descriptor is the more stable since the associated correlation curve reached its maximum value at each revisited resting state. In fact, the proposed feature exhibited a perfect curve showing a repeated maximum of correlation very close to 1. 

% When the bladder falls down by maximum of inspiration, the application of the proposed feature shows that the organ topology is most deformed laterally.

\begin{figure*}[!h]
\centering
\subfigure[Proposed feature (in mm$^{-1}$)]
{\includegraphics[scale=.39]{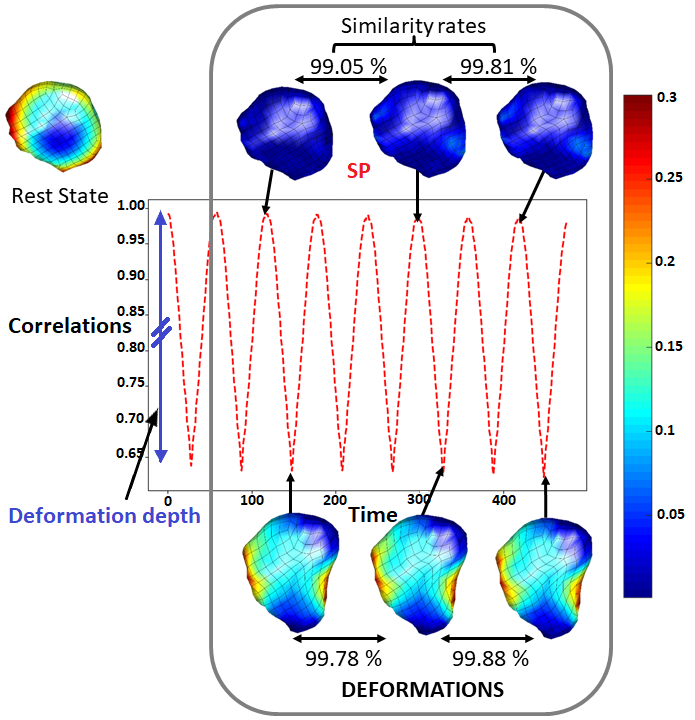}}
\subfigure[Mean curvature (in mm$^{-1}$)]
{\includegraphics[scale=.39]{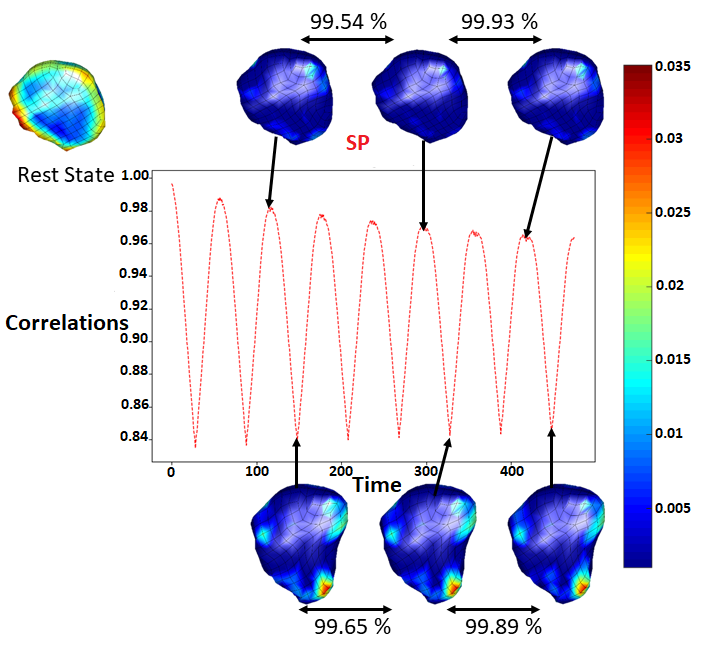}}
\subfigure[Mesh elongations (in mm)]
{\includegraphics[scale=.39]{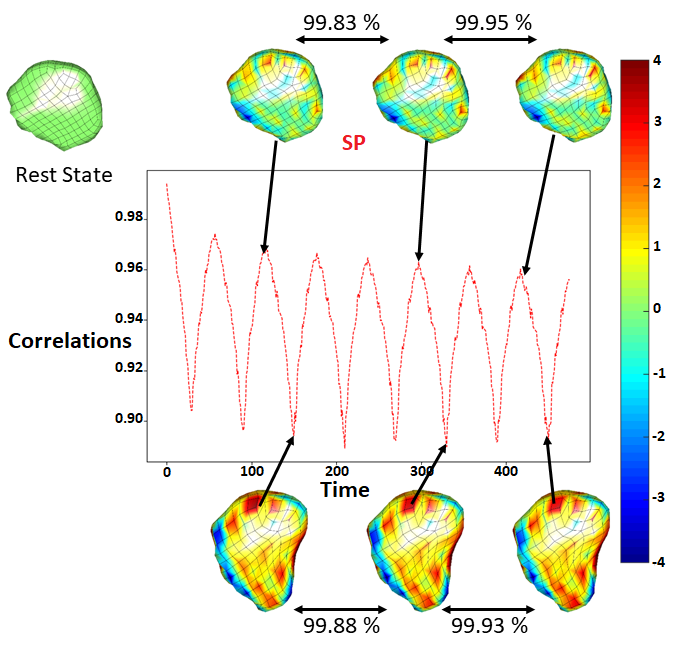}}
\subfigure[Mesh distortions (in degrees)]
{\includegraphics[scale=.39]{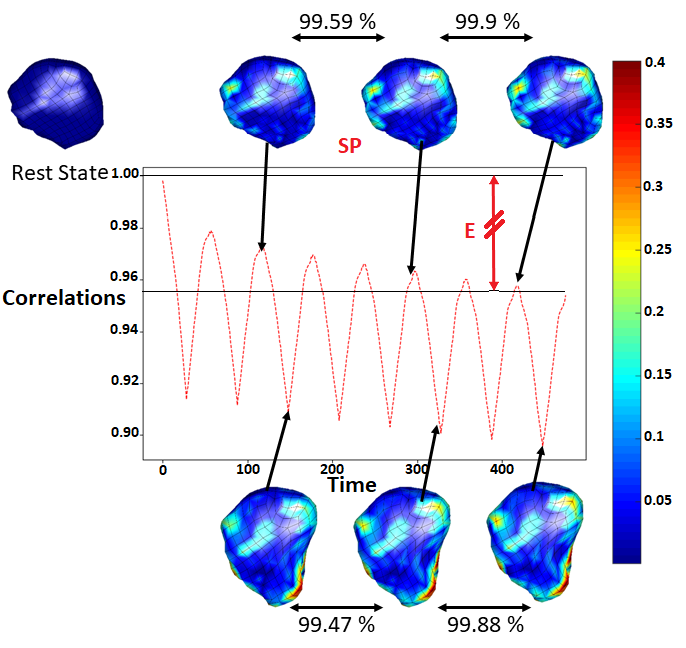}}
    \caption{Feature map dynamics (changes) w.r.t the resting state for the synthetic sequence. Elongations and distortions showed the largest sensitivity rate to tracking-error propagation: $E = 4.5 \% $. Results show up to $99 \%$ correlation (similarity rate) between feature maps at first and last (9th) resting states for our descriptor, more than $3 \%$ better than those of other descriptors. For (a), (b) and (d), colorbars range from blue (no deformation) to red (high deformation). For (c), the green color indicates that no deformation has occured.}
    \label{fig:temporal_stability2}
\end{figure*}

% \hl{Figure}~\ref{fig:temporal_stability_realistic} \hl{shows also that the geodesic-based features are better suited as classifiers as they are numerically much more stable than mesh elongations and distortions for characterizing realistic organ trajectories.}
In fact, the propagated tracking-errors may affect the mesh regularity in the neighbourhood of some slightly perturbed vertices. Such small errors can be from any source. For instance, they may have originated from velocity smoothing within the LDDMM, which can also be mixed with round-off errors propagated in the numerical integration of differential equations when computing the features themselves.
Assuming that the effect of error propagation increases with time, neighbourhood-based features may thus be more or less sensitive to these cumulative errors.
To evaluate the robustness of each descriptor with respect to error propagation, we computed the following error ratio that is indicated in Fig.~\ref{fig:temporal_stability2}:
\begin{equation}
E= 1 - normcorr{(\mathcal{F}_{r}(0), \mathcal{F}_r(t_{max})})
\end{equation}
where $normcorr(., .)$ is the normalized correlation function, $\mathcal{F}_{r}(0)$ is the reference feature map, and $\mathcal{F}_{r}(t_{max})$ is the obtained feature map at the last time the resting state is revisited.\\
Error estimates for the characterization of simulated organ shape trajectory were: $0.045$ for mesh elongations and distortions, $0.034$ for Riemannian mean curvature, and $3.10^{-6}$ for the proposed feature.

Fig.~\ref{fig:temporal_stability2} shows different feature map dynamics across time for the simulated sequence. The top meshes represent the shape at each revisited resting state. Using the geodesic-based feature for which the computations were totally independent of vertex neighbourhood characteristics, the feature map differences were very close to zero over all the surface but we can observe the presence of small feature changes arising around some vertices, caused mainly by propagation of neglectable registration errors. This means that this feature is capable of detecting these perturbations as well as providing correct correspondence trajectories in a bijective fashion. Moreover, the use of the proposed feature exhibited a perfect correlation curve repeatedly showing a correlation value which is very close to 1 at each revisited resting state.

To evaluate descriptors in terms of their capacity to detect motion abnormality, we have defined the deformation depth parameter by $Depth = 1 - min corr$, where $min corr$ is the minimal correlation value achieved at maximum of inspiration (see Fig.~\ref{fig:temporal_stability2}). The obtained deformation depths were: $0.12$ for mesh elongations, $0.115$ for mesh distortions, $0.17$ for Riemann mean curvature, and $0.36$ for the proposed feature. 
Since we have simulated a large deformation in a specific direction (see Section \ref{sec:simu_patho}), it would be expected that the corresponding correlation trajectory will exhibit an important deformation depth. Which was the case using our descriptor, in a more remarkable way.\\

On the other hand, sensitivity assessment consisted of testing the robustness of feature-based statistical characterizations in the presence of some uncertainties related to the data acquisitions. For example, respiratory depth and rythm may vary slightly over time and cannot be totally controlled during MR scanning. Therefore, we cannot assure that the resting state will reappear somewhere else in the sequence while preserving exactly the same initial organ shape patterns. To compare between descriptors in this context, we used a realistic organ trajectory, for which explicit error models are not known.

% we interpret these results to indicate that the proposed feature is more capable of describing how the maximal deformation through motion is large \hl{in addition to its capacity to provide a deformation depth that remains reasonably
% stable over long trajectories.}

\begin{figure}[!h]
\centering
\subfigure[Proposed feature (in mm$^{-1}$)]{\includegraphics[scale=.39]{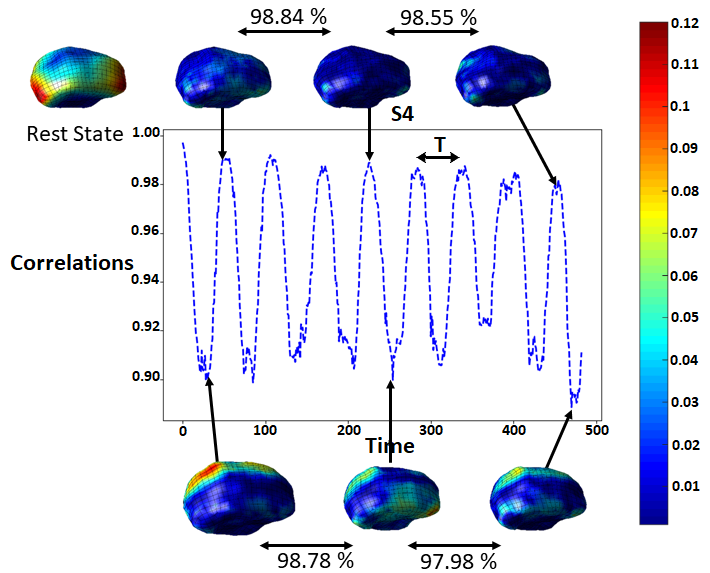}}
\hspace{1cm}
\subfigure[Mean curvature (in mm$^{-1}$)]{\includegraphics[scale=.39]{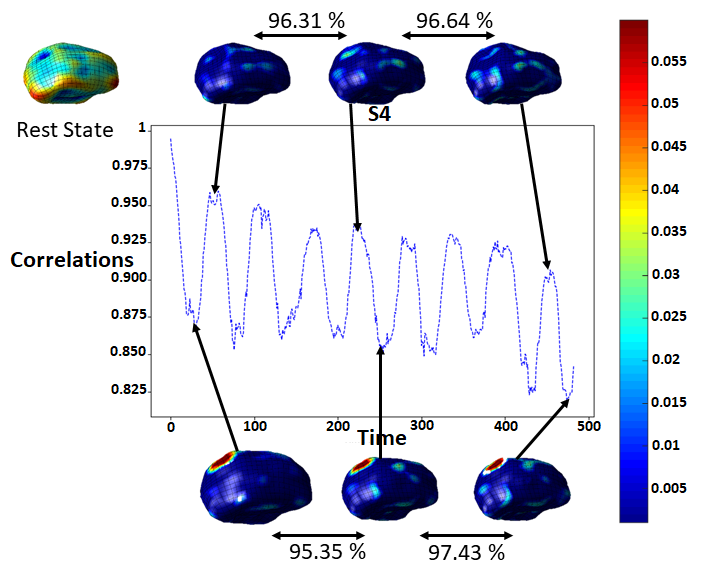}}
\subfigure[Mesh elongations (in mm)]{\includegraphics[scale=.39]{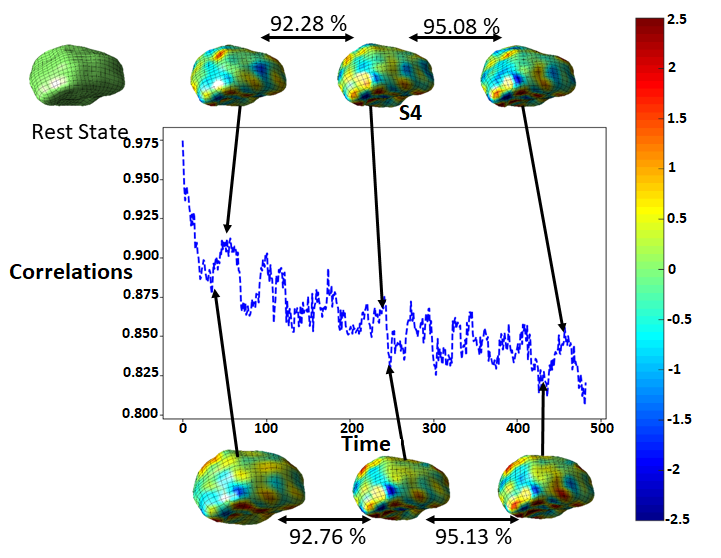}}
\hspace{1cm}
\subfigure[Mesh distortions (in degrees)]{\includegraphics[scale=.39]{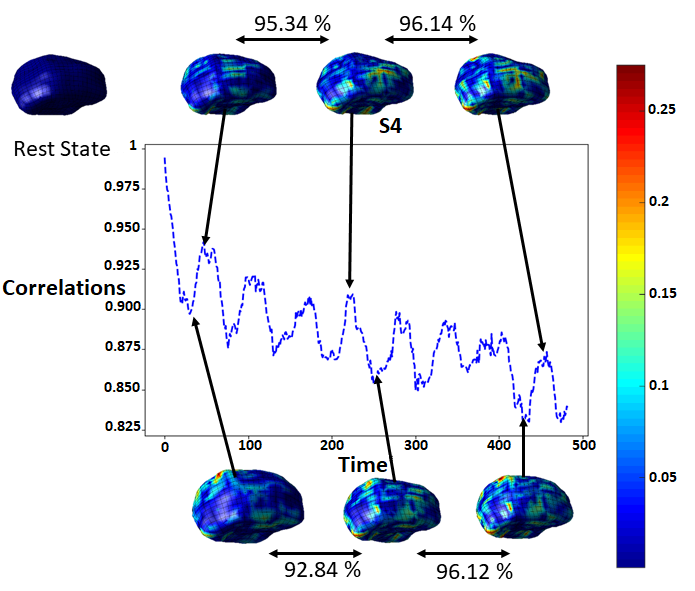}}
\caption{Feature map dynamics (changes) w.r.t the resting state for one realistic organ shape trajectory. As shown in (a) and (b), a fine characterization of organ shape trajectory using geodesic-based features can allow for approximating the respiratory frequency, $f_r \approx \frac{1}{T}$, where $T$ is the median time interval between two successive maximums of the correlation curve. For (a), (b), and (d), colorbars range from blue (no deformation) to red (high deformation). For (c), the green color indicates that no deformation has occured.}
    \label{fig:temporal_stability_realistic}
\end{figure}

Fig.~\ref{fig:temporal_stability_realistic} shows that the correlation curve associated to the proposed feature is characterized by greater spatio-temporal stability than those produced by employing each of the other descriptors for characterizing one realistic organ shape trajectory. These experiments show also that mesh elongations are very sensitive to noise and tracking-error propagation which makes it difficult to quantify breathing frequency. It is also clear that the stability of distortion measures decreases significantly with time. This confirms the assumption mentioned above and also confirms the fact that the geodesic-based features are more capable of describing respiratory frequency and depth.
% More results for comparing between features in terms of their classification power are given in Figure}~\ref{fig:intersub_distMat}. 

% \begin{figure}[ht] 
% \centering
% \includegraphics[scale=.37]{Figs/mapping_four_features_AOL.png}
% \caption{\label{fig:bladder_defo} Reference feature maps (left column) and motion patterns or feature differences with respect to the reference at extreme positions (right columns) for one subject.}
% \end{figure}

\subsection{Subject-specific organ dynamics}
\label{sec:subj_specific_charac}
\begin{figure}[ht]
\centering
\subfigure{\includegraphics[scale=.31]{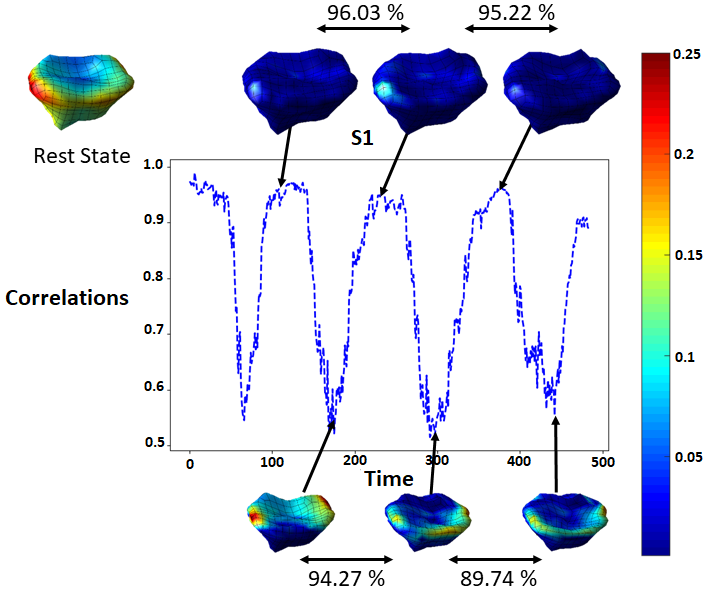}}
\hspace{0.7cm}
\subfigure{\includegraphics[scale=.32]{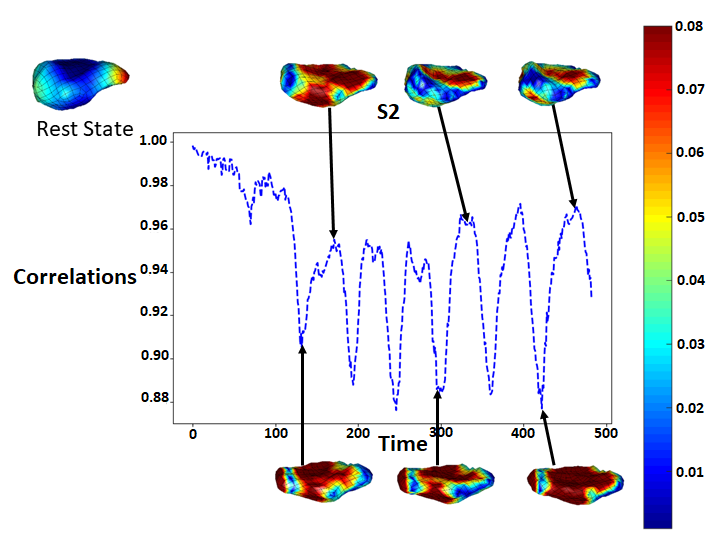}}
\vspace{0.4cm}

\subfigure{\includegraphics[scale=.31]{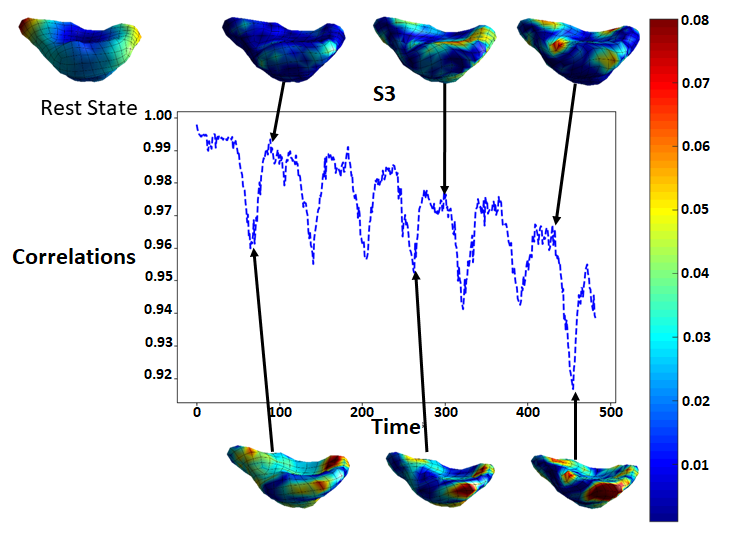}}
\hspace{0.7cm}
\subfigure{\includegraphics[scale=.31]{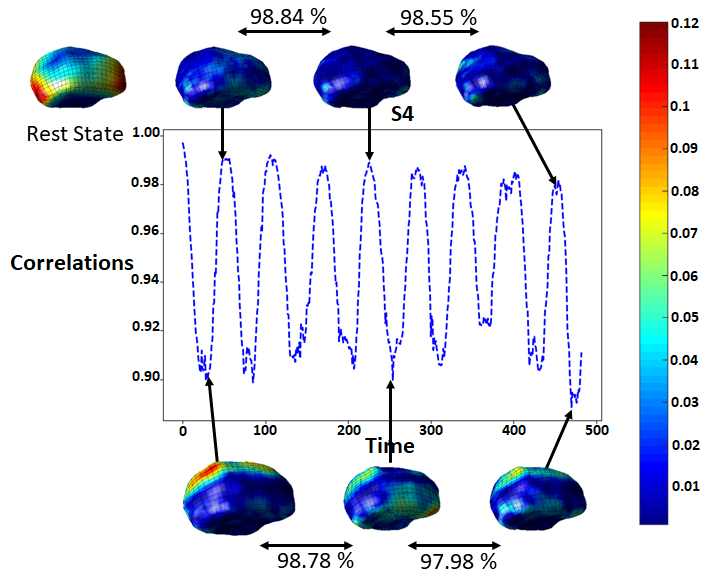}}
\vspace{0.4cm}

\subfigure{\includegraphics[scale=.31]{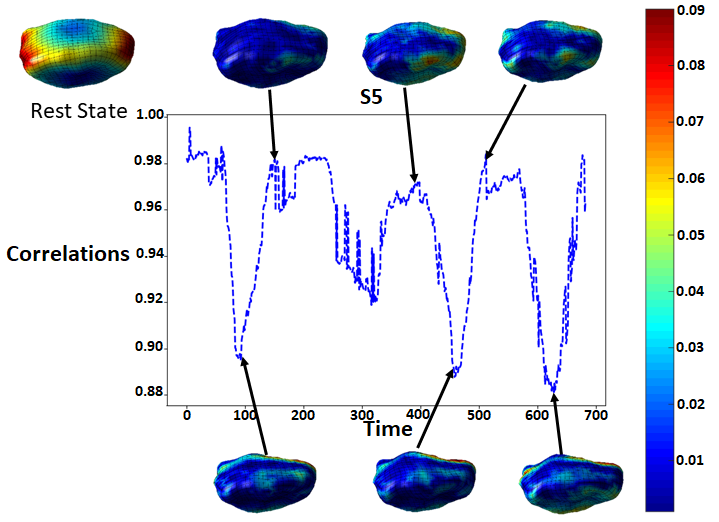}}
\hspace{0.7cm}
\subfigure{\includegraphics[scale=.31]{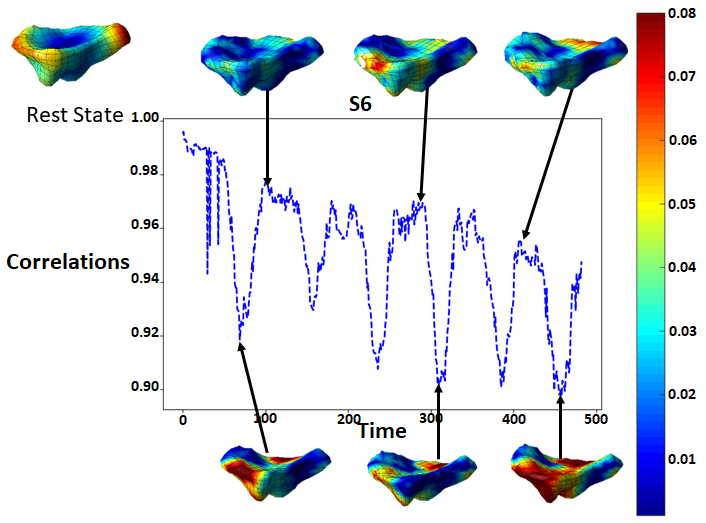}}
\vspace{0.3cm}

\subfigure{\includegraphics[scale=.31]{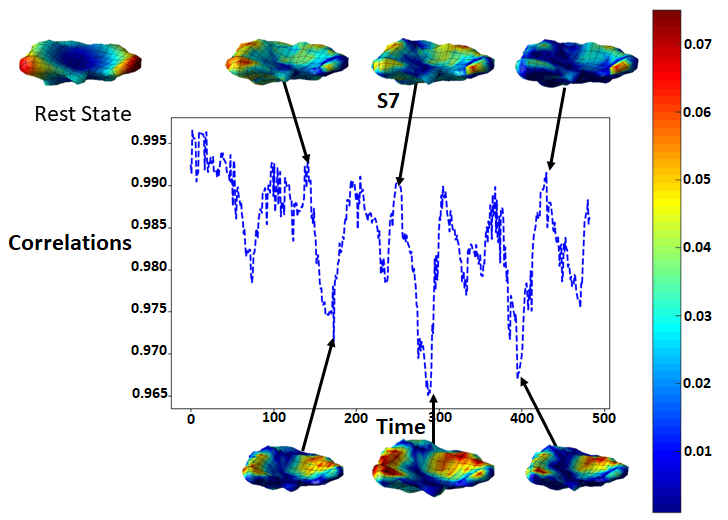}}
\hspace{-0.7cm}
\caption{Characterization of the different loading phases of the bladder using the proposed feature. Each subfigure corresponds to one subject. For each subject, we show the cyclical behavior in surface motion patterns: at maximum of expiration states (top meshes), and at maximum of inspiration states (buttom meshes). Colormaps depict feature changes in mm$^{-1}$ and range from blue (no deformation) to red (high deformation).}.
\label{fig:subjects_corr_maps}
\end{figure}

%\hl{To quantify organ surface deformations across time from feature maps, we use the Pearson correlation coefficient as metric, thus taking advantage of its invariance to scale and shifting.}
Results presented in Fig~\ref{fig:subjects_corr_maps} illustrate the organ motion patterns for all subjects using the proposed descriptor. For each motion cycle, a decrease in correlation coefficients reflects changes in surface geometry, relative to the reference state. This can be interpreted by the fact that forced inspiration involves an action of the diaphragm and abdominal muscles which induces deformation of the internal organs. Otherwise, when the correlation values increase, the patient releases the pressure and consequently the bladder relaxes and returns to its initial shape.
Respiratory motions were expected to be perfectly regular during scanning. However, depth and rhythm may vary over time. The patient's breathing patterns then becomes irregular with time which can be noticed from the correlation trajectories.
For all healthy subjects, the organ was highly deformed by maximum of inspiration and the deformations occured essentially in the top lateral regions. In terms of motion patterns, the organ shape at the maximum of expiration state was very close to the shape at resting state, especially for subjects $S_1$, $S_4$, and $S_5$ (see top meshes in each subfigure). Clearly, the resting state has never been revisited throughout the organ shape trajectory for the other subjects. This can explain the reduction in correlation across cycles.

% old caption : Feature map dynamics w.r.t the resting state for the synthetic sequence, using, clockwise from top-left: our feature (in mm$^{-1}$); mean curvature (in a.u.); mesh distortions (in degrees); and mesh elongations (in mm). Mesh elongations and distortions showed the largest sensitivity rate to tracking-error propagation: $E = 4.5 \% $. Results show up to $99 \%$ correlation (similarity rate) between feature maps at first and last (9th) resting states for our descriptor, more than $3 \%$ better than those of other descriptors.

% old caption : Feature map dynamics w.r.t the resting state for one realistic organ shape trajectory, using, clockwise from top-left: the proposed feature; the mean curvature; mesh distortions; and mesh elongations. A fine characterization of organ shape trajectory using geodesic-based features can allow for approximating the respiratory frequency, $f_r \approx \frac{1}{T}$, where $T$ is the median time interval between two successive maximums of the correlation curve.

%%... \hl{Karim inteprete cette partie}

\subsection{Inter-subject variability in motion patterns}

In addition to the organ shape inter-variability, depth and rhythm of respiration may vary across subjects. Depending on individual lung functions and capacities, healthy individuals exhibit different motion patterns and thus different bladder volume changes. In this work, inter-subject comparisons were performed in two ways:
i) a global metric is achieved using the organ deformation depth which reflects the respiratory depth (section~\ref{sec_mdbc}). And ii) a more fine comparison is performed by establishing point-to-point anatomical correspondences to compute inter-correlations by means of topological characteristics (section~\ref{sec_iscutsp}).

\subsubsection{Maximal-deformation based comparison}
\label{sec_mdbc}

\begin{figure*}[!h]
\centering
\includegraphics[scale=.6]{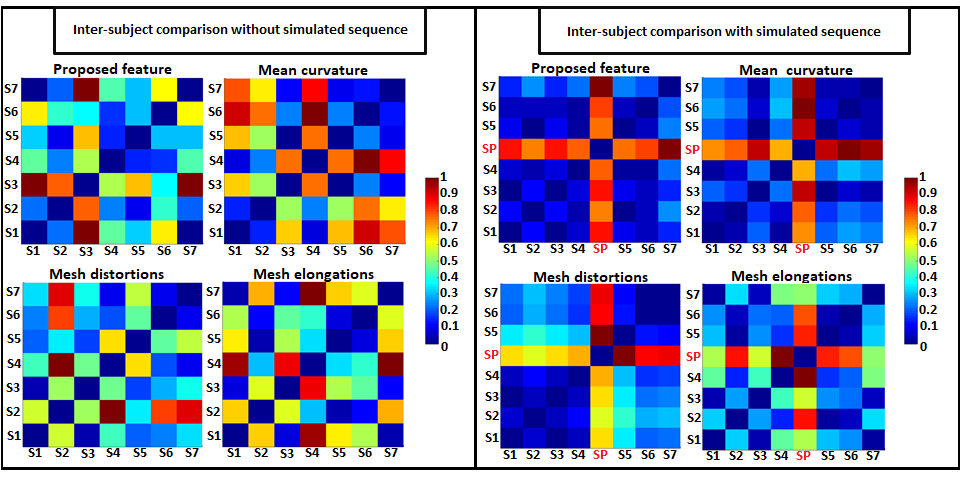}
\caption{\label{fig:intersub_distMat} Inter-subject comparison: normalized distance matrices between maximal temporal deformations (i.e. deformation depths). The left panel depicts the results for the volunteer-only comparisons. The right panel shows the comparisons after introducing the artificial sequence into the mix.}
% From left to right, from up to down: using the proposed feature, using the mean curvature, using mesh distortions, and using mesh elongations.}
\end{figure*}

Fig~\ref{fig:intersub_distMat} depicts the normalized distance matrices based on the absolute differences of deformation depths between each pair of subjects. We have introduced the simulated sequence into the data set for comparing between two different classes of motion, \textit{i.e.} normal and abnormal. 
The objective of these experiments is then two-fold: 1) to show how deformation depths vary across healthy subjects, depending on their individual breathing capacities (left panel), 2) to evaluate the capacity of each descriptor to filter out the introduced artificial data from the mix, and thus to recognize any kind of abnormal motion, such as the pathological motion type described in Section~\ref{sec:simu_patho} (right panel).
% The maximal deformation is identified here by the minimum correlation value between the temporal feature vectors w.r.t the reference one, achieved throughout the dynamic sequence, which reflects the maximum of inspiration time where the bladder is deformed the most.
As illustrated in the figure, the artificial sequence was poorly correlated with the realistic ones in terms of deformation depth. This difference was significant when using  the proposed feature and the Riemannian curvature which makes them better suited as classifiers, but less significant when using mesh distortions and elongations for the characterization of organ dynamics. These results are in line with those obtained in previous sections~\ref{sec:geom_desc} and~ \ref{sec:subj_specific_charac}.

% \subsubsection{Distance matrix-based method}

\subsubsection{Inter-subject comparison using the spherical parameterization}

\label{sec_iscutsp}

Because of the large inter-variability between subjects in terms of organ volume and geometry, the organ surface parameterization differed across subjects (see Section~\ref{surface_param}). We then used the eigenfunctions of the LBO to find point correspondence between two vertex sets in order to compare between different feature-based characterizations.
Each subject $X$, for which the organ surface was encoded with $n$ samples, was compared with each other subject $Y$ in the database with a unique parameterization (that of $X$). We therefore adopted the spherical parameterization method introduced in \cite{lefevre2015spherical}. This method considers only the three first non-trivial eigenfunctions of the LBO of the closed (genus$-0$) surfaces. However, it is not guaranteed that the three first eigenfunctions possess only two nodal domains which may lead to some irregular situations such as singular vertices in sphere-mesh. To deal with this problem, we followed~\cite{bohi2019global} by taking into account the six first eigenfunctions and then selecting among them the first three eigenfunctions with only two nodal domains. Further details of the methods are provided in the Appendix~\ref{Appendix B}.
% in which the velocity vector field on the tangent plane vanishes.

For the proposed geodesic-based feature, the spherical mapping was straightforward since we have the geometric flow that maps the mesh vertices to their corresponding vertices on a sphere surface (see Appendices~\ref{Appendix A} and~\ref{Appendix B} for more details), and only the surface resampling (i.e. the next step) was required to compute inter-correlations between individuals' motion patterns. 

The second step consisted of measuring similarities between individual's motion patterns  (i.e. between the individual time-average feature vectors) using a temporal inter-subject comparison (vertex-to-vertex) based on a spherical interpolation of vertices and their corresponding texture values in the unit sphere using the Kdtree interpolation approach for surface resampling, included in the Spherical Demons framework~\cite{yeo2009spherical}.

\begin{figure*}[!h]
\centering
\subfigure[Proposed feature (in mm$^{-1}$)]
{\includegraphics[scale=.31]{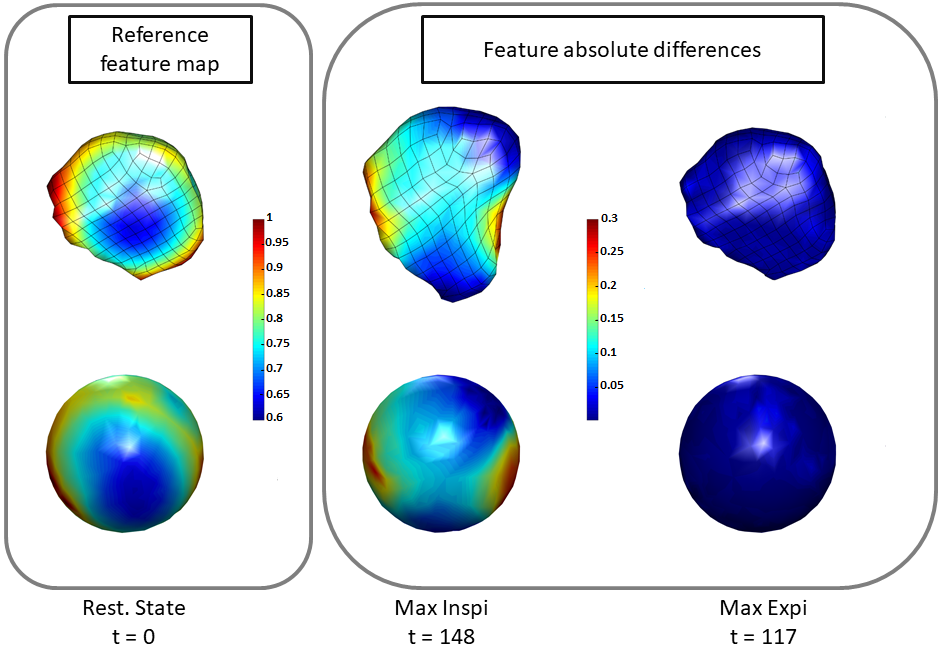}}
\hspace{0.5cm}
\subfigure[Mean curvature (in mm$^{-1}$)]
{\includegraphics[scale=.31]{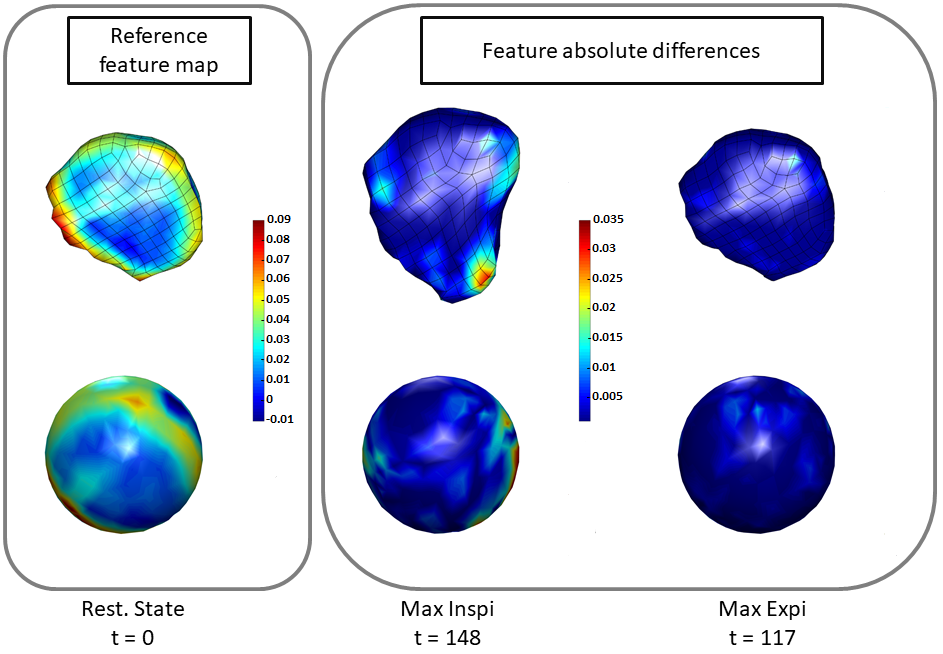}}
\hspace{0.5cm}
\subfigure[Mesh distortions (in degrees)]
{\includegraphics[scale=.31]{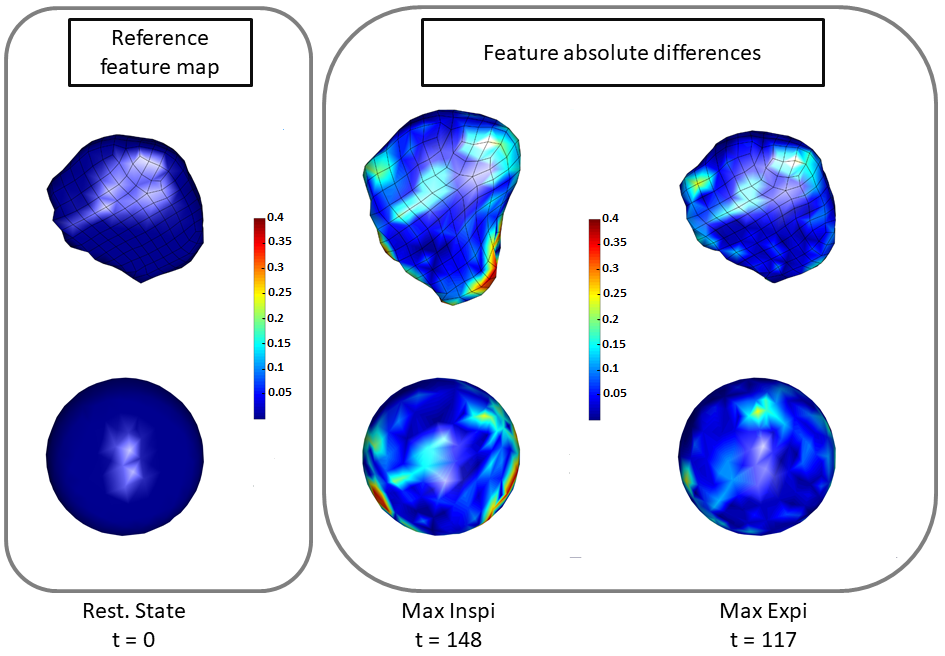}}
\hspace{0.5cm}
\subfigure[Mesh elongations (in mm)]
{\includegraphics[scale=.31]{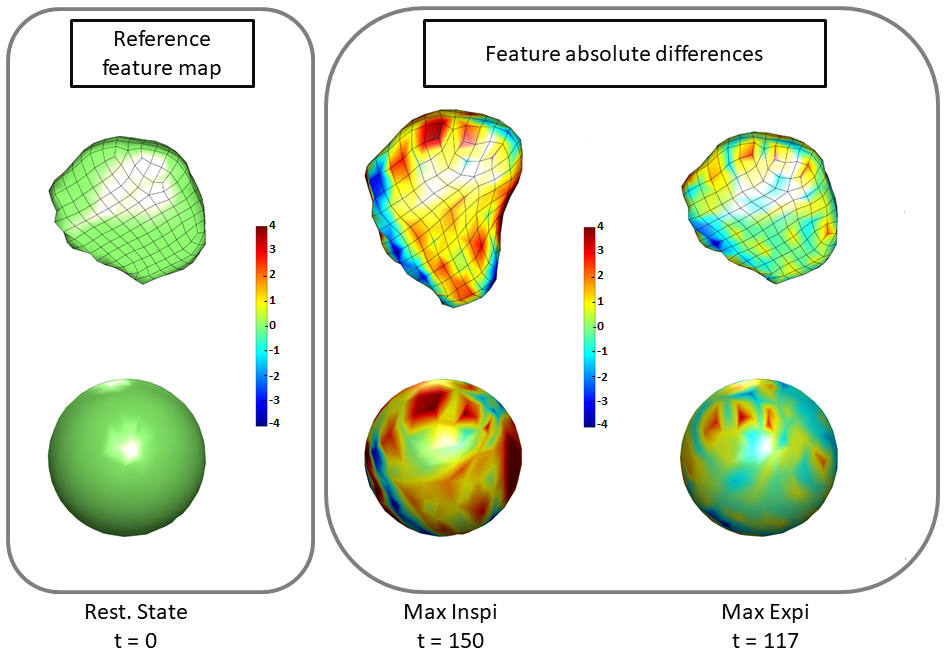}}
    \caption{Feature projection on the sphere. For each panel, the first column represents the reference feature map, while the $2^{nd}$ and $3^{rd}$ columns represent the feature changes at maximal range of motion w.r.t the reference state (patterns of the highest simulated organ deformation).} 
    \label{fig:spherical_maps}
\end{figure*}

Fig.~\ref{fig:spherical_maps} shows the quality of feature projections on the unit sphere for the simulated sequence.
In clinical practice, such projections will provide full 4D information using a common and simple shape representation to easily localize the most important and common deformations at the population level.

\begin{figure}[ht]
\centering
\includegraphics[scale=0.6]{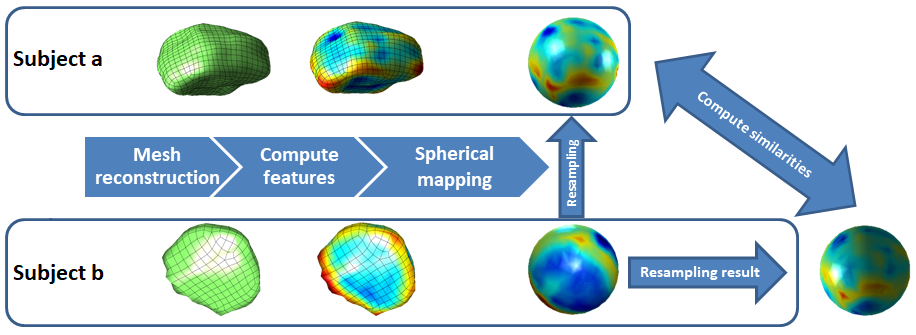}
\caption{\label{fig:pipeline_sphMap} The pipeline of the proposed framework of comparing two bladder surfaces with different parameterizations. This Figure summarizes the key steps leading to perform inter-subject comparisons. In this example, comparisons are based on the proposed feature. To complete this process, we resample the projected feature texture for "subject b" to fit the feature map projection for "subject a"  using a KdTree interpolation.}
\label{fig:inter_subject_pipeline}
\end{figure}

\begin{figure*}
\centering
\includegraphics[scale=.6]{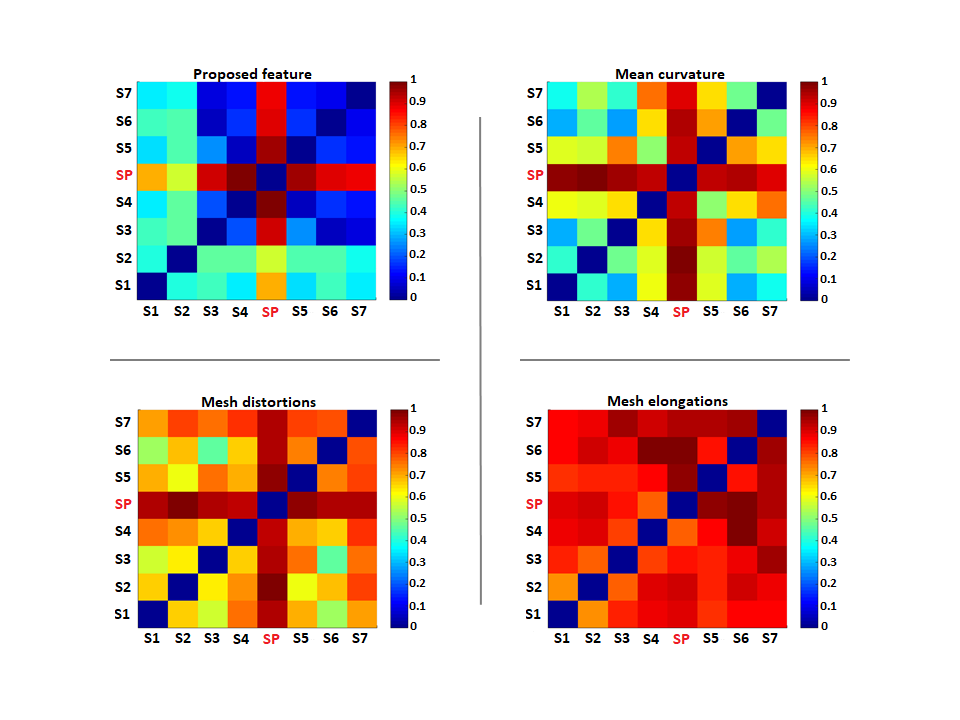}
\vspace{-1cm}
\caption{\label{fig:distMat_sphAllFeat} Inter-subject comparison using LBO spherical mapping with a Kdtree interpolation. Comparisons were based on the correlations between subjects' motion patterns in a common shape space (the unit sphere): For each feature, the matrix elements were filled using the distance function $dist(\bar{\mathcal{F}_i}, \bar{\mathcal{F}_j}) = 1 - normcorr(\bar{\mathcal{F}_i}, \bar{\mathcal{F}_j}) $, where $\bar{\mathcal{F}_i}$ is the resampled average map for distances between temporal feature vectors and the reference one, for subject $S_i$, and $normcorr$ is the normalized correlation function.}
\end{figure*}

Fig.~\ref{fig:pipeline_sphMap} illustrates the complete pipeline for measuring the similarity between two surfaces with different resolutions based on the normalized proposed feature. All the results of inter-subject comparison using LBO spherical parameterization are given in Fig.~\ref{fig:distMat_sphAllFeat}. These results are consistent with those obtained using deformation depth as a global metric for inter-subject comparisons. They confirm the ability of our descriptor to discriminate between normal and abnormal motion.

\section{Discussion and conclusion}
\subsection{Comparison with related works and their clinical relevance}
Actually, only 4D simulations (biomechanical models) of the 4D pelvic organ motion are available in the literature~\cite{chen2015female, courtecuisse2020three}. This work represents a research initiative that aims to assess the 4D motion of such organs from realistic observations using noninvasive imaging techniques. Hence, the data reconstruction itself reveals clinically relevant insights
into the understanding of pelvic floor dynamics. Furthermore, there is a need for a consensus on the image acquisition protocols themselves~\cite{garetier2020dynamic}. And the results of the current work will be used to adjust the current acquisition protocols for dynamic MRI sequences to make them more exploitable and well adapted for 3D reconstruction and post-processing.
This study represents an improved extension of the works of~\cite{rahim2013diffeomorphic} from 2D+$t$ to 3D+$t$ for characterizing pelvic floor organ dynamics.
In 2D+$t$ studies, the out of plane problem related to the acquisition protocols and conditions can lead to incomplete or even misleading interpretations, which can prohibit their use in clinical settings. Moreover, this natural extension was required because of the variability of abdominal organ shapes across time and subjects. To address these competing concerns, statistical shape analyses were performed in the high resolution domain to allow quantification of the 4D motion of the bladder. Given the fact that this extension has pointed out that the bladder deformations occur essentially in top lateral regions, then it becomes obvious that this valuable information cannot be directly extracted from dynamical sagittal slices. This makes it unnecessary to perform a direct comparison between the two studies.

% \hl{Furthermore, our work providing more realistic boundary conditions to analyse organ deformations will help biomechanists and physicians to improve simulation accuracy and to assess the dynamic behavior of the pelvic organs that is governed typically by local changes in surface characteristics.\\}

In terms of methodologies and tools, the novelty in this work is that we characterize the organ dynamics only from surfacic information (i.e. using surface meshes) contrary to finite element modeling for stress analysis using volumetric meshes which also discretize the interior structure of the organ~\cite{chen2015female, courtecuisse2020three}. Our fine characterization can be clinically useful to measure respirations indirectly through the deformation induced by their actions on bladder shape: in addition to the assessment of respiration depth that can be deduced from the deformation depth parameter, Fig.~\ref{fig:temporal_stability_realistic} illustrates its use as a promising tool for identifying subject-specific respiratory frequencies by exploiting properties of periodicity in correlation trajectories.
% Our fine characterization of bladder shape trajectories will provide valuable information to clinicians about individual respiratory frequencies (from periodicity of correlation trajectories) and lungs capacities (from deformation depths).

\subsection{Surface representation and motion characterization}
In this study, we combine parametric methods (LDDMM, for surface representation) and non-parametric methods (an Eulerian PDE approach, for surface characterization). The LDDMM was employed to represent the organ surfaces as 4D smooth quadrilateral meshes, empty from irregular situations such as singular vertices near the organ shape poles, as encountered in~\cite{chen2015female}. The idea behind was to parameterize large deformations with only representative samples of surface points encoding the major shape variability with reasonable computational costs within large datasets. The use of the LDDMM, employing Hamiltonian statistical mechanics, aimed at providing an hypothesis compatible with the physics of deformations. This helped for establishing a compact shape representation relying on structured meshes for the organ surface with fixed connectivity information.
A set of 3D geometric descriptors was employed to extract meaningful features characterizing the bladder shape dynamics during loading exercices. Elongations and distortions are a well known biomechanical parameters which have been used in~\cite{rahim2013diffeomorphic} to classify subjects into pathological patients and healthy control groups. 
However, since we would like to characterize the geometry of highly curved organs, then statistical tools derived from Euclidean geometry are not the most appropriate to deal with such clinical issues. In addition to the extension of elongations and distortions from 2D to 3D, we have also employed two shape descriptors involving the notion of geodesic distances for a finer characterization of organ dynamics: the mean curvature and our proposed feature were used to identify salient motion patterns.

To study the temporal changes in mean curvature, we have used the method proposed in~\cite{rusinkiewicz2004estimating}, to estimate curvatures on triangle meshes. This required the conversion of temporal quad meshes to triangular ones and the major limitation was the dependency of the algorithm on mesh resolution and vertex neighborhood size (this algorithm is well adapted to high-resolution meshes). To cope with this dependency, we proposed the use of a fully non-parametric Eulerian PDE approach to explore the space of continuous maps from shape surface into a surrounding sphere surface.  

To analyze descriptors' performances, we have deformed the bladder volume in the static MRI scan for one subject to simulate a biologically plausible model of abnormal shape trajectory using the LEPF (this aimed to locally control the deformation kinematics, while providing globally nonlinear deformation). The results of our experiments on several descriptors show the effectiveness of the proposed shape descriptor to characterize surface dynamics through its application to this cyclic trajectory, characterized by its large deformation-depth and long-term time variations (see Fig~\ref{fig:temporal_stability2}).

% While the performance of other descriptors decreases with time, the proposed descriptor provide a stable deformation depth throughout motion, for different initial mesh qualities. (see Figures~\ref{fig:temporal_stability2} and~\ref{fig:temporal_stability_realistic}).

Both these results on synthetic data and the results illustrated in Fig~\ref{fig:temporal_stability_realistic} for characterization of a realistic organ shape trajectory show that the numerical stability of elongation and distortion measures decreases drastically across cycles. These examples also show that the proposed feature is numerically much more stable than tensor-based mean curvature which confirms the findings of \textit{Nava-Yazdani et al.}~\cite{nava2019geodesic} that computing tensors impacts numerical stability in time-dependent shape data analysis. In fact, mean curvature is an extrinsic measure of curvature which corresponds to layman's understanding of curvature before we were ever introduced to differential geometry. It is often employed in the Level set method for curvature-driven segmentation tasks~\cite{paragios2003level}. Its expression for non-parametric surfaces is also known as the divergence formula~\cite{goldman2005curvature}.
In~\cite{albin2016computational}, \textit{Albin et al.} showed that curvature estimates from implicit surfaces (for which the computations are performed in Cartesian coordinates) are more accurate than those calculated from meshes. This can explain the fact that non-parametric methods are more accurate and stable than parametric ones. In the same context, we have studied the sensitivity of parametric methods to mesh quality for estimating brain curvature tensor in~\cite{makki2021towards}. In appendix~\ref{Appendix A}, we show that the non-parametric approaches can successfully avoid the need to compute metric tensors explicitly while taking into account the non-Euclidean geometry of organ curved surfaces.\\

On the other hand, comparisons between two or more patients may help identify movement abnormalities and may serve for subject classification: into healthy and pathological, or into subgroups, sharing similar organ motion characteristics within large data sets.\\
An inter-individual comparison is then performed by introducing the simulated sequence to the realistic dataset.
A global metric is first achieved using the deformation depth. It reflects the maximum breathing capacity of each subject since the organ surface deforms the most by maximum of inspiration.
As illustrated in Fig.~\ref{fig:intersub_distMat}, a quantification of subject-specific deformation depths during deep breathing exercices is capable of determining differences between patients (left panel) and groups (right panel). This may serve to optimize both surgical and non-surgical treatments of pelvic floor disorders with respect to individual breathing capacities.
To make this comparison more statistically meaningful, local motion patterns derived from different descriptors and for different subjects (different parameterizations) were compared in a common space, the unit sphere, through the use of the LBO eigenfuctions for spherical mapping. Results in Figures~\ref{fig:intersub_distMat} and~\ref{fig:distMat_sphAllFeat} illustrate the fact that motion patterns differ across healthy volunteers and highlight the capacity of each descriptor to distinguish the simulated abnormal motion type from statistical measures. The results show also that the geodesic-based features (including curvature), for which the simulated sequence was less correlated with realistic ones, are the best classifiers against traditional classifiers based on Euclidean geometry. In fact, each feature have described an organ shape trajectory by integrating over time all the available information from each frame. However, we observed that for a deformation characterization task, not all frames contain salient spatio-temporal informations which are discriminative to different classes of deformations. Indeed, many frames contain non-salient motions which are irrelevant to the performed deformation. The characterization of the simulated trajectory represents a good example to illustrate these findings (Fig.~\ref{fig:temporal_stability2}). See also Fig.~\ref{fig:temporal_stability_realistic}, in which mesh distortions and elongations provided more "non-salient" motion patterns, thus affecting the quality of the corresponding correlation trajectories.

\subsection{Limitations of the study and future scopes}

Although the fact that we propose a pipeline with minimal user intervention, let us recall that it is not yet fully automatic, since it requires manual segmentation of the acquired dynamical slices used for the reconstruction of organ volumes. Furthermore, it is required to establish only the initial quad mesh for the organ surface using Instant meshes. Concerning the proposed descriptor, we are considering only isotropic volumes for now.

Future work will be towards addressing these limitations with the help of deep learning approaches for slice segmentation~\cite{ogier2021overview}, and to fully integrate the open source software of Instant Meshes~\cite{jakob2015instant} in a way to make the entire pipeline easy to handle for physicians and nurses. 
% \hl{Moreover, all the results presented in Section}~\ref{sec:geom_desc} \hl{can exclude the following possibility: the observed differences for the geodesic-based feature can be due to differences in the estimated correspondences rather than the features themselves.}
All the source codes developed and used in this work are available at \url{https://github.com/k16makki/dynPelvis/tree/master/Dynpelvis3D}.

Future works will also include clinico-pathological data from age-matched women with uterine and bladder prolapse. The proposed methods will be employed to identify morphological differences between normal and pathological groups.  These techniques could also be applied to study and characterize the dynamics of other functional human tissues and organs such as the heart. While it cannot be inferred from this study, it seems reasonable to hypothesize that the proposed tools can be employed to perform longitudinal analysis during organ disease development or following a therapy for pelvic floor disorders.

\section*{Declaration of Competing Interests}
The authors declare that they have no conflict of interest.

\section*{Acknowledgments}
This research was supported in part by the AMIDEX - Institut Carnot STAR under the DynPelvis3D grant.

\bibliographystyle{model1-num-names}
\renewcommand{\refname}{References}
\bibliography{refs.bib}

\begin{thebibliography}{68}
\expandafter\ifx\csname natexlab\endcsname\relax\def\natexlab#1{#1}\fi
\providecommand{\bibinfo}[2]{#2}
\ifx\xfnm\relax \def\xfnm[#1]{\unskip,\space#1}\fi
%Type = Article
\bibitem[{Law and Fielding(2008)}]{law2008mri}
\bibinfo{author}{Y.~M. Law}, \bibinfo{author}{J.~R. Fielding},
\newblock \bibinfo{title}{{MRI} of pelvic floor dysfunction},
\newblock \bibinfo{journal}{American Journal of Roentgenology}
  \bibinfo{volume}{191} (\bibinfo{year}{2008}) \bibinfo{pages}{S45--S53}.
%Type = Article
\bibitem[{El~Sayed et~al.(2017)El~Sayed, Alt, Maccioni, Meissnitzer, Masselli,
  Manganaro, Vinci, Weishaupt, ESUR, Group et~al.}]{el2017magnetic}
\bibinfo{author}{R.~F. El~Sayed}, \bibinfo{author}{C.~D. Alt},
  \bibinfo{author}{F.~Maccioni}, \bibinfo{author}{M.~Meissnitzer},
  \bibinfo{author}{G.~Masselli}, \bibinfo{author}{L.~Manganaro},
  \bibinfo{author}{V.~Vinci}, \bibinfo{author}{D.~Weishaupt},
  \bibinfo{author}{ESUR}, \bibinfo{author}{E.~P. F.~W. Group}, et~al.,
\newblock \bibinfo{title}{Magnetic resonance imaging of pelvic floor
  dysfunction-joint recommendations of the {ESUR} and {ESGAR} {P}elvic {F}loor
  {W}orking {G}roup},
\newblock \bibinfo{journal}{European radiology} \bibinfo{volume}{27}
  (\bibinfo{year}{2017}) \bibinfo{pages}{2067--2085}.
%Type = Article
\bibitem[{Jourdan et~al.(2021)Jourdan, Le~Troter, Daude, Rapacchi, Masson,
  B{\`e}ge, and Bendahan}]{jourdan2021semiautomatic}
\bibinfo{author}{A.~Jourdan}, \bibinfo{author}{A.~Le~Troter},
  \bibinfo{author}{P.~Daude}, \bibinfo{author}{S.~Rapacchi},
  \bibinfo{author}{C.~Masson}, \bibinfo{author}{T.~B{\`e}ge},
  \bibinfo{author}{D.~Bendahan},
\newblock \bibinfo{title}{Semiautomatic quantification of abdominal wall
  muscles deformations based on dynamic {MRI} image registration},
\newblock \bibinfo{journal}{NMR in Biomedicine}  (\bibinfo{year}{2021})
  \bibinfo{pages}{e4470}.
%Type = Article
\bibitem[{Garetier et~al.(2020)Garetier, Borotikar, Makki, Brochard, Rousseau,
  and Ben~Salem}]{garetier2020dynamic}
\bibinfo{author}{M.~Garetier}, \bibinfo{author}{B.~Borotikar},
  \bibinfo{author}{K.~Makki}, \bibinfo{author}{S.~Brochard},
  \bibinfo{author}{F.~Rousseau}, \bibinfo{author}{D.~Ben~Salem},
\newblock \bibinfo{title}{Dynamic {MRI} for articulating joint evaluation on
  1.5 {T} and 3.0 {T} scanners: setup, protocols, and real-time sequences},
\newblock \bibinfo{journal}{Insights into Imaging} \bibinfo{volume}{11}
  (\bibinfo{year}{2020}) \bibinfo{pages}{1--10}.
%Type = Article
\bibitem[{Chen et~al.(2021)Chen, van Sloun, Turco, Wijkstra, Filomena, Agati,
  Houthuizen, and Mischi}]{chen2021blood}
\bibinfo{author}{P.~Chen}, \bibinfo{author}{R.~J. van Sloun},
  \bibinfo{author}{S.~Turco}, \bibinfo{author}{H.~Wijkstra},
  \bibinfo{author}{D.~Filomena}, \bibinfo{author}{L.~Agati},
  \bibinfo{author}{P.~Houthuizen}, \bibinfo{author}{M.~Mischi},
\newblock \bibinfo{title}{Blood flow patterns estimation in the left ventricle
  with low-rate 2{D} and 3{D} dynamic contrast-enhanced ultrasound},
\newblock \bibinfo{journal}{Computer Methods and Programs in Biomedicine}
  \bibinfo{volume}{198} (\bibinfo{year}{2021}) \bibinfo{pages}{105810}.
%Type = Article
\bibitem[{Gorji and Pourmomeny(2020)}]{gorji2020effect}
\bibinfo{author}{Z.~Gorji}, \bibinfo{author}{A.~A. Pourmomeny},
\newblock \bibinfo{title}{The effect of pelvic floor muscles training using
  biofeedback on symptoms of pelvic prolapse and quality of life in affected
  females},
\newblock \bibinfo{journal}{International Journal of Biomedicine and Public
  Health} \bibinfo{volume}{3} (\bibinfo{year}{2020}) \bibinfo{pages}{5--9}.
%Type = Article
\bibitem[{Chen et~al.(2015)Chen, Joli, Feng, Rahim, Pirr{\'o}, and
  Bellemare}]{chen2015female}
\bibinfo{author}{Z.-W. Chen}, \bibinfo{author}{P.~Joli}, \bibinfo{author}{Z.-Q.
  Feng}, \bibinfo{author}{M.~Rahim}, \bibinfo{author}{N.~Pirr{\'o}},
  \bibinfo{author}{M.-E. Bellemare},
\newblock \bibinfo{title}{Female patient-specific finite element modeling of
  pelvic organ prolapse ({POP})},
\newblock \bibinfo{journal}{Journal of biomechanics} \bibinfo{volume}{48}
  (\bibinfo{year}{2015}) \bibinfo{pages}{238--245}.
%Type = Article
\bibitem[{Courtecuisse et~al.(2020)Courtecuisse, Jiang, Mayeur, Witz,
  Lecomte-Grosbras, Cosson, Brieu, and Cotin}]{courtecuisse2020three}
\bibinfo{author}{H.~Courtecuisse}, \bibinfo{author}{Z.~Jiang},
  \bibinfo{author}{O.~Mayeur}, \bibinfo{author}{J.~Witz},
  \bibinfo{author}{P.~Lecomte-Grosbras}, \bibinfo{author}{M.~Cosson},
  \bibinfo{author}{M.~Brieu}, \bibinfo{author}{S.~Cotin},
\newblock \bibinfo{title}{Three-dimensional physics-based registration of
  pelvic system using {2D} dynamic magnetic resonance imaging slices},
\newblock \bibinfo{journal}{Strain} \bibinfo{volume}{56} (\bibinfo{year}{2020})
  \bibinfo{pages}{e12339}.
%Type = Inproceedings
\bibitem[{Ogier et~al.(2019)Ogier, Rapacchi, Le~Troter, and
  Bellemare}]{ogier20193d}
\bibinfo{author}{A.~C. Ogier}, \bibinfo{author}{S.~Rapacchi},
  \bibinfo{author}{A.~Le~Troter}, \bibinfo{author}{M.-E. Bellemare},
\newblock \bibinfo{title}{{3D} dynamic {MRI} for pelvis observation-a first
  step},
\newblock in: \bibinfo{booktitle}{2019 IEEE 16th International Symposium on
  Biomedical Imaging (ISBI 2019)}, \bibinfo{organization}{IEEE}, pp.
  \bibinfo{pages}{1801--1804}.
%Type = Article
\bibitem[{Makki et~al.(2020)Makki, Bohi, Ogier, and Bellemare}]{makki2020new}
\bibinfo{author}{K.~Makki}, \bibinfo{author}{A.~Bohi}, \bibinfo{author}{A.~C.
  Ogier}, \bibinfo{author}{M.-E. Bellemare},
\newblock \bibinfo{title}{A new geodesic-based feature for characterization of
  {3D} shapes: application to soft tissue organ temporal deformations},
\newblock \bibinfo{journal}{25th International Conference on Pattern
  Recognition (ICPR2020), Jan 2021, Milan, Italy}  (\bibinfo{year}{2020}).
%Type = Article
\bibitem[{Br{\"u}ning et~al.(2020)Br{\"u}ning, Hildebrandt, Heppt, Schmidt,
  Lamecker, Szengel, Amiridze, Ramm, Bindernagel, Zachow
  et~al.}]{bruning2020characterization}
\bibinfo{author}{J.~Br{\"u}ning}, \bibinfo{author}{T.~Hildebrandt},
  \bibinfo{author}{W.~Heppt}, \bibinfo{author}{N.~Schmidt},
  \bibinfo{author}{H.~Lamecker}, \bibinfo{author}{A.~Szengel},
  \bibinfo{author}{N.~Amiridze}, \bibinfo{author}{H.~Ramm},
  \bibinfo{author}{M.~Bindernagel}, \bibinfo{author}{S.~Zachow}, et~al.,
\newblock \bibinfo{title}{Characterization of the airflow within an average
  geometry of the healthy human nasal cavity},
\newblock \bibinfo{journal}{Scientific Reports} \bibinfo{volume}{10}
  (\bibinfo{year}{2020}) \bibinfo{pages}{1--12}.
%Type = Incollection
\bibitem[{Pennec(2020)}]{pennec2020advances}
\bibinfo{author}{X.~Pennec},
\newblock \bibinfo{title}{Advances in geometric statistics for manifold
  dimension reduction},
\newblock in: \bibinfo{booktitle}{Handbook of Variational Methods for Nonlinear
  Geometric Data}, \bibinfo{publisher}{Springer}, \bibinfo{year}{2020}, pp.
  \bibinfo{pages}{339--359}.
%Type = Article
\bibitem[{Zheng et~al.(2019)Zheng, Delingette, Fung, Petersen, and
  Ayache}]{zheng2019unsupervised}
\bibinfo{author}{Q.~Zheng}, \bibinfo{author}{H.~Delingette},
  \bibinfo{author}{K.~Fung}, \bibinfo{author}{S.~E. Petersen},
  \bibinfo{author}{N.~Ayache},
\newblock \bibinfo{title}{Unsupervised shape and motion analysis of 3822
  cardiac 4{D} {MRI}s of {UK} biobank},
\newblock \bibinfo{journal}{arXiv preprint arXiv:1902.05811}
  (\bibinfo{year}{2019}).
%Type = Inproceedings
\bibitem[{Abbas et~al.(2019)Abbas, Fishbaugh, Petchprapa, Lattanzi, and
  Gerig}]{abbas2019analysis}
\bibinfo{author}{B.~Abbas}, \bibinfo{author}{J.~Fishbaugh},
  \bibinfo{author}{C.~Petchprapa}, \bibinfo{author}{R.~Lattanzi},
  \bibinfo{author}{G.~Gerig},
\newblock \bibinfo{title}{Analysis of the kinematic motion of the wrist from
  4{D} magnetic resonance imaging},
\newblock in: \bibinfo{booktitle}{Medical Imaging 2019: Image Processing},
  volume \bibinfo{volume}{10949}, \bibinfo{organization}{International Society
  for Optics and Photonics}, p. \bibinfo{pages}{109491E}.
%Type = Inproceedings
\bibitem[{Hong et~al.(2018)Hong, Fishbaugh, and Gerig}]{hong20184d}
\bibinfo{author}{S.~Hong}, \bibinfo{author}{J.~Fishbaugh},
  \bibinfo{author}{G.~Gerig},
\newblock \bibinfo{title}{4{D} continuous medial representation trajectory
  estimation for longitudinal shape analysis},
\newblock in: \bibinfo{booktitle}{International Workshop on Shape in Medical
  Imaging}, \bibinfo{organization}{Springer}, pp. \bibinfo{pages}{125--136}.
%Type = Article
\bibitem[{Makki et~al.(2019)Makki, Borotikar, Garetier, Brochard, Salem, and
  Rousseau}]{makki2019vivo}
\bibinfo{author}{K.~Makki}, \bibinfo{author}{B.~Borotikar},
  \bibinfo{author}{M.~Garetier}, \bibinfo{author}{S.~Brochard},
  \bibinfo{author}{D.~B. Salem}, \bibinfo{author}{F.~Rousseau},
\newblock \bibinfo{title}{In vivo ankle joint kinematics from dynamic magnetic
  resonance imaging using a registration-based framework},
\newblock \bibinfo{journal}{Journal of biomechanics} \bibinfo{volume}{86}
  (\bibinfo{year}{2019}) \bibinfo{pages}{193--203}.
%Type = Article
\bibitem[{Heimann and Meinzer(2009)}]{heimann2009statistical}
\bibinfo{author}{T.~Heimann}, \bibinfo{author}{H.-P. Meinzer},
\newblock \bibinfo{title}{Statistical shape models for 3{D} medical image
  segmentation: a review},
\newblock \bibinfo{journal}{Medical image analysis} \bibinfo{volume}{13}
  (\bibinfo{year}{2009}) \bibinfo{pages}{543--563}.
%Type = Article
\bibitem[{Pennec(2020)}]{pennec2020statistical}
\bibinfo{author}{X.~Pennec},
\newblock \bibinfo{title}{Statistical analysis of organs' shapes and
  deformations: the {R}iemannian and the affine settings in computational
  anatomy}  (\bibinfo{year}{2020}).
%Type = Article
\bibitem[{Fishbaugh et~al.(2017)Fishbaugh, Durrleman, Prastawa, and
  Gerig}]{fishbaugh2017geodesic}
\bibinfo{author}{J.~Fishbaugh}, \bibinfo{author}{S.~Durrleman},
  \bibinfo{author}{M.~Prastawa}, \bibinfo{author}{G.~Gerig},
\newblock \bibinfo{title}{Geodesic shape regression with multiple geometries
  and sparse parameters},
\newblock \bibinfo{journal}{Medical image analysis} \bibinfo{volume}{39}
  (\bibinfo{year}{2017}) \bibinfo{pages}{1--17}.
%Type = Misc
\bibitem[{Zhang and Golland(2016)}]{zhang2016statistical}
\bibinfo{author}{M.~Zhang}, \bibinfo{author}{P.~Golland},
  \bibinfo{title}{Statistical shape analysis: From landmarks to
  diffeomorphisms}, \bibinfo{year}{2016}.
%Type = Inproceedings
\bibitem[{Fishbaugh et~al.(2013)Fishbaugh, Prastawa, Gerig, and
  Durrleman}]{fishbaugh2013geodesic}
\bibinfo{author}{J.~Fishbaugh}, \bibinfo{author}{M.~Prastawa},
  \bibinfo{author}{G.~Gerig}, \bibinfo{author}{S.~Durrleman},
\newblock \bibinfo{title}{Geodesic shape regression in the framework of
  currents},
\newblock in: \bibinfo{booktitle}{International Conference on Information
  Processing in Medical Imaging}, \bibinfo{organization}{Springer}, pp.
  \bibinfo{pages}{718--729}.
%Type = Article
\bibitem[{Abi~Nader et~al.(2020)Abi~Nader, Ayache, Robert, Lorenzi, Initiative
  et~al.}]{abi2020monotonic}
\bibinfo{author}{C.~Abi~Nader}, \bibinfo{author}{N.~Ayache},
  \bibinfo{author}{P.~Robert}, \bibinfo{author}{M.~Lorenzi},
  \bibinfo{author}{A.~D.~N. Initiative}, et~al.,
\newblock \bibinfo{title}{Monotonic gaussian process for spatio-temporal
  disease progression modeling in brain imaging data},
\newblock \bibinfo{journal}{NeuroImage} \bibinfo{volume}{205}
  (\bibinfo{year}{2020}) \bibinfo{pages}{116266}.
%Type = Inproceedings
\bibitem[{Zolfaghari et~al.(2014)Zolfaghari, Epain, Jin, Tew, and
  Glaunes}]{zolfaghari2014multiscale}
\bibinfo{author}{R.~Zolfaghari}, \bibinfo{author}{N.~Epain},
  \bibinfo{author}{C.~T. Jin}, \bibinfo{author}{A.~Tew},
  \bibinfo{author}{J.~Glaunes},
\newblock \bibinfo{title}{A multiscale {LDDMM} template algorithm for studying
  ear shape variations},
\newblock in: \bibinfo{booktitle}{2014 8th International Conference on Signal
  Processing and Communication Systems (ICSPCS)}, \bibinfo{organization}{IEEE},
  pp. \bibinfo{pages}{1--6}.
%Type = Inproceedings
\bibitem[{B{\^o}ne et~al.(2018)B{\^o}ne, Colliot, and
  Durrleman}]{bone2018learning}
\bibinfo{author}{A.~B{\^o}ne}, \bibinfo{author}{O.~Colliot},
  \bibinfo{author}{S.~Durrleman},
\newblock \bibinfo{title}{Learning distributions of shape trajectories from
  longitudinal datasets: a hierarchical model on a manifold of
  diffeomorphisms},
\newblock in: \bibinfo{booktitle}{Proceedings of the IEEE Conference on
  Computer Vision and Pattern Recognition}, pp. \bibinfo{pages}{9271--9280}.
%Type = Incollection
\bibitem[{Peyr{\'e} and Cohen(2009)}]{peyre2009geodesic}
\bibinfo{author}{G.~Peyr{\'e}}, \bibinfo{author}{L.~D. Cohen},
\newblock \bibinfo{title}{Geodesic methods for shape and surface processing},
\newblock in: \bibinfo{booktitle}{Advances in Computational Vision and Medical
  Image Processing}, \bibinfo{publisher}{Springer}, \bibinfo{year}{2009}, pp.
  \bibinfo{pages}{29--56}.
%Type = Incollection
\bibitem[{Malament(2012)}]{malament2012remark}
\bibinfo{author}{D.~B. Malament},
\newblock \bibinfo{title}{A remark about the “geodesic principle” in
  general relativity},
\newblock in: \bibinfo{booktitle}{Analysis and Interpretation in the Exact
  Sciences}, \bibinfo{publisher}{Springer}, \bibinfo{year}{2012}, pp.
  \bibinfo{pages}{245--252}.
%Type = Inproceedings
\bibitem[{Sun et~al.(2015)Sun, Lelieveldt, and Staring}]{sun2015fast}
\bibinfo{author}{Z.~Sun}, \bibinfo{author}{B.~P. Lelieveldt},
  \bibinfo{author}{M.~Staring},
\newblock \bibinfo{title}{Fast linear geodesic shape regression using coupled
  logdemons registration},
\newblock in: \bibinfo{booktitle}{2015 IEEE 12th International Symposium on
  Biomedical Imaging (ISBI)}, \bibinfo{organization}{IEEE}, pp.
  \bibinfo{pages}{1276--1279}.
%Type = Article
\bibitem[{Kim et~al.(2020)Kim, Styner, Piven, and Gerig}]{kim2020framework}
\bibinfo{author}{H.~Kim}, \bibinfo{author}{M.~Styner},
  \bibinfo{author}{J.~Piven}, \bibinfo{author}{G.~Gerig},
\newblock \bibinfo{title}{A framework to construct a longitudinal {DW-MRI}
  infant atlas based on mixed effects modeling of d{ODF} coefficients},
\newblock \bibinfo{journal}{arXiv preprint arXiv:2003.05091}
  (\bibinfo{year}{2020}).
%Type = Article
\bibitem[{Rios et~al.(2017)Rios, De~Crevoisier, Ospina, Commandeur, Lafond,
  Simon, Haigron, Espinosa, and Acosta}]{rios2017population}
\bibinfo{author}{R.~Rios}, \bibinfo{author}{R.~De~Crevoisier},
  \bibinfo{author}{J.~D. Ospina}, \bibinfo{author}{F.~Commandeur},
  \bibinfo{author}{C.~Lafond}, \bibinfo{author}{A.~Simon},
  \bibinfo{author}{P.~Haigron}, \bibinfo{author}{J.~Espinosa},
  \bibinfo{author}{O.~Acosta},
\newblock \bibinfo{title}{Population model of bladder motion and deformation
  based on dominant eigenmodes and mixed-effects models in prostate cancer
  radiotherapy},
\newblock \bibinfo{journal}{Medical image analysis} \bibinfo{volume}{38}
  (\bibinfo{year}{2017}) \bibinfo{pages}{133--149}.
%Type = Article
\bibitem[{Luo et~al.(2016)Luo, Jin, Yang, Wang, Li, Guo, Ran, Liu, Zhou, and
  Wu}]{luo2016interfractional}
\bibinfo{author}{H.~Luo}, \bibinfo{author}{F.~Jin}, \bibinfo{author}{D.~Yang},
  \bibinfo{author}{Y.~Wang}, \bibinfo{author}{C.~Li}, \bibinfo{author}{M.~Guo},
  \bibinfo{author}{X.~Ran}, \bibinfo{author}{X.~Liu},
  \bibinfo{author}{Q.~Zhou}, \bibinfo{author}{Y.~Wu},
\newblock \bibinfo{title}{Interfractional variation in bladder volume and its
  impact on cervical cancer radiotherapy: Clinical significance of portable
  bladder scanner},
\newblock \bibinfo{journal}{Medical physics} \bibinfo{volume}{43}
  (\bibinfo{year}{2016}) \bibinfo{pages}{4412--4419}.
%Type = Article
\bibitem[{Kendall(1984)}]{kendall1984shape}
\bibinfo{author}{D.~G. Kendall},
\newblock \bibinfo{title}{Shape manifolds, procrustean metrics, and complex
  projective spaces},
\newblock \bibinfo{journal}{Bulletin of the London mathematical society}
  \bibinfo{volume}{16} (\bibinfo{year}{1984}) \bibinfo{pages}{81--121}.
%Type = Incollection
\bibitem[{Nava-Yazdani et~al.(2019)Nava-Yazdani, Hege, and von
  Tycowicz}]{nava2019geodesic}
\bibinfo{author}{E.~Nava-Yazdani}, \bibinfo{author}{H.-C. Hege},
  \bibinfo{author}{C.~von Tycowicz},
\newblock \bibinfo{title}{A geodesic mixed effects model in {K}endall’s shape
  space},
\newblock in: \bibinfo{booktitle}{Multimodal Brain Image Analysis and
  Mathematical Foundations of Computational Anatomy},
  \bibinfo{publisher}{Springer}, \bibinfo{year}{2019}, pp.
  \bibinfo{pages}{209--218}.
%Type = Inproceedings
\bibitem[{Billet et~al.(2008)Billet, Sermesant, Delingette, and
  Ayache}]{billet2008cardiac}
\bibinfo{author}{F.~Billet}, \bibinfo{author}{M.~Sermesant},
  \bibinfo{author}{H.~Delingette}, \bibinfo{author}{N.~Ayache},
\newblock \bibinfo{title}{Cardiac motion recovery by coupling an
  electromechanical model and cine-{MRI} data: First steps},
\newblock in: \bibinfo{booktitle}{Proc. of the Workshop on Computational
  Biomechanics for Medicine III.(Workshop MICCAI-2008)},
  volume~\bibinfo{volume}{55}, \bibinfo{organization}{Citeseer}, p.
  \bibinfo{pages}{176}.
%Type = Article
\bibitem[{Lee et~al.(2014)Lee, Woo, Xing, Murano, Stone, and
  Prince}]{lee2014semi}
\bibinfo{author}{J.~Lee}, \bibinfo{author}{J.~Woo}, \bibinfo{author}{F.~Xing},
  \bibinfo{author}{E.~Z. Murano}, \bibinfo{author}{M.~Stone},
  \bibinfo{author}{J.~L. Prince},
\newblock \bibinfo{title}{Semi-automatic segmentation for 3{D} motion analysis
  of the tongue with dynamic {MRI}},
\newblock \bibinfo{journal}{Computerized Medical Imaging and Graphics}
  \bibinfo{volume}{38} (\bibinfo{year}{2014}) \bibinfo{pages}{714--724}.
%Type = Inproceedings
\bibitem[{Makki et~al.(2018)Makki, Borotikar, Garetier, Brochard, Salem, and
  Rousseau}]{makki2018high}
\bibinfo{author}{K.~Makki}, \bibinfo{author}{B.~Borotikar},
  \bibinfo{author}{M.~Garetier}, \bibinfo{author}{S.~Brochard},
  \bibinfo{author}{D.~B. Salem}, \bibinfo{author}{F.~Rousseau},
\newblock \bibinfo{title}{High-resolution temporal reconstruction of ankle
  joint from dynamic {MRI}},
\newblock in: \bibinfo{booktitle}{2018 IEEE 15th International Symposium on
  Biomedical Imaging (ISBI 2018)}, \bibinfo{organization}{IEEE}, pp.
  \bibinfo{pages}{1297--1300}.
%Type = Article
\bibitem[{Arsigny et~al.(2009)Arsigny, Commowick, Ayache, and
  Pennec}]{arsigny2009fast}
\bibinfo{author}{V.~Arsigny}, \bibinfo{author}{O.~Commowick},
  \bibinfo{author}{N.~Ayache}, \bibinfo{author}{X.~Pennec},
\newblock \bibinfo{title}{A fast and log-euclidean polyaffine framework for
  locally linear registration},
\newblock \bibinfo{journal}{Journal of Mathematical Imaging and Vision}
  \bibinfo{volume}{33} (\bibinfo{year}{2009}) \bibinfo{pages}{222--238}.
%Type = Article
\bibitem[{Rahim et~al.(2013)Rahim, Bellemare, Bulot, and
  Pirr{\'o}}]{rahim2013diffeomorphic}
\bibinfo{author}{M.~Rahim}, \bibinfo{author}{M.-E. Bellemare},
  \bibinfo{author}{R.~Bulot}, \bibinfo{author}{N.~Pirr{\'o}},
\newblock \bibinfo{title}{A diffeomorphic mapping based characterization of
  temporal sequences: application to the pelvic organ dynamics assessment},
\newblock \bibinfo{journal}{Journal of mathematical imaging and vision}
  \bibinfo{volume}{47} (\bibinfo{year}{2013}) \bibinfo{pages}{151--164}.
%Type = Article
\bibitem[{Lewiner et~al.(2003)Lewiner, Lopes, Vieira, and
  Tavares}]{lewiner2003efficient}
\bibinfo{author}{T.~Lewiner}, \bibinfo{author}{H.~Lopes},
  \bibinfo{author}{A.~W. Vieira}, \bibinfo{author}{G.~Tavares},
\newblock \bibinfo{title}{Efficient implementation of marching cubes' cases
  with topological guarantees},
\newblock \bibinfo{journal}{Journal of graphics tools} \bibinfo{volume}{8}
  (\bibinfo{year}{2003}) \bibinfo{pages}{1--15}.
%Type = Article
\bibitem[{Jakob et~al.(2015)Jakob, Tarini, Panozzo, and
  Sorkine-Hornung}]{jakob2015instant}
\bibinfo{author}{W.~Jakob}, \bibinfo{author}{M.~Tarini},
  \bibinfo{author}{D.~Panozzo}, \bibinfo{author}{O.~Sorkine-Hornung},
\newblock \bibinfo{title}{Instant field-aligned meshes.},
\newblock \bibinfo{journal}{ACM Trans. Graph.} \bibinfo{volume}{34}
  (\bibinfo{year}{2015}) \bibinfo{pages}{189--1}.
%Type = Article
\bibitem[{Beg et~al.(2005)Beg, Miller, Trouv{\'e}, and
  Younes}]{beg2005computing}
\bibinfo{author}{M.~F. Beg}, \bibinfo{author}{M.~I. Miller},
  \bibinfo{author}{A.~Trouv{\'e}}, \bibinfo{author}{L.~Younes},
\newblock \bibinfo{title}{Computing large deformation metric mappings via
  geodesic flows of diffeomorphisms},
\newblock \bibinfo{journal}{International journal of computer vision}
  \bibinfo{volume}{61} (\bibinfo{year}{2005}) \bibinfo{pages}{139--157}.
%Type = Inproceedings
\bibitem[{V{\'a}{\v{s}}a and Rus(2012)}]{vavsa2012dihedral}
\bibinfo{author}{L.~V{\'a}{\v{s}}a}, \bibinfo{author}{J.~Rus},
\newblock \bibinfo{title}{Dihedral angle mesh error: a fast perception
  correlated distortion measure for fixed connectivity triangle meshes},
\newblock in: \bibinfo{booktitle}{Computer Graphics Forum},
  volume~\bibinfo{volume}{31}, \bibinfo{organization}{Wiley Online Library},
  pp. \bibinfo{pages}{1715--1724}.
%Type = Article
\bibitem[{Dijkstra(1959)}]{dijkstra1959note}
\bibinfo{author}{E.~W. Dijkstra},
\newblock \bibinfo{title}{A note on two problems in connexion with graphs},
\newblock \bibinfo{journal}{Numerische mathematik} \bibinfo{volume}{1}
  (\bibinfo{year}{1959}) \bibinfo{pages}{269--271}.
%Type = Article
\bibitem[{Sethian(1996)}]{sethian1996fast}
\bibinfo{author}{J.~A. Sethian},
\newblock \bibinfo{title}{A fast marching level set method for monotonically
  advancing fronts},
\newblock \bibinfo{journal}{Proceedings of the National Academy of Sciences}
  \bibinfo{volume}{93} (\bibinfo{year}{1996}) \bibinfo{pages}{1591--1595}.
%Type = Article
\bibitem[{Crane et~al.(2017)Crane, Weischedel, and Wardetzky}]{crane2017heat}
\bibinfo{author}{K.~Crane}, \bibinfo{author}{C.~Weischedel},
  \bibinfo{author}{M.~Wardetzky},
\newblock \bibinfo{title}{The heat method for distance computation},
\newblock \bibinfo{journal}{Communications of the ACM} \bibinfo{volume}{60}
  (\bibinfo{year}{2017}) \bibinfo{pages}{90--99}.
%Type = Article
\bibitem[{Yezzi and Prince(2003)}]{yezzi2003eulerian}
\bibinfo{author}{A.~J. Yezzi}, \bibinfo{author}{J.~L. Prince},
\newblock \bibinfo{title}{An {E}ulerian {PDE} approach for computing tissue
  thickness},
\newblock \bibinfo{journal}{IEEE transactions on medical imaging}
  \bibinfo{volume}{22} (\bibinfo{year}{2003}) \bibinfo{pages}{1332--1339}.
%Type = Article
\bibitem[{Acosta et~al.(2009)Acosta, Bourgeat, Zuluaga, Fripp, Salvado,
  Ourselin, Initiative et~al.}]{acosta2009automated}
\bibinfo{author}{O.~Acosta}, \bibinfo{author}{P.~Bourgeat},
  \bibinfo{author}{M.~A. Zuluaga}, \bibinfo{author}{J.~Fripp},
  \bibinfo{author}{O.~Salvado}, \bibinfo{author}{S.~Ourselin},
  \bibinfo{author}{A.~D.~N. Initiative}, et~al.,
\newblock \bibinfo{title}{Automated voxel-based 3{D} cortical thickness
  measurement in a combined {L}agrangian--{E}ulerian {PDE} approach using
  partial volume maps},
\newblock \bibinfo{journal}{Medical image analysis} \bibinfo{volume}{13}
  (\bibinfo{year}{2009}) \bibinfo{pages}{730--743}.
%Type = Inproceedings
\bibitem[{Cedilnik et~al.(2019)Cedilnik, Duchateau, Sacher, Ja{\"\i}s, Cochet,
  and Sermesant}]{cedilnik2019fully}
\bibinfo{author}{N.~Cedilnik}, \bibinfo{author}{J.~Duchateau},
  \bibinfo{author}{F.~Sacher}, \bibinfo{author}{P.~Ja{\"\i}s},
  \bibinfo{author}{H.~Cochet}, \bibinfo{author}{M.~Sermesant},
\newblock \bibinfo{title}{Fully automated electrophysiological model
  personalisation framework from {CT} imaging},
\newblock in: \bibinfo{booktitle}{International Conference on Functional
  Imaging and Modeling of the Heart}, \bibinfo{organization}{Springer}, pp.
  \bibinfo{pages}{325--333}.
%Type = Article
\bibitem[{Crane et~al.(2013)Crane, Weischedel, and
  Wardetzky}]{crane2013geodesics}
\bibinfo{author}{K.~Crane}, \bibinfo{author}{C.~Weischedel},
  \bibinfo{author}{M.~Wardetzky},
\newblock \bibinfo{title}{Geodesics in heat: A new approach to computing
  distance based on heat flow},
\newblock \bibinfo{journal}{ACM Transactions on Graphics (TOG)}
  \bibinfo{volume}{32} (\bibinfo{year}{2013}) \bibinfo{pages}{1--11}.
%Type = Article
\bibitem[{Varadhan(1967)}]{varadhan1967behavior}
\bibinfo{author}{S.~R.~S. Varadhan},
\newblock \bibinfo{title}{On the behavior of the fundamental solution of the
  heat equation with variable coefficients},
\newblock \bibinfo{journal}{Communications on Pure and Applied Mathematics}
  \bibinfo{volume}{20} (\bibinfo{year}{1967}) \bibinfo{pages}{431--455}.
%Type = Article
\bibitem[{Wang et~al.(2015)Wang, Zhang, Su, Shi, Caselli, Wang, Initiative
  et~al.}]{wang2015novel}
\bibinfo{author}{G.~Wang}, \bibinfo{author}{X.~Zhang}, \bibinfo{author}{Q.~Su},
  \bibinfo{author}{J.~Shi}, \bibinfo{author}{R.~J. Caselli},
  \bibinfo{author}{Y.~Wang}, \bibinfo{author}{A.~D.~N. Initiative}, et~al.,
\newblock \bibinfo{title}{A novel cortical thickness estimation method based on
  volumetric {L}aplace--{B}eltrami operator and heat kernel},
\newblock \bibinfo{journal}{Medical image analysis} \bibinfo{volume}{22}
  (\bibinfo{year}{2015}) \bibinfo{pages}{1--20}.
%Type = Book
\bibitem[{Grigoryan(2009)}]{grigoryan2009heat}
\bibinfo{author}{A.~Grigoryan}, \bibinfo{title}{Heat kernel and analysis on
  manifolds}, volume~\bibinfo{volume}{47}, \bibinfo{publisher}{American
  Mathematical Soc.}, \bibinfo{year}{2009}.
%Type = Book
\bibitem[{LeVeque(2007)}]{leveque2007finite}
\bibinfo{author}{R.~J. LeVeque}, \bibinfo{title}{Finite difference methods for
  ordinary and partial differential equations: steady-state and time-dependent
  problems}, \bibinfo{publisher}{SIAM}, \bibinfo{year}{2007}.
%Type = Article
\bibitem[{Wessner et~al.(2006)Wessner, Cervenka, Heitzinger, Hossinger, and
  Selberherr}]{wessner2006anisotropic}
\bibinfo{author}{W.~Wessner}, \bibinfo{author}{J.~Cervenka},
  \bibinfo{author}{C.~Heitzinger}, \bibinfo{author}{A.~Hossinger},
  \bibinfo{author}{S.~Selberherr},
\newblock \bibinfo{title}{Anisotropic mesh refinement for the simulation of
  three-dimensional semiconductor manufacturing processes},
\newblock \bibinfo{journal}{IEEE Transactions on Computer-Aided Design of
  Integrated Circuits and Systems} \bibinfo{volume}{25} (\bibinfo{year}{2006})
  \bibinfo{pages}{2129--2139}.
%Type = Book
\bibitem[{Krantz and Krantz(1999)}]{krantz1999handbook}
\bibinfo{author}{S.~G. Krantz}, \bibinfo{author}{S.~G. Krantz},
  \bibinfo{title}{Handbook of complex variables}, \bibinfo{publisher}{Springer
  Science \& Business Media}, \bibinfo{year}{1999}.
%Type = Book
\bibitem[{Greene and Krantz(2006)}]{greene2006function}
\bibinfo{author}{R.~E. Greene}, \bibinfo{author}{S.~G. Krantz},
  \bibinfo{title}{Function theory of one complex variable},
  volume~\bibinfo{volume}{40}, \bibinfo{publisher}{American Mathematical Soc.},
  \bibinfo{year}{2006}.
%Type = Misc
\bibitem[{Durrleman et~al.(2018)Durrleman, Prastawa, Routier, and et.
  al.}]{deformetrica}
\bibinfo{author}{S.~Durrleman}, \bibinfo{author}{M.~Prastawa},
  \bibinfo{author}{A.~Routier}, \bibinfo{author}{et. al.},
  \bibinfo{title}{{Deformetrica: learn from shapes}},
  \bibinfo{howpublished}{\url{http://www.deformetrica.org/}},
  \bibinfo{year}{2018}.
%Type = Inproceedings
\bibitem[{Lef{\`e}vre and Auzias(2015)}]{lefevre2015spherical}
\bibinfo{author}{J.~Lef{\`e}vre}, \bibinfo{author}{G.~Auzias},
\newblock \bibinfo{title}{Spherical parameterization for genus zero surfaces
  using {L}aplace-{B}eltrami eigenfunctions},
\newblock in: \bibinfo{booktitle}{International Conference on Geometric Science
  of Information}, \bibinfo{organization}{Springer}, pp.
  \bibinfo{pages}{121--129}.
%Type = Inproceedings
\bibitem[{Bohi et~al.(2019)Bohi, Wang, Harrach, Dinomais, Rousseau, and
  Lef{\`e}vre}]{bohi2019global}
\bibinfo{author}{A.~Bohi}, \bibinfo{author}{X.~Wang},
  \bibinfo{author}{M.~Harrach}, \bibinfo{author}{M.~Dinomais},
  \bibinfo{author}{F.~Rousseau}, \bibinfo{author}{J.~Lef{\`e}vre},
\newblock \bibinfo{title}{Global perturbation of initial geometry in a
  biomechanical model of cortical morphogenesis},
\newblock in: \bibinfo{booktitle}{2019 41st Annual International Conference of
  the IEEE Engineering in Medicine and Biology Society (EMBC)},
  \bibinfo{organization}{IEEE}, pp. \bibinfo{pages}{442--445}.
%Type = Article
\bibitem[{Yeo et~al.(2009)Yeo, Sabuncu, Vercauteren, Ayache, Fischl, and
  Golland}]{yeo2009spherical}
\bibinfo{author}{B.~T. Yeo}, \bibinfo{author}{M.~R. Sabuncu},
  \bibinfo{author}{T.~Vercauteren}, \bibinfo{author}{N.~Ayache},
  \bibinfo{author}{B.~Fischl}, \bibinfo{author}{P.~Golland},
\newblock \bibinfo{title}{Spherical demons: fast diffeomorphic landmark-free
  surface registration},
\newblock \bibinfo{journal}{IEEE transactions on medical imaging}
  \bibinfo{volume}{29} (\bibinfo{year}{2009}) \bibinfo{pages}{650--668}.
%Type = Inproceedings
\bibitem[{Rusinkiewicz(2004)}]{rusinkiewicz2004estimating}
\bibinfo{author}{S.~Rusinkiewicz},
\newblock \bibinfo{title}{Estimating curvatures and their derivatives on
  triangle meshes},
\newblock in: \bibinfo{booktitle}{Proceedings. 2nd International Symposium on
  3D Data Processing, Visualization and Transmission, 2004. 3DPVT 2004.},
  \bibinfo{organization}{IEEE}, pp. \bibinfo{pages}{486--493}.
%Type = Article
\bibitem[{Paragios(2003)}]{paragios2003level}
\bibinfo{author}{N.~Paragios},
\newblock \bibinfo{title}{A level set approach for shape-driven segmentation
  and tracking of the left ventricle},
\newblock \bibinfo{journal}{IEEE transactions on medical imaging}
  \bibinfo{volume}{22} (\bibinfo{year}{2003}) \bibinfo{pages}{773--776}.
%Type = Article
\bibitem[{Goldman(2005)}]{goldman2005curvature}
\bibinfo{author}{R.~Goldman},
\newblock \bibinfo{title}{Curvature formulas for implicit curves and surfaces},
\newblock \bibinfo{journal}{Computer Aided Geometric Design}
  \bibinfo{volume}{22} (\bibinfo{year}{2005}) \bibinfo{pages}{632--658}.
%Type = Inproceedings
\bibitem[{Albin et~al.(2016)Albin, Knikker, Xin, Paschereit, and
  d’Angelo}]{albin2016computational}
\bibinfo{author}{E.~Albin}, \bibinfo{author}{R.~Knikker},
  \bibinfo{author}{S.~Xin}, \bibinfo{author}{C.~O. Paschereit},
  \bibinfo{author}{Y.~d’Angelo},
\newblock \bibinfo{title}{Computational assessment of curvatures and principal
  directions of implicit surfaces from 3{D} scalar data},
\newblock in: \bibinfo{booktitle}{International Conference on Mathematical
  Methods for Curves and Surfaces}, \bibinfo{organization}{Springer}, pp.
  \bibinfo{pages}{1--22}.
%Type = Article
\bibitem[{Makki et~al.(2021)Makki, Salem, and ben Amor}]{makki2021towards}
\bibinfo{author}{K.~Makki}, \bibinfo{author}{D.~B. Salem},
  \bibinfo{author}{B.~ben Amor},
\newblock \bibinfo{title}{Towards the assessment of intrinsic geometry of
  implicit brain {MRI} manifolds},
\newblock \bibinfo{journal}{IEEE Access} \bibinfo{volume}{9}
  (\bibinfo{year}{2021}) \bibinfo{pages}{131054 -- 131071}.
%Type = Article
\bibitem[{Ogier et~al.(2021)Ogier, Hostin, Bellemare, and
  Bendahan}]{ogier2021overview}
\bibinfo{author}{A.~C. Ogier}, \bibinfo{author}{M.-A. Hostin},
  \bibinfo{author}{M.-E. Bellemare}, \bibinfo{author}{D.~Bendahan},
\newblock \bibinfo{title}{Overview of {MR I}mage segmentation strategies in
  neuromuscular disorders},
\newblock \bibinfo{journal}{Frontiers in Neurology} \bibinfo{volume}{12}
  (\bibinfo{year}{2021}) \bibinfo{pages}{255}.
%Type = Article
\bibitem[{Hamilton et~al.(1982)}]{hamilton1982three}
\bibinfo{author}{R.~S. Hamilton}, et~al.,
\newblock \bibinfo{title}{Three-manifolds with positive {R}icci curvature},
\newblock \bibinfo{journal}{J. Differential geom} \bibinfo{volume}{17}
  (\bibinfo{year}{1982}) \bibinfo{pages}{255--306}.
%Type = Book
\bibitem[{Weatherburn(2016)}]{weatherburn2016differential}
\bibinfo{author}{C.~E. Weatherburn}, \bibinfo{title}{Differential geometry of
  three dimensions}, volume~\bibinfo{volume}{1}, \bibinfo{publisher}{Cambridge
  University Press}, \bibinfo{year}{2016}.
%Type = Article
\bibitem[{Avenel et~al.(2014)Avenel, M{\'e}min, and
  P{\'e}rez}]{avenel2014stochastic}
\bibinfo{author}{C.~Avenel}, \bibinfo{author}{E.~M{\'e}min},
  \bibinfo{author}{P.~P{\'e}rez},
\newblock \bibinfo{title}{Stochastic level set dynamics to track closed curves
  through image data},
\newblock \bibinfo{journal}{Journal of mathematical imaging and vision}
  \bibinfo{volume}{49} (\bibinfo{year}{2014}) \bibinfo{pages}{296--316}.

\end{thebibliography}

\section{Appendices}
\begin{appendices}
% \appendix
\section{Riemannian manifolds and geodesic flows}
\label{Appendix A}

In Section~\ref{sec: Dirichlet_problem}, we have demonstrated the geodesic nature of the obtained geometric flows from a numerical point of view. In this Appendix, we will try to establish an analogy with the classical geodesic flows in differential geometry for parametric surfaces.
First we shall give an overview of the mathematical background of geodesics on Riemannian manifolds with an aim to help further explain the geodesic nature of the obtained curves and lengths.\\ 
A 2-manifold is just a surface in three dimensional space. In differential geometry, manifolds are "topological spaces" which locally resemble Euclidean. $(\mathcal{M}, g)$ is a called Riemannian manifold equipped with a metric $g$ if it is a differential manifold where the metric satisfies:
\begin{equation}
    g\in Sym^2_{++}(\mathcal{T}_x\mathcal{M}), \quad \forall \quad x \in \mathcal{M} 
\end{equation}

where $Sym^2_{++}$ is the group of symmetric positive definite bilinear forms defined on the tangent space.\\
While the Euclidean metric is the conventional Euclidean distance (\textit{i.e.} the distance of the straight line between two points), the Riemannian metric controlling intrinsic properties of a curved surface allows one to define distances and angles on the manifold, that best preserves the Euclidean geometry of the local tangent space. \\
Let $u$ and $v$ be two vectors of the tangent plane $\mathcal{T}_{x_0}\mathcal{M}$  with origin $x_0$, then the Riemann metric can be defined in the sense of inner (scalar) product by: $g(t,x_0,u,v)=||u||.||v||.cos(u,v)$ with $t \in[0,1]$.\\
In Riemannian geometry, a curve with tangential acceleration  zero is called a \textit{geodesic}. Geodesics are also characterized by the property of having vanishing intrinsic curvature. The shortest geodesic between two surface points $x_0$ and $x_1$ can be obtained by minimizing the following Dirichlet energy, subject to fixed endpoints $\gamma(0)=x_0$ and $\gamma(1)=x_1$:
\begin{equation}
    E_D(\gamma) = \frac{1}{2} \int_{0}^{1} |\gamma^{'} (t)|^2 dt = \frac{1}{2} \int_{0}^{1} g_{\gamma(t)} (\gamma^{'} (t), \gamma^{'}(t))  dt  
    \label{dirichlet_energy}
\end{equation}

The length of the shortest geodesic curve $\gamma$ is called a geodesic distance, it is like a flexibility when we define distance:
\begin{equation}
    L(\gamma) = \int_{0}^{1} \sqrt{g_{\gamma(t)} (\gamma^{'} (t), \gamma^{'}(t))}  dt 
    \label{geod_equation}
\end{equation}

If the metric $g$ reduces to the Kronecker delta $\delta_{ij}$, then the Eq.~\ref{geod_equation} reduces to:
\begin{equation}
    L(\gamma) = \int_{0}^{1} ||\gamma^{'} (t)|| dt 
    \label{geod_scalar_equation}
\end{equation}

\begin{figure}[h!]
    \centering
    \includegraphics[width=0.35\linewidth]{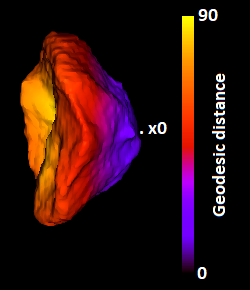}
    %\vspace{0.7cm}
    \caption{Geodesic distances on Riemannian manifolds (expressed in mm for the bladder example): the lengths of the shortest paths connecting each vertex to the vertex $x_0$, along the manifold.}
    \label{fig:Riemannian_geodesics}
 \end{figure}

Fig.~\ref{fig:Riemannian_geodesics} illustrates the Riemannian geodesic distances between the point $x_0 \in \mathcal{M}$ and each surface point $x_1 \in  \mathcal{M}$ (example for one reconstructed bladder surface).

In Riemannian geometry, each spherical mapping method is subsumed by Ricci flow (the generic model). Like the heat equation developed by Fourier $\frac{\partial h}{\partial t}-\alpha \Delta h = 0$, where $\alpha$ is the diffusivity constant, the Ricci flow is a non-linear PDE, but resulting from geometry rather than physics~\cite{hamilton1982three}.
The Ricci flow is therefore a way of changing the metric tensor over time until it converges toward a positive constant (the sphere curvature), according to the parabolic PDE $\frac{\partial g}{\partial t} + 2 Ric_g=0$.
Like the Riemannian metric $g$, the Ricci curvature tensor $Ric_g$ is a symmetric bilinear form defined as the trace of the curvature tensor. It is therefore a mean of the information contained in the Riemann curvature tensor. \\
With a separation of space and time variables, the above defined heat equation (parabolic PDE) can be splitted into two elliptic PDEs that can be easily solved separately. In this particular case, the separation of variables is a result of the spectral theorem. More details are provided in Appendix~\ref{Appendix B}. 

In the same analogy, and assuming that the solution is becoming time-independent under the imposed constraints, the Dirichlet energy to be minimized is the one given in Eq.~\ref{Dirichlet_energy}. 
The harmonic map $h$ minimizing this energy is the optimal solution of an elliptic PDE. By definition, a map $h$ is called harmonic if its Laplacian vanishes, $\Delta h = 0$. We then propose to recover the curve-shortening flow by integrating the normalized gradient of $h$.\\

\begin{figure}[ht]
\centering
\subfigure{\includegraphics[scale=0.5]{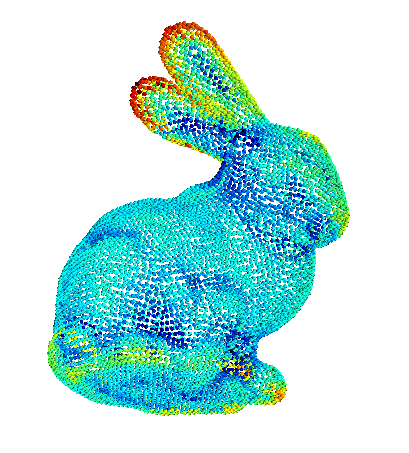}}
\hspace{0.2cm}
\subfigure{\includegraphics[scale=0.5]{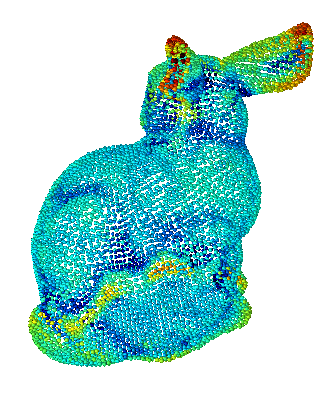}}
\hspace{0.2cm}
\subfigure{\includegraphics[scale=0.5]{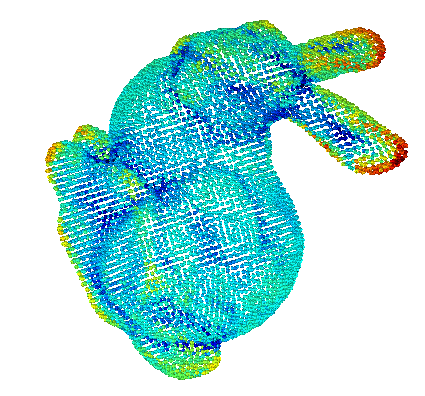}}

\caption{\label{mean_curv} Obtained mean curvature for the isosurface $h=0.98$.}
\end{figure}

The resulting geometric flow is gradient like, i.e. a process that modify the surface curve toward the sphere by moving its points perpendicularly to the curve (the tangential acceleration vanishes), at a speed proportional to the curvature. The mean curvature (extrinsic) of the surface $\mathcal{M}$ can be derived from the obtained harmonic interpolant according to the divergence formula: $\kappa(x) = \frac{1}{2} div (\frac{\nabla h(x)}{||\nabla h(x)||}), \forall x \in \mathcal{M}$. A visual example is provided in Fig~\ref{mean_curv}.

%\begin{figure}[ht]
%\centering
%\subfigure{\includegraphics[scale=.2]{mc1.png}}
%\hspace{0.7cm}
%\subfigure{\includegraphics[scale=.2]{mc2.png}}
%\vspace{0.4cm}

%\subfigure{\includegraphics[scale=.2]{mc3.png}}
%\hspace{0.7cm}
%\subfigure{\includegraphics[scale=.2]{mc4.png}}
%\vspace{0.4cm}

%\subfigure{\includegraphics[scale=.2]{mc5.png}}
%\hspace{0.7cm}
%\subfigure{\includegraphics[scale=.2]{mc6.png}}
%\vspace{0.3cm}

%\subfigure{\includegraphics[scale=.2]{mc7.png}}
%\hspace{0.7cm}
%\subfigure{\includegraphics[scale=.2]{mc9.png}}
%\vspace{0.3cm}
%\caption{\label{mean_curv} Obtained surface mean curvature from the function $h$, minimizing the Dirichlet energy. The results are displayed on the 2D surface meshes extracted from 3D volumes using the marching cubes algorithm (isosurface value, $h=0.01$).}
%\end{figure}

%\hl{ To make the surface speed totally independent of the curvature (\textit{i.e.} to have a unit speed geometric/gradient flow overall $\Omega$), we normalize the gradient of $h$ to obtain $N$.}

Replacing $\gamma '$ with $N= \frac{\nabla h}{||\nabla h||}$ in Eq.~\ref{geod_scalar_equation}, and assuming again that the solution is time-independent, then the geodesic lengths/distances can be defined (in the sense of optimization) by:

\begin{equation}
    L^*(\Phi_N) = \operatorname*{min}_L ( \int_{\partial_1 \Omega}^{\partial_0 \Omega}\left || \nabla L - N \right ||^2 dx)%\int_{x \in \Omega}  N (x) \quad dx 
    \label{geod_scal_equation}
\end{equation}
The numerical integration of the above equation over $\Omega$ is performed using an iterative Gauss-Seidel relaxation method, for locally solving the following PDE:

\begin{equation}
\nabla L.N =  ||\nabla L||.||N||.cos(\nabla L,N)= 1
\end{equation}

where the left term ($||\nabla L||.||N||.cos(\nabla L,N)$) is the scalar product between the two vectors, while the right term (1) controls the distances and angles: by forcing infinitesimal lengths to be unitary and by specifying the direction (angle=0) in such a way that the gradient of the scalar distance function $\nabla L$ coincides with the velocity vector field $N$. Thus, the integrated trajectory is qualified as a geodesic and its length is locally defined and shortest in the sense of scalar product.

% Furthermore, since $\vec{v}_x$ is a unit vector, then its derivative (i.e. the local acceleration) vanishes in Cartesian coordinates. This means that continuously integrating the velocity vector field between boundaries results in a geodesic flow.

% \appendix

\section{Normalized gradient vector field}
\label{Appendix C}

\begin{figure}[!t]
\centering
\subfigure[Input shape]{\includegraphics[scale=.15]{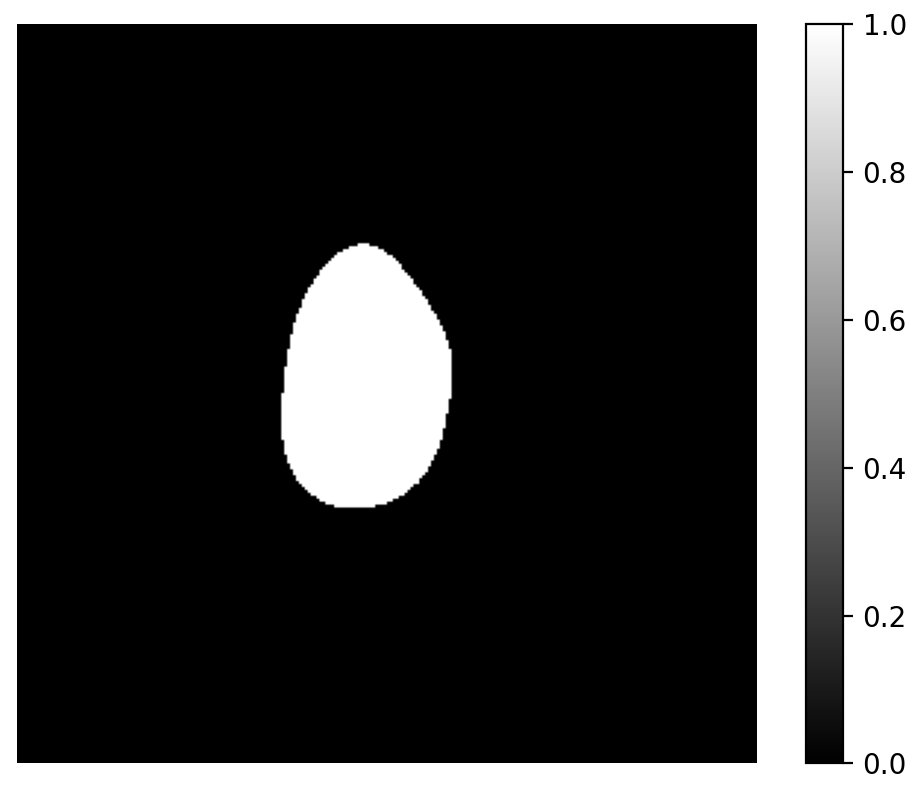}}
\hspace{0.5cm}
\subfigure[$h$]{\includegraphics[scale=.15]{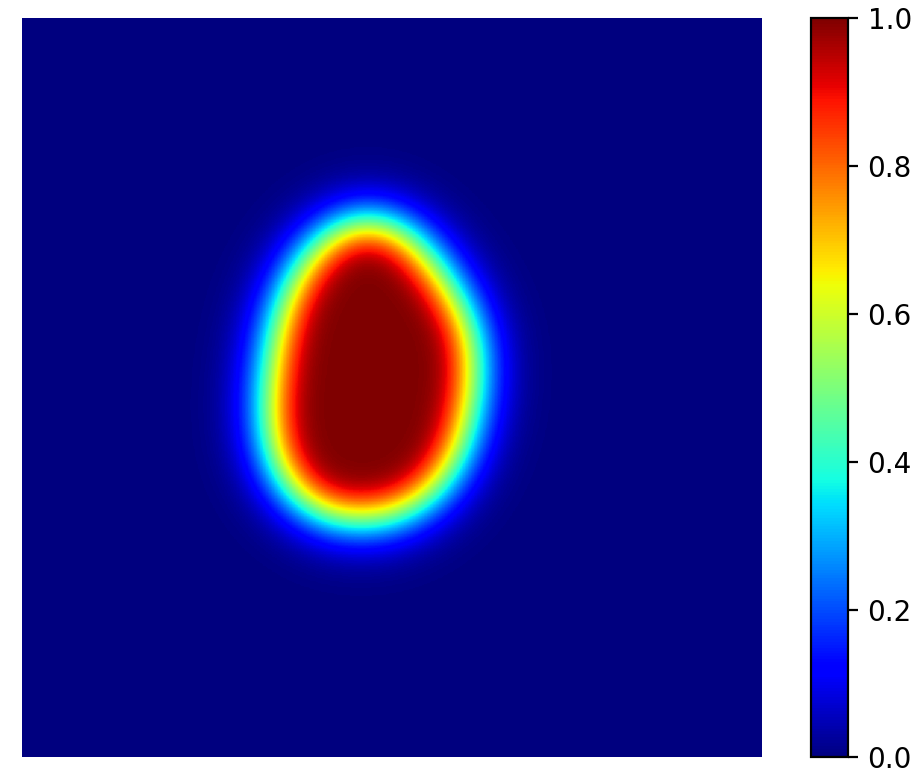}}
\hspace{0.5cm}
\subfigure[$||\nabla h||_2$]{\includegraphics[scale=.15]{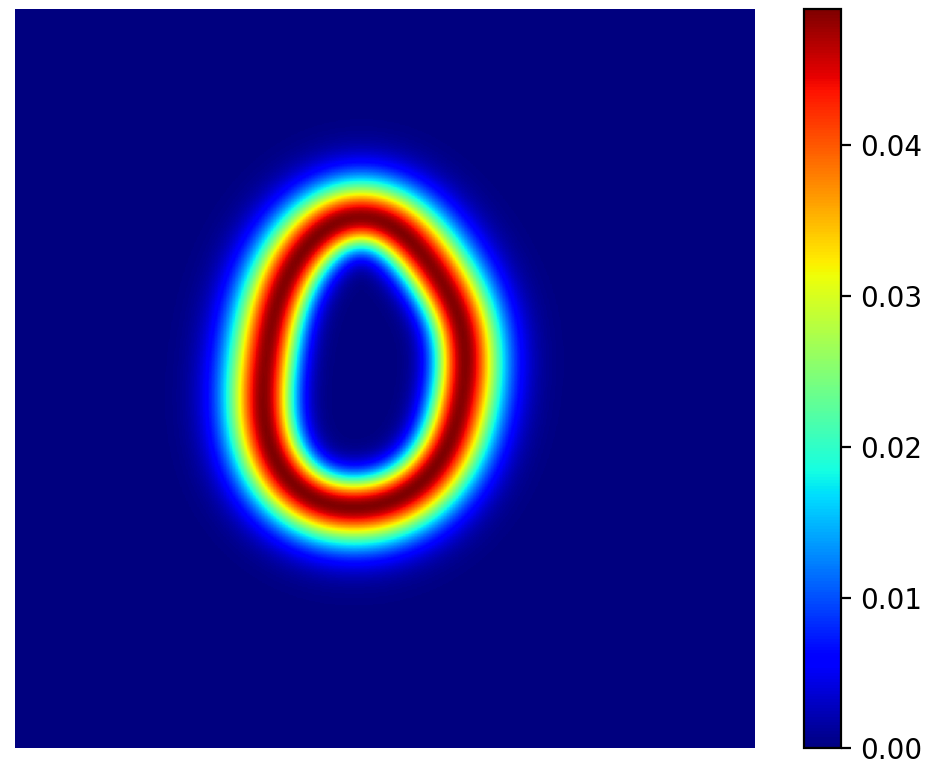}}

\subfigure[Fast marching distance $L_{FM}$]{\includegraphics[scale=.15]{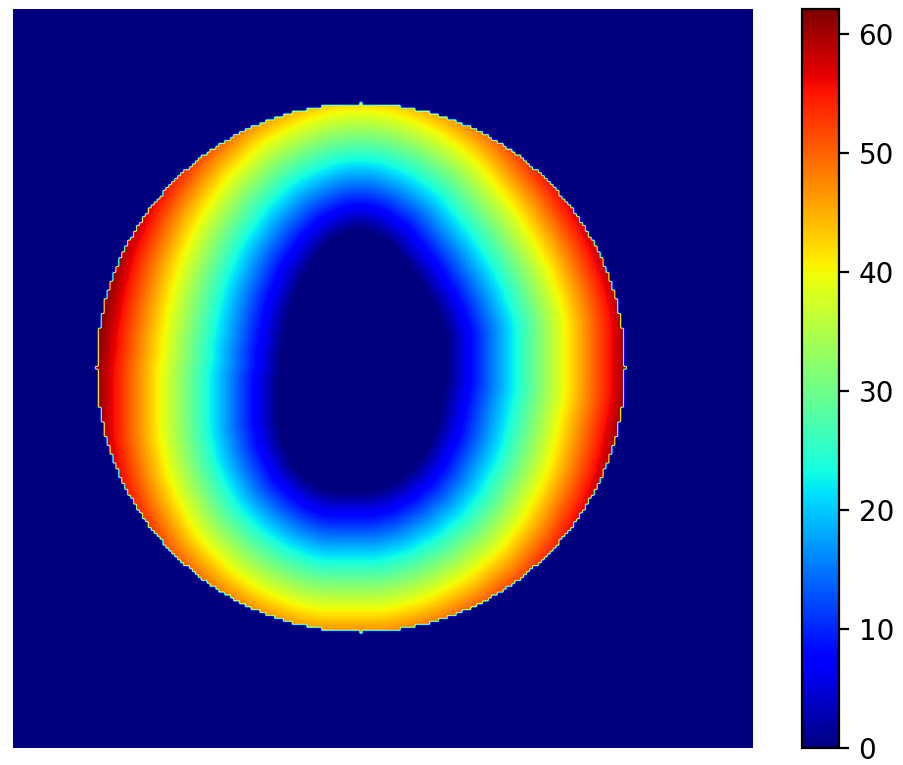}}
\hspace{0.5cm}
\subfigure[Our distance $L$]{\includegraphics[scale=.15]{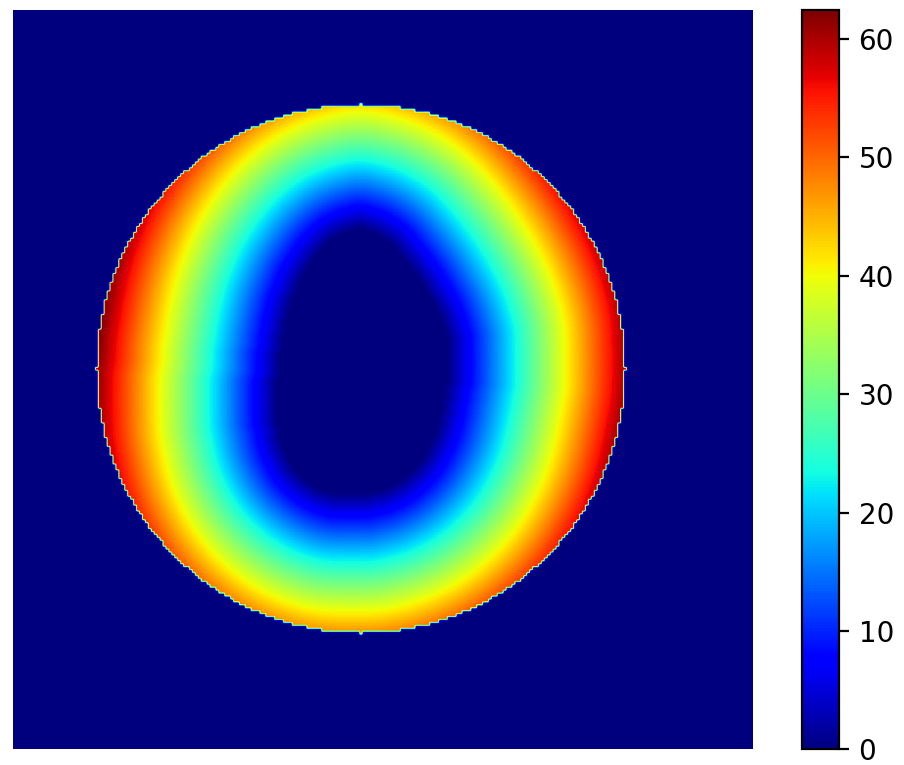}}
\hspace{0.5cm}
\subfigure[$L - L_{FM}$]{\includegraphics[scale=.15]{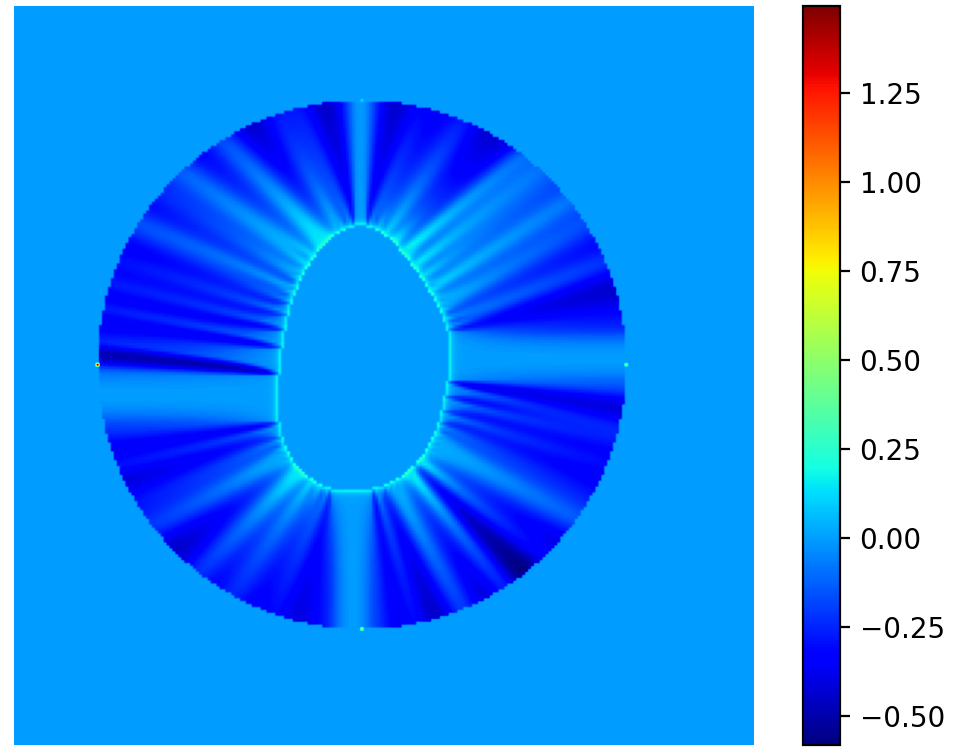}}

\subfigure[$\nabla L_{FM}$]{\includegraphics[scale=.15]{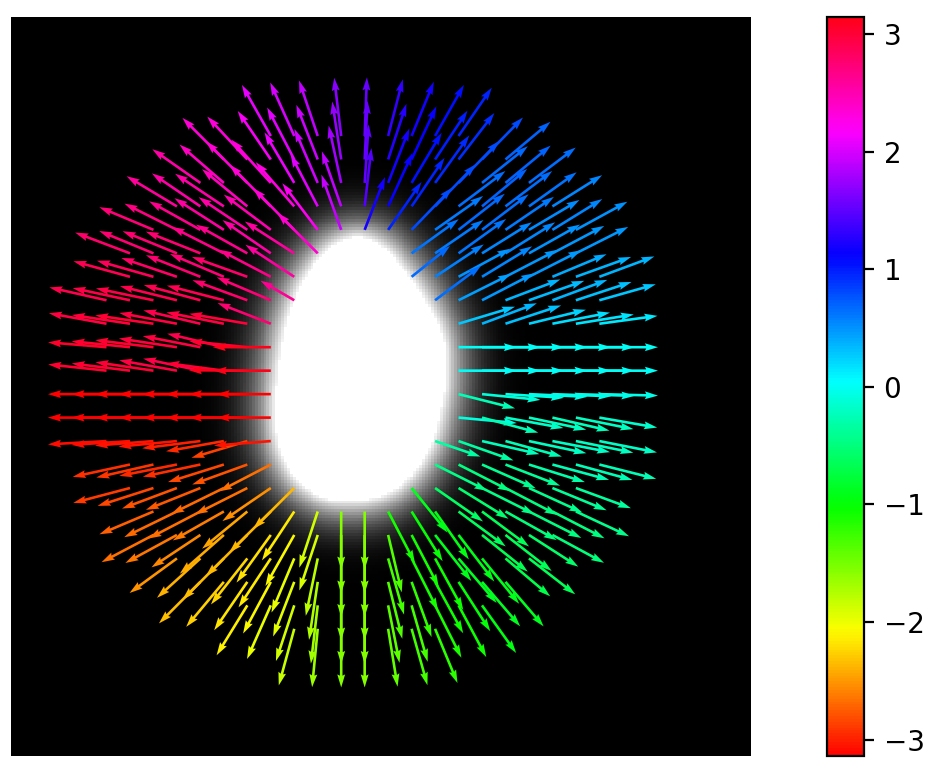}}
\hspace{0.5cm}
\subfigure[$N$]{\includegraphics[scale=.15]{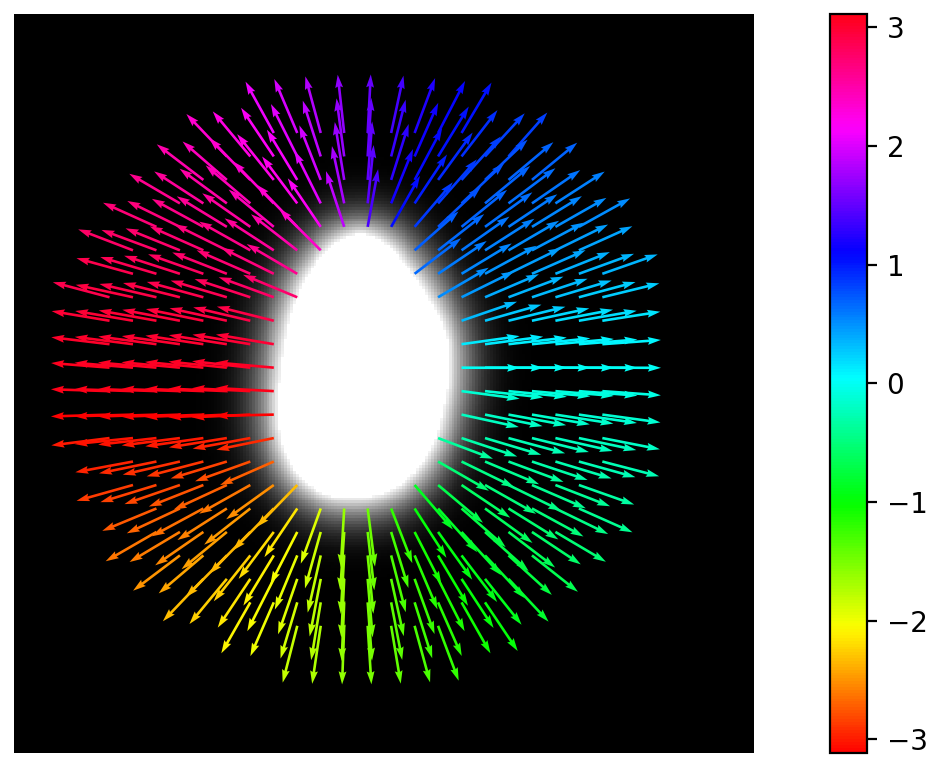}}
\hspace{0.5cm}
\subfigure[Scalar product: $\nabla L_{FM} \cdot N$]{\includegraphics[scale=.15]{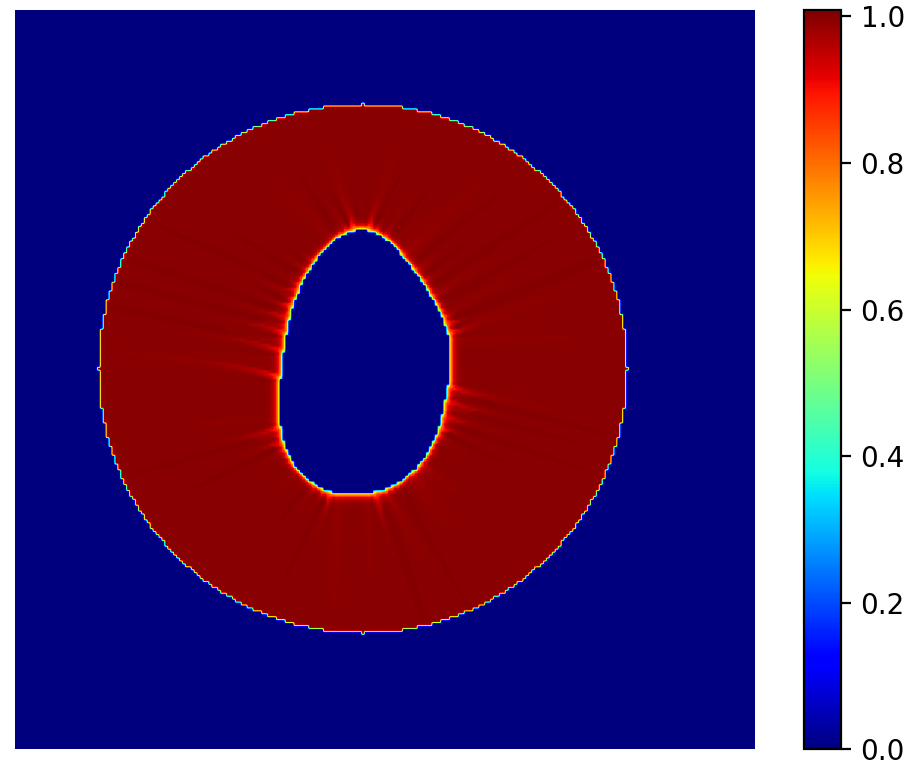}}

\caption{Method comparison with the Fast marching method. Comparison is based on the scalar product of their gradient vector fields.}
    \label{fig:conservativeness}
\end{figure}

To prove that the computed correspondence trajectories are geodesics, it is mandatory to prove that $N$ is conservative. For this, it is sufficient to prove that $N$ is curl free (irrotational) in the vicinity of each heat isosurface. But, this is non-trivial, since $||\nabla h|| \neq cste$ generally so that:\\
$curl(N) = \nabla \times N =  \nabla \times (\frac{\nabla h}{||\nabla h||}) = \nabla (\frac{1}{||\nabla h||}) \times \nabla h + \frac{1}{||\nabla h||} \underbrace{\nabla \times \nabla h}_{\text{=~$curl(\nabla h) =0$.}} = \frac{1}{||\nabla h||^2} \nabla(||\nabla h||) \times \nabla h$.\\ 
%And since $||\nabla h|| \neq cste$ generally, then the problem arises.\\
Overall $\Omega$, for each isosurface $h = h_{iso}$, the fact that $curl(N) = 0$ depends essentially on how the vector field $N$ is defined away from that isosurface. However, on each isosurface, $N$ is well defined and conservative~\cite{weatherburn2016differential}.\\
\begin{itemize}
\item \textbf{Formulation 1:} Suppose $N$ to be the normalized gradient of $h$ over $\Omega$. In this case, $\nabla \times N = 0$, when evaluated on the isosurface $h_{iso}$ if and only if $||\nabla h|| = cste$.
\item \textbf{Formulation 2:} Define the function $f = \frac{1}{||\nabla h||} (h - h_{iso})$ inside $\Omega$. Suppose now that the surface we are interested in is $\Gamma=\{f=0\}$. Developing the expression of the gradient of $f$, we obtain: 
\begin{equation}
    \nabla f = \frac{\nabla h}{||\nabla h||}- \frac{(h-h_{iso}) \nabla h \cdot \nabla^2 h}{||\nabla h||^3},
    \label{deriv}
\end{equation}
where $\nabla^2(.)$ is the Hessian operator. It is clear that the second term of the right-hand side of the Eq.~\eqref{deriv} vanishes on $\Gamma$ (since $h = h_{iso}$). Consequently, $\nabla f$ restricted to the surface is still the unit normal vector field $N$. Moreover, $\nabla \times \nabla f$ is clearly zero as argued before.
\end{itemize}
More generally, for a compact smooth surface $\Gamma \subset \mathbb{R}^n$, there exists a real $r>0$ such that on the set $\omega = \{ x \in \mathbb{R}^3 \quad | \quad dist(x, \Gamma) <r\}$ one can solve the Eikonal PDE  $|\nabla f| = 1$ to get a continuous function $f: \omega \mapsto \mathbb{R}$ which satisfies $\Gamma = f^{-1}(0)$ and such that $\nabla f$ is the unit normal vector field for any level set $f^{-1}(c)$. Under this formulation, the unit normal vector field $N = \nabla f$ is curl-free in a narrowband $\omega$ of the surface $\Gamma$. The real number $r$ is related to the radius of the surface mean curvature. Fig.~\ref{fig:conservativeness}.(c) illustrates that for each set $\omega_c \subset \Omega$ and for a small radius $r$, we have $|\nabla h| = cste$.

%In the attached Notebook (see also Fig.~\ref{fig:grad_norm}), we provide a detailed and complete example showing that the gradient norm $|\nabla h$| is always constant in a narrowband of each each isosurface.}

\section{Background on harmonic analysis for 2-manifolds}
\label{Appendix B}

The harmonic analysis on Riemannian manifolds is in fact a generalization of Fourier spectral shape analysis. For which, the Fourier modes on $[0,1]$ are solutions of $f''= -\lambda f$ with boundary conditions (Dirichlet, Neumann, etc.). \\
Again, the more general heat equation is the principal ingredient. This equation reads: 
\begin{equation}
    \frac{\partial h}{\partial t}= \alpha \Delta h
    \label{heat_ref}
\end{equation}

where the function $h$ describes the temperature distribution over the manifold.\\
In the homogeneous case, the boundary conditions satisfy $h(\partial_{x_0} \mathcal{M})= h(\partial_{x_L} \mathcal{M})=0$.
Looking now for a solution which is not identically zero satisfying the boundary conditions and for which the dependence of $h$ on $x$, $t$ is separated, where $x \in \mathcal{M}$ and $t$ denotes the time, that is:
\begin{equation}
    h(x,t) = f(x).T(t).
\end{equation}

Using the product rule and substituting $h$ back into equation~\ref{heat_ref}, we obtain:
\begin{equation}
    \frac{T'(t)}{\alpha T(t)}= \frac{\Delta f(x)}{f(x)}
\end{equation}

Since the left hand side depends only on $t$ and the right hand side only on $x$, then both sides are equal to some constant value $- \lambda$. Thus:

\begin{equation}
    T'(t) = - \alpha \lambda T(t),
\end{equation}

and

\begin{equation}
    \Delta f(x) = -  \lambda f(x)
    \label{eigenvalue_problem}
\end{equation}

where: $-\lambda$ represent the eigenvalues for both differential operators, and $T(t)$ and $f(x)$ are corresponding eigenfunctions. If we only consider the heat vibration modes in the spatial domain, then the heat equation reduces to Eq.~\ref{eigenvalue_problem} (a special case of the Poisson equation $\Delta \phi = f$, with $f = \lambda \phi$), for which, the solutions become time-independent.\\
\textbf{Definition:}\\
Given a complete 2-Riemannian manifold without boundary (closed surface), $(\mathcal{M},g)$ equipped with a local metric tensor $g$ and with a local coordinate system $u: x \in \mathcal{M} \rightarrow \mathbb{R}^2/$ $u(x)=(u_1,u_2)$, the LBO acting on $C^{\infty}$ functions is defined by:
\begin{equation}
\label{eq:LBO}
    \Delta_{\mathcal{M}}f(u)= \frac{1}{\sqrt{{det (g)}}} \sum_{i,j} \partial_{u_j}(\sqrt{{det(g)}}g^{i,j}\partial_{u_i}f(u))
\end{equation} 

\begin{figure}[!t] 
\centering
\includegraphics[scale=.7]{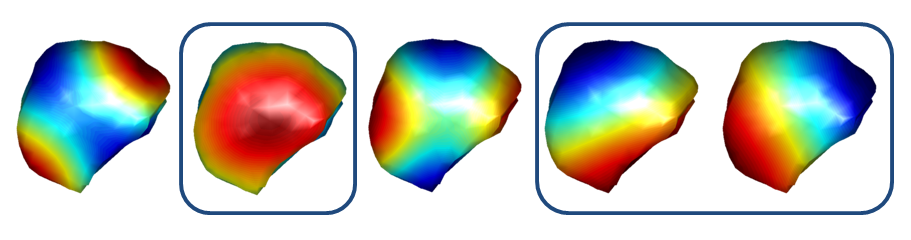}
\caption{\label{fig:eigenfunctions} Surface spectral analysis: five first non-trivial LBO eigenfunctions (sorted from left to right). Colormap goes from
blue (negative) to red (positive). The eigenfunctions with only two nodal domains $\Phi_2$, $\Phi_4$, and $\Phi_5$ were used for spherical parameterization.}
\end{figure}

Consider now the Dirichlet eigenvalue problem:

\begin{equation}
    \left\{
\begin{array}{l}
  \Delta f = - \lambda f  \quad in \quad \Omega,  \\
  f = 0 \quad on \quad \partial \Omega.  \\
\end{array} \right.
\end{equation}

where: $\Omega \subset \mathcal{M}$. Solving this problem gives an infinite set of  sorted eigenvalues $0 \leq \lambda_1 \leq \lambda_2 \leq \lambda_3 \ldots$ and a set of corresponding eigenfunctions $\Phi_1, \Phi_2, \Phi_3 \ldots$ (orthogonal in the sense of the scalar product), which form together an orthonormal basis of $L^2(\mathcal{M})$. An example showing the five first eigenfunctions for a bladder surface is given in Fig.~\ref{fig:eigenfunctions}.

As proposed in~\cite{lefevre2015spherical}, the principle of spherical mapping is as follows: for a surface point $x \in \mathcal{M}$, there exists a $C^{\infty}$ diffeomorphism $\Phi$, satisfying:
\begin{equation}
\begin{matrix}
\Phi : \mathcal{M} \rightarrow \mathbb{S}^2 \\
x \longmapsto \frac{(\Phi_1(x),\Phi_2(x),\Phi_3(x))}{\sqrt{{\Phi_1(x)^2}+\Phi_2(x)^2+\Phi_3(x)^2}}
\end{matrix}
\end{equation}

\begin{figure}[!h]
\centering
%\begin{minipage}{0.36\textwidth}
\subfigure{\includegraphics[scale=0.18]{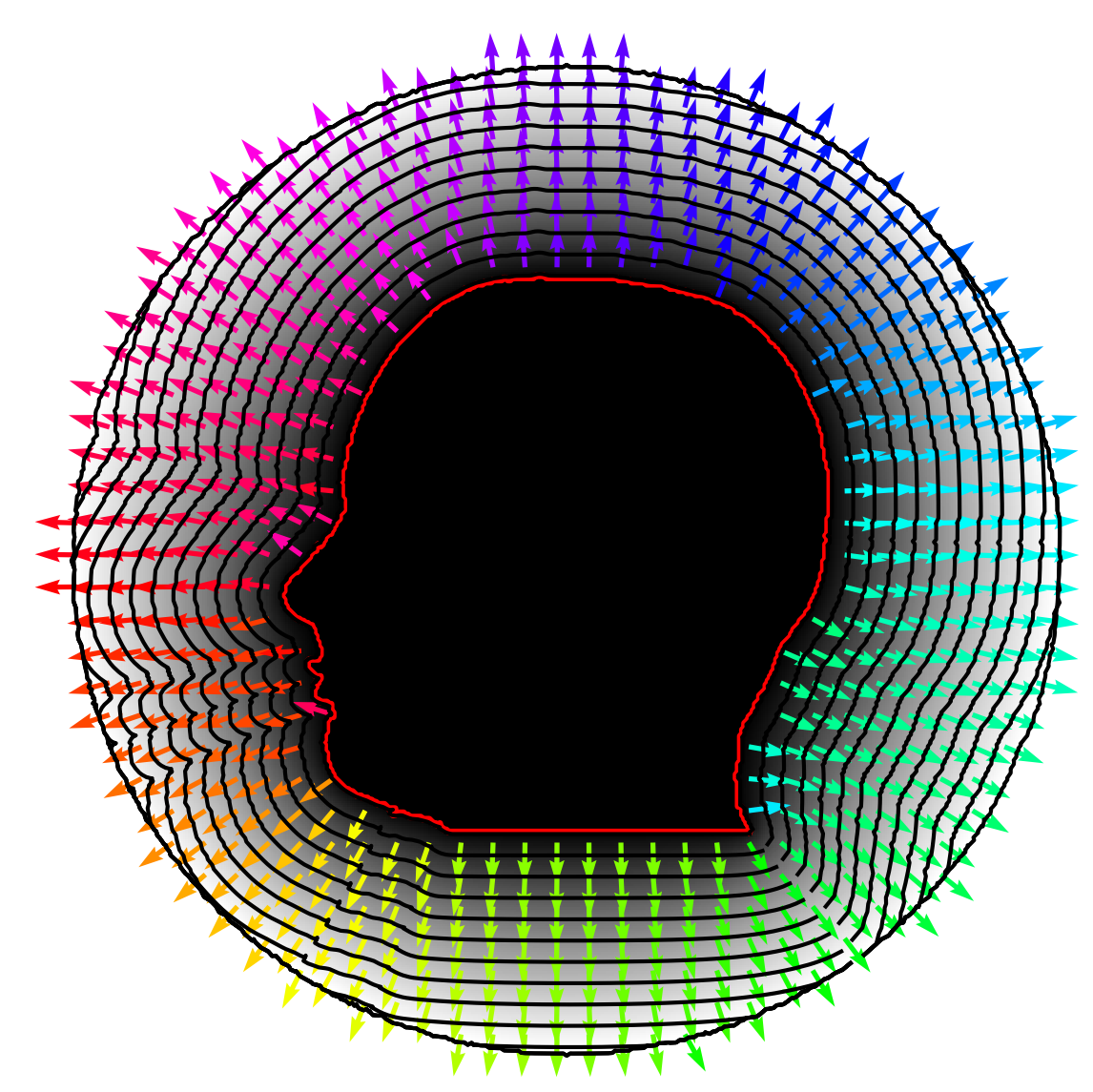}}
\hspace{0.2cm}
\subfigure{\includegraphics[scale=1.0]{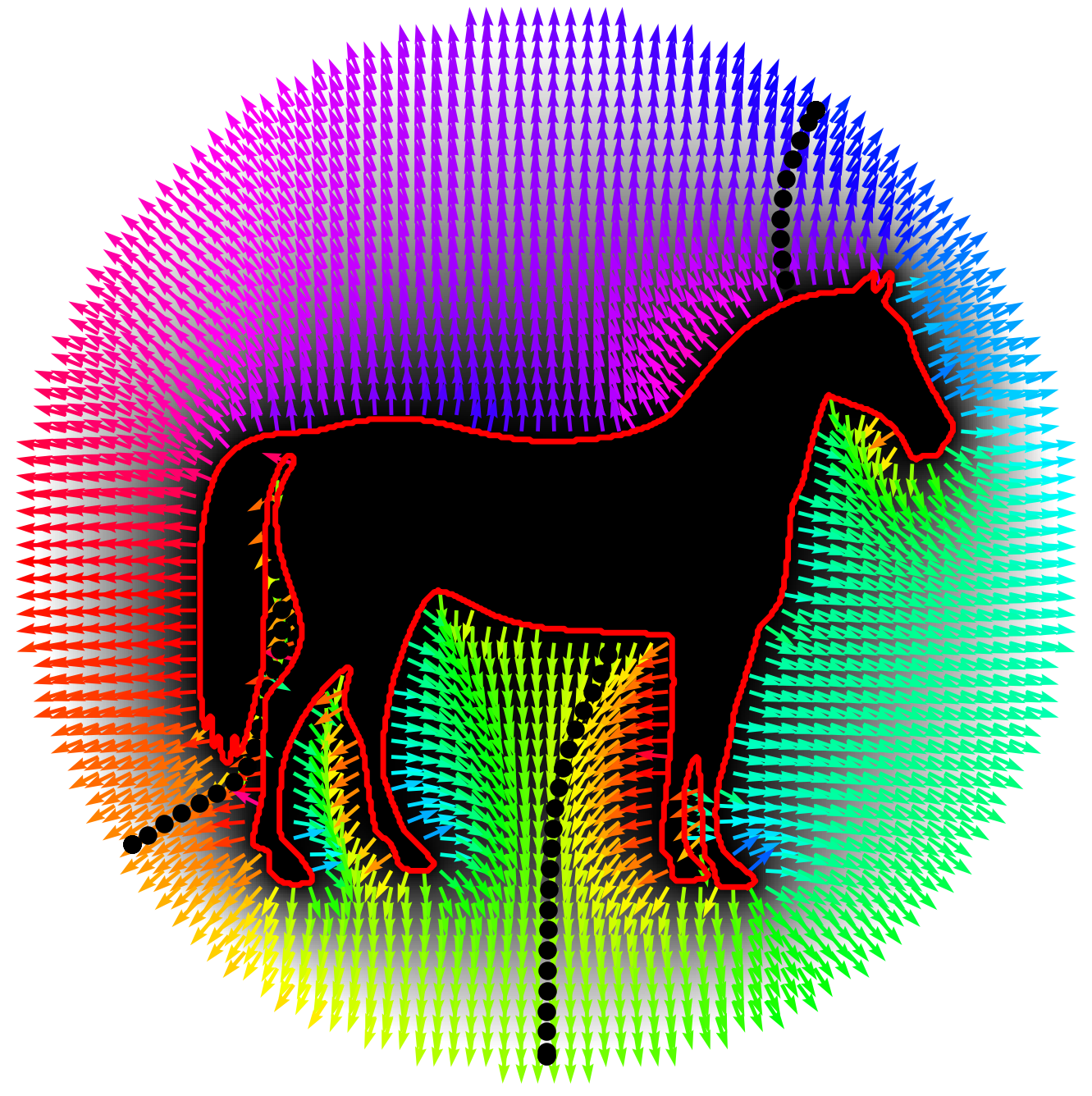}}
\hspace{0.2cm}
\subfigure{\includegraphics[scale=0.7]{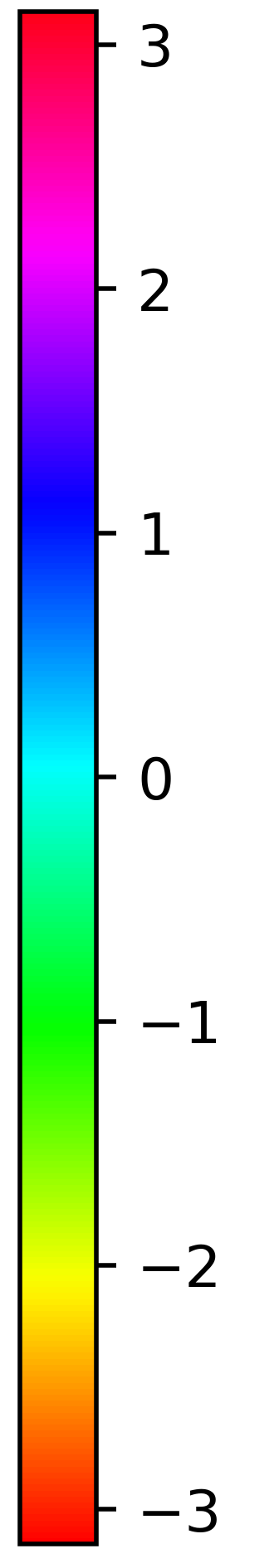}}
\caption{\label{velocity_integration} Numerical integration of the velocity vector field (2D examples). Left: the dark curves indicate the trajectory of shape contour,or equivalently, the isocurves of the distance function. Right: the entire trajectories of some contour points, where the colorbar encodes the direction of the gradient vector field $N$ in radians. Note that the principle remains the same in 3D.}
\end{figure}

By solving Laplace equation under the condition of the equilibrium state, one can redefine the above defined diffeomorphic spherical mapping with the following application:
\begin{equation}
\begin{matrix}
\Phi_n : \mathcal{M} \rightarrow \mathbb{S}^2 \\
x \longmapsto \frac{\Phi_N(x)}{||\Phi_N(x)||}=\frac{(\Phi_{N_i}(x),\Phi_{N_j}(x),\Phi_{N_k}(x))}{\sqrt{{\Phi_{N_i}(x)^2}+\Phi_{N_j}(x)^2+\Phi_{N_k}(x)^2}},
\end{matrix}
\end{equation}

such that $\Phi_N(x_0) = (x_0 - c_0) + \int_{x_0}^{\partial_0 \Omega} N(x) dx$, where $c_0$ is the center of the surrounding sphere, and $x_0 \in \mathcal{M}$. Intuitively, integrating the normal velocity vector field $v=N$ (defined in Section~\ref{sec: Dirichlet_problem}) gives a smooth bijective map $\Phi_N$ which transforms each surface point $x_0 \in \mathcal{M}$ into its unique corresponding point in the surrounding sphere $x_1$, modulo translation (see examples of Fig.~\ref{velocity_integration}). The way in which surface points move along the curve normal directions within the Eulerian Framework resembles the principle used in level set methods to track evolving curves while providing a notion of correspondence~\cite{avenel2014stochastic}. 
%The use of this proposed technique will solve the singularity problem in the resulting sphere-mesh since the correspondence trajectories may never intersect.

% \hl{Assuming that the LBO eigenfunction-based mapping can only deal with genus-0 surfaces, the results of the application of the proposed technique to the torus (genus-1) were surprising, but promising, simply because the torus is not homeomorphic to the 2-sphere according to the Poincaré conjecture. These particular results can open a new avenue for future research in topology.}
\end{appendices}

% where $\Phi_1$, $\Phi_2$ and $\Phi_3$ are the three first orthogonal eigenfunctions of the LBO, satisfying $-\Delta \Phi_i = \lambda_i \Phi_i $ with $\lambda_i$ is the positive eigenvalue corresponding to $\Phi_i$.\\

\end{document}